\documentclass[]{article}

\usepackage{microtype}
\usepackage{subfigure}
\usepackage{booktabs} 
\usepackage{thmtools}
\usepackage{thm-restate}

\newcommand\numberthis{\addtocounter{equation}{1}\tag{\theequation}}
\usepackage{hyperref}

\usepackage[accepted]{icml2021}

\icmltitlerunning{Matching bandits}

\usepackage[utf8]{inputenc}

\usepackage{amsmath,amsfonts,hyperref}
\usepackage{dsfont}
\usepackage{array,multirow,graphicx}
\usepackage{float}
\usepackage{latexsym}
\usepackage{amssymb}
\usepackage{caption}

\usepackage{comment}

\usepackage{slantsc}
\DeclareMathOperator*{\argmax}{arg\,max} 

\makeatletter
\newenvironment{algosketch}[1][htb]{%
    \renewcommand{\ALG@name}{Algorithm Sketch}
   \begin{algorithm}[#1]%
  }{\end{algorithm}}
\makeatother

\newcommand{\STAB}[1]{\begin{tabular}{@{}c@{}}#1\end{tabular}}

\newtheorem{theorem}{Theorem}[section]

\newtheorem{lemma}[theorem]{Lemma}
\newtheorem{proposition}[theorem]{Proposition}

\newtheorem{hyp}{Assumption}

\newcommand{\E}{\mathds{E}}

\newcommand{\Pb}{\mathds{P}}
\newcommand{\cA}{\mathcal{A}}
\newcommand{\cB}{\mathcal{B}}

\newcommand{\cS}{\mathcal{S}}

\newcommand{\cU}{\mathcal{U}}
\newcommand{\cV}{\mathcal{V}}

\newcommand{\SRankElim}{{\sc Pair-Elim}}
\newcommand{\RankElim}{{\sc Rank1Elim}}
\newcommand{\IBLearning}{{\sc Pair-Select}}

\newcommand{\RMono}{{\sc Pair-Elim-Mono}}
\newcommand{\SMDC}{{\sc simple-Adaptive-Matching}}
\newcommand{\SMDCname}{Simple Adaptive Matching}
\newcommand{\MatchingId}{{\sc Matching-id}}
\newcommand{\MDC}{{\sc Adaptive-Matching}}

\newcommand{\CMSampling}{{\sc Sample-Matching}}

\newcommand{\fs}[1]{\textcolor{red}{#1}}
\newcommand{\mv}[1]{\textcolor{blue}{#1}}

\begin{document}

\twocolumn[
\icmltitle{Pure Exploration and Regret Minimization in Matching Bandits}



\icmlsetsymbol{equal}{*}

\begin{icmlauthorlist}
\icmlauthor{Flore Sentenac}{equal,ENSAE}
\icmlauthor{Jialin Yi}{equal,LSE}
\icmlauthor{Cl\'ement Calauz\`enes}{CRITEO}
\icmlauthor{Vianney Perchet}{ENSAECRITEO}
\icmlauthor{Milan Vojnovi\'c}{LSE}
\end{icmlauthorlist}

\icmlaffiliation{ENSAE}{CREST, ENSAE Paris, Palaiseau, France}
\icmlaffiliation{ENSAECRITEO}{CREST, ENSAE Paris, Palaiseau, France \& Criteo AI Lab, Paris, France}
\icmlaffiliation{LSE}{London School of Economics, London, UK}
\icmlaffiliation{CRITEO}{Criteo AI Lab, Paris, France}

\icmlcorrespondingauthor{Flore Sentenac}{flore.sentenac@gmail.com}
\icmlcorrespondingauthor{Jialin Yi}{J.Yi8@lse.ac.uk}

\icmlkeywords{Matching, Bandit Algorithm, Regret Minimization, Pure Exploration, Combinatorial Bandits}

\vskip 0.3in
]



\printAffiliationsAndNotice{\icmlEqualContribution} 

\begin{abstract}
Finding an optimal matching in a weighted graph is a standard combinatorial problem. We consider its  semi-bandit version where either a pair or a full matching is sampled sequentially. We prove that it is possible to leverage a rank-1 assumption on the adjacency matrix to reduce the sample complexity and the regret of off-the-shelf algorithms up to reaching a linear dependency in the number of vertices (up to $\mathrm{poly}\log$ terms).
\end{abstract}

\section{Introduction}



Finding \emph{matchings} in 
graphs, i.e., subsets of edges without common vertices, is a long standing problem \cite{lovasz2009matching} with many different applications in economics \cite{roth2004kidney}, operations research \cite{wheaton1990vacancy}, and machine learning \cite{mehta2012online}. We consider here its sequential variant where at each time step $t$, an agent chooses a matching $m_t$ of some graph, defined by its unknown weighted adjacency matrix $W$ (with bounded elements in $(0,1)$), and observes noisy evaluations of the chosen entries $\{X_{i,j,t}: (i,j)\in m_t\}$, with $\E[ X_{i,j,t}]=W_{i,j}$. This problem obviously falls in the realm of combinatorial bandits \cite{cesa2012combinatorial}, but we aim at leveraging a specific  structural property: in many relevant examples, $W$ is a rank 1 matrix.

Two different types of graphs are relevant for matchings, \emph{bipartite} and \emph{monopartite}, and we are going to consider both of them (even though the latter is  more intriguing, the former, maybe more intuitive, will serve as a warm-up and to convey insights). In the bipartite case, the set of vertices is separated in two distinct subsets $\cU$ and $\cV$ (of respective sizes $N$ and $M$) and edges only exist across subsets, not within. The rank-1 adjacency matrix $W$ is then a $N\times M$ matrix, that can be written as $W=uv^\top$ for some $u \in (0,1)^N$ and $v\in(0,1)^M$. The canonical application of this setting is online advertising, where the probability that a user clicks on an ad depends on both the position at which the ad is displayed and its relevance to the user \cite{pmlr-v54-katariya17a}. Other motivations come from two-sided markets, where matching occurs between offers and demands, e.g. in online labor markets, $u_i$ may represent the utility for a user seeking a solution to a project and $v_j$ may represent the expertise of a project solver.

On the other hand, monopartite graphs  have $2N$ vertices and their rank-1 adjacency matrix $W$ can be written as $W=uu^\top$ for some $u\in(0,1)^{2N}$. This setting models collaborative activities that arise in teamwork, online gaming, and online labor platforms \cite{teamformation}. For instance, some online gaming apps match players together (e.g. Go, competitive quizzes, and drawings) but players will participate only if they both want to. In a simple model, player $i$ decides to participate with probability $u_i$, and matched players $i$ and $j$ participate in a game with probability $u_iu_j$. The expected number of played games from a proposed matching $m$ is then $\sum_{(i,j)\in m} u_i u_j$. The revenue of such apps typically comes from ads displayed during games, so the more games played the better. In these examples, the app will match (say, everyday) as many pairs of players as possible and not just one (as in the bipartite example).  

We will consider the aforementioned two variants of these sequential choices of matchings: either the matching has to be  ``minimal'', i.e., it has to be a single pair of vertices, or it has to be ``maximal'', i.e., a choice of $N$ distinct pairs. We will refer the former to as \emph{pair selection}  and the latter as \emph{matching selection} problem. As standard in multi-armed bandits, we shall investigate both the regret minimization over an arbitrary given time horizon and the pure exploration in a PAC learning setting.

\subsection{Related work}

The matching problems  defined above are special classes of combinatorial bandit problems with semi-bandit feedback \cite{cesa2012combinatorial} with many  recent improvements for regret minimization \cite{combes2015combinatorial,cuvelier2021statistically,degenne,perrault2021statistical,wang2021thompson} as well as pure exploration \cite{garivier2016optimal1,NIPS2014_5433}. The  combinatorial structure is quite clear, as the cardinality of the set of matchings is equal to $(2N)!/(2^N N!)\sim \sqrt{2}\big(2N/e\big)^N$.

Off-the-shelf combinatorial bandits algorithms would incur a regret scaling as $\tilde{O}(N^2\log(T)/\Delta_{\mathrm{min}})$, where $\Delta_{\mathrm{min}}$ denotes the expected reward gap between an optimal matching and the best sub-optimal matching. This has been recently improved, but only in the aforementioned bipartite case, where the rank-1 structure has been leveraged in the line of work of \emph{stochastic rank-1 bandits} \cite{pmlr-v54-katariya17a, bernoullirank1b, trinh2019solving1}, yet either with sub-optimal parameter dependencies or with asymptotic performances. This quadratic dependency would also appear in pure exploration, as standard algorithms would require in the bipartite case  $O(NM\log(1/\delta))$ iterations to find the best pair with probability at least $1-\delta$ \cite{garivier2016optimal1}.

The classical matching problem has strong connections with  ranking/sorting; it is obviously the same with their sequential variants  \cite{pmlr-v70-zoghi17a,pmlr-v117-rejwan20a} even though they do not directly handle the  bandit feedback.

\subsection{Organization of the paper and our contributions}

The remaining of the paper is divided in four main parts. First, we formally introduce the general model in Section~\ref{SE:Model}. Then we investigate the pair selection problem (both for regret minimization and pure exploration) in Section~\ref{SE:Pair_Selection}, and afterwards the matching selection problem (again, for regret and pure exploration) in Section~\ref{SE:Matching_Selection}. Finally, we present numerical results in Section~\ref{SE:Simul}. They validate the tightness of our results and demonstrate competitiveness and performance gains obtained by our proposed algorithms over some state-of-the-art baseline algorithms. 

Our contributions can be summarized as follows:
\begin{itemize}
    \item[i)] For the pair selection problem in the bipartite case, we introduce a new algorithm, called \SRankElim, in Section \ref{SE:RegretPair} with an optimal (up to a multiplicative constant) regret bound: perhaps interestingly, the algorithm eliminates sub-optimal rows and columns on different timescales. This result is extended to the monopartite case in the same Section \ref{SE:RegretPair}.
     
     For pure exploration, we simply adapt \SRankElim; it  still leverages the rank-1 structure to find the optimal pair  with a linear (instead of quadratic) sampling complexity in $O((N+M)\log(1/\delta))$. 
     \item[ii)] The monopartite case, still with pair sampling, is investigated in Section \ref{SE:MonopartitePair}; we transform the above algorithm into \RMono, that can handle both regret minimization and pure exploration. It is also extended for best matching identification, with again optimal bound (up to a multiplicative constant).
    \item[iii)] Section \ref{SE:ExploMatching} is dedicated to  pure exploration with matching sampling; a new algorithm is developed with optimal sample complexity for non-degenerate ranges of parameters (i.e., it equals the new lower bounds proved up to multiplicative constants).
    
    \item[iv)] Finally, regret minimization in the matching selection problem is investigated in Section \ref{SE:REgretMatching}; we introduce a new \MDC\ algorithm with a linear (instead of quadratic) dependency in $N$ since its regret  scales as  $\tilde{O}(N\log(T)/\Delta_{\mathrm{min}})$.  
    
    Roughly speaking, this algorithm relies on a divide and conquer type of approach. 
\end{itemize}

\section{Objectives and problem statement}
\label{SE:Model}

\paragraph{Noise model.} We assume the noisy observation $X_t$ of $W$ is generated as follow: for any $(i,j,t), X_{i,j,t} = W_{i,j} + \varepsilon_{i,j,t}$ where $\varepsilon_{i,j,t}$ are independent, zero-mean, sub-Gaussian random variables.

\paragraph{Optimal matching.} The objective is to find a matching $m$, either minimal or maximal depending on the setting, that maximizes the expected reward $\mathds{E}[\sum_{(i,j)\in m}X_{i,j,t}] = \sum_{(i,j)\in m} W_{i,j}$. It turns out that in both the bipartite case and the monopartite one, under the rank-1 assumption, the optimal matching is the one that pairs better items together. More formally and without loss of generality, for the bipartite case ($W=uv^\top$), we assume that $u_1 \geq \cdots \geq u_N$ and $v_1 \geq \cdots \geq v_M$. The optimal matching is the one that associates $(u_1, v_1)$, then $(u_2, v_2)$, and so on\footnote{This is a direct consequence of the \emph{rearrangement inequality}.}. Similarly, for the monopartite case ($W=uu^\top$), we assume that $u_1\geq u_1\geq \cdots \geq u_{2N}$ and the optimal matching associates any odd index with its successor, i.e. $(u_1, u_2)$ then $(u_3, u_4)$ and so on. In both cases finding the optimal matching boils down to finding the order of the entries of $u$ and $v$.

\paragraph{Pure exploration.} A first objective the agent can aim for is to identify the best matching with \emph{high probability} and \emph{as fast as possible}. Formally, given a confidence level $0<\delta<1/2$, the agent seeks to minimize the worst-case number of samples $\tau_\delta$ needed for the algorithm to finish and return the optimal matching with probability at least $1-\delta$.

\paragraph{Regret minimization.} Another objective for the agent is to find the best matching while \emph{playing sub-optimally as few times as possible} in the process. Formally, her goal is to minimize the regret, i.e., the difference between the cumulative reward of the oracle (that knows the best pair or the best matching) and her cumulative reward. Denoting by $\mathcal{M}$ the set of matching considered -- e.g. minimal matchings for \emph{pair selection} or maximal matchings for \emph{matching selection} -- the regret after $T$ steps is defined as:
\begin{align}
    R(T) = T \max_{m\in\mathcal{M}} \sum_{(i,j)\in m} W_{i,j} - \sum_{t=1}^T \sum_{(i,j)\in m_t} W_{i,j}\,.
\end{align}
\paragraph{Universal vs.\ parameter dependent constants.} In order to avoid cumbersomeness, we shall use the notations $c_\mathfrak{u}$ to denote some universal constant and $c_\mathfrak{p}$  to denote constants (w.r.t.\ $T$) but that can depend on other problem parameters. They might change from one statement to another, but they are always defined explicitly in the proofs.

\section{Pair selection problem}
\label{SE:Pair_Selection}

In this section we consider the pair selection problem. Even though playing one pair $(i,j)$ of items at each time step may seem very similar to dueling bandits \cite{yue2012duelingbandit} in the monopartite case, the reward information structure is very different. In dueling bandits the information is \emph{competitive}, one observes which $i$ or $j$ is best (in expectation). Here, the information is \emph{collaborative}, the higher the parameters of both $i$ and $j$, the higher the observation (in expectation). Thus, in our case, playing the pair $(i,j)$ does not provide information on the relative order of $i$ and $j$. Instead, to get information about the relative order of $i$ and $j$, it is necessary to use a third item $j'$ as a point of comparison and play both $(i, j')$ and $(j,j')$. We will refer to this as comparing $i$ with $j$ \emph{against} $j'$. A crucial idea, that is key to several of the algorithms presented in the paper, is that the fastest way to compare two items $i$ and $j$ is to compare them against the item $j'$ with the highest possible parameter value. It turns out this last remark also holds in the bipartite case.

\subsection{Bipartite case}
\label{SE:RegretPair}

 
 
 

As the bipartite case has already been studied and might be simpler to grasp, we start with it and then extend the results to the monopartite case. The fastest way to compare row items is to compare them against the best column, and reciprocally for columns. Similarly to {\sc Rank1Elim} \cite{pmlr-v54-katariya17a}, the algorithm maintains a list of \emph{active} rows (resp. columns) that are, with high probability, \emph{non-provably dominated} as defined by confidence sets computed from the samples. As {\sc Rank1Elim}, \SRankElim\ performs an Explore Then Commit (ETC) strategy, playing all active rows against randomly chosen active columns to collect samples and update the confidence sets. Then it uses a similar ETC strategy on columns. The main difference with {\sc Rank1Elim} resides in \SRankElim\ eliminating row and columns at different timescales. \SRankElim\ implements an ETC policy with horizon $T$ for rows. Simultaneously, it runs an ETC policy for columns, but over shorter time windows $w$ (referred to as ``blocks'') between steps $T_{w-1}$ and $T_w-1$ where $T_w:=T_{w-1} +2^{2^w}$. 
Within a block, columns  are only \emph{temporarily} eliminated. They are reinstated as active  at the beginning of the next block, when a new instance of the ETC is run in the new horizon. 
The detailed pseudo-code is given in Appendix \ref{SRankElimAppendix}.

\begin{algosketch}
\caption{\SRankElim}
\begin{algorithmic}\label{PairElimcode}
\STATE Set target precision to $\delta$
\FOR{$t=0...$}
    \STATE Identify active rows and columns
    \STATE Sample all active columns against a random active row
    \STATE Sample all active rows against a random active column
    \IF{$t>\sum_{s=0}^{w}2^{2^s}$} 
        \STATE Reset samples for columns
        \STATE $w=w+1$
    \ENDIF
    \STATE Update confidence sets on rows and columns
    \IF{Optimal pair detected with confidence  $ \geq 1-\delta$}
        \STATE Output optimal pair
    \ENDIF
\ENDFOR
\end{algorithmic}
\end{algosketch}


The key intuition is that the algorithm is more aggressive in the elimination of columns (i.e. with lower confidence) as they are only temporarily eliminated. Thus, sub-optimal rows will be eliminated after $\log(T)$ samples while the number of samples of sub-optimal columns  only increases logarithmically in the current number of samples. This implies that pairs $(i,j)$ that are ``doubly-suboptimal'', i.e. both $i$ and $j$ are sub-optimal, are eliminated after $\log(\log(T))$ samples. On the other hand, pairs that are only suboptimal, but not doubly, are eliminated after $\log(T)$ samples.

 It is noteworthy that this cannot be achieved with a standard Explore Then Commit (ETC) independent on rows and columns, as the number of samples of sub-optimal decisions would then scale linearly (not logarithmically) with the number of samples -- before  elimination. This is the reason why the algorithm \RankElim\ is sub-optimal.

The complexity of bandit problems is characterized by the \textsl{gaps}, i.e., the difference in expectation between basic item performances. In this case, we need to differentiate gaps on rows and columns, namely,
$\Delta_i^U=\max_{i'\in [N]} u_{i'}-u_i$ and $\Delta_j^V=\max_{j'\in [M]} v_{j'}-v_j$ that appear both in the regret and in the sample complexity.

\begin{restatable}{theorem}{PairElimRegret}\label{Regret2S}

For any time horizon $T > 0$, the expected regret of \SRankElim\ with $\delta = (1/T)$ is upper bounded as,
$$
\E[R(T)] \leq c_{\mathfrak{u}} A(u,v) \log(T) + c_{\mathfrak{p}}\log(\log(T))
$$
where 
$$
A(u,v)=\sum_{i\in [N]: \Delta_i^U > 0} \frac{1}{v_1 \Delta_i^U} +  \sum_{j\in [M]: \Delta_j^V > 0} \frac{1}{u_1 \Delta_j^V}.
$$
\end{restatable}

The proof of  Theorem \ref{Regret2S} is available in Appendix \ref{SRankElimAppendix}.

\begin{restatable}{theorem}{PairElimExplore}\label{UBIB-L} For any $\delta \in (0,1)$,  \SRankElim\ outputs the best pair with probability at least $1-\delta$ at stage $\tau_\delta$ s.t.
  \begin{equation*}
 \E[\tau_{\delta}] \leq c_\mathfrak{u} A(u,v)\log(1/\delta) + c_\mathfrak{p} \log\log(1/\delta)
\label{equ:etau}
 \end{equation*}
 where 
 $$
A(u,v)= \sum_{i\in [N]: \Delta_i^U > 0}\frac{1}{(v_1 \Delta^U_i)^2} + \sum_{j\in [M]: \Delta^V_j > 0}\frac{1}{(u_1 \Delta^V_j)^2}
$$
which is tight up to a multiplicative constant.
\end{restatable}

Proofs of the upper bound of Theorem \ref{UBIB-L} is in Appendix \ref{SRankElimAppendix}. The lower was proven in previous work \cite{pmlr-v54-katariya17a}.
 

The improvement compared to \RankElim\ is visible, as the leading term of the regret scales as the inverse of the parameters mean for \RankElim, while it only scales as the inverse of the best parameter, for \SRankElim.

\subsection{Monopartite case}\label{SE:MonopartitePair}

We introduce a new algorithm, \RMono, that generalizes the  main ideas of \SRankElim. Instead of working on monopartite graph (of size $2N$), it first duplicates items and create a bipartite graph with $\cU=\cV=[2N]$. Then, as in the previous section, an elimination policy is run over rows  with horizon $T$ and over columns  by blocks.

The major difference between the mono and bipartite case is that, in the former, two items of $\cU$ and $\cV$ are optimal (instead of only one, because of the initial duplication). As a consequence, active pairs are tracked instead of active rows and columns. Pairs containing item $i$ are all eliminated after they have been deemed smaller than two other distinct items. If $i$ is deemed smaller than another item $j$, all pairs containing item $i$ are eliminated except $(i,j)$. If entry $(i,j)$ is eliminated as a consequence of row $i$'s sub-optimality, entry $(j,i)$ is also eliminated.

Note that the fastest way to compare item $1$ with item $i>2$ is to compare them against item $2$ and the fastest way to compare item $2$ with item $i>2$ is to compare them against item $1$. Similarly, it is harder to identify the second best item than the best item, as simple computations yield
\begin{equation*}
u_1 \Delta_{2,i} \leq u_2 \Delta_{1,i}.
\end{equation*}
where $\Delta_{i,j}=u_i-u_j$.


\begin{algosketch}
\caption{\RMono}
\begin{algorithmic}\label{PairElimMonocode}
\STATE Set target precision to $\delta$
\STATE Initiate rows $\cU=[2N]$ and columns $\cV=[2N]$
\FOR{$t=0,1,\ldots$}
    \STATE Identify active pairs
    \STATE Sample all active pairs
    \IF{$t>\sum_{s=0}^{w}2^{2^s}$} 
        \STATE Reset samples for columns
        \STATE $w=w+1$
    \ENDIF
    \STATE Update confidence sets on rows and columns
    \IF{Optimal pair detected with confidence  $ \geq 1-\delta$}
        \STATE Output optimal pair
    \ENDIF
\ENDFOR
\end{algorithmic}
\end{algosketch}


Apart from the algorithm itself, another difference with the bipartite case is the way to measure the gaps, but the guarantees are very similar to the bipartite case.

\begin{restatable}{theorem}{PairElimMonoregret}\label{monopartite}
The expected regret of \RMono\ satisfies, for any time horizon $T > 0$,
\begin{equation}\label{equ:Runi}
\E[R(T)] \leq c_\mathfrak{u}A(u)\log(T) + c_\mathfrak{p}\log\log(T)
\end{equation}
where 
$$A(u)=\sum_{i\in\{3,\ldots,2N\}:\Delta_{2,i}>0} \frac{1}{u_1 \Delta_{2,i}}.
$$
\end{restatable}

The proof is provided in Appendix \ref{RMonoAppendix}.

\begin{restatable}{theorem}{PairElimMonoexplore}\label{TH:mono_id} For any $\delta \in (0,1)$, the sample complexity of the \RMono\ algorithm satisfies
$$
\tau_{\delta} \leq c_\mathfrak{u}A(u)\log(1/\delta)+c_\mathfrak{p}\log\log(1/\delta)
$$
with probability at least $1-\delta$, where 
$$
A(u)=\sum_{i\in [2N]: \Delta_i > 0} \frac{1}{(u_1 \Delta_{2,i})^2} 
$$
which is tight up to a multiplicative constant.
\end{restatable}
The proof is provided in Appendix \ref{RMonoAppendix}.
\paragraph{Towards maximal matchings.} The \emph{matching selection} setting can be seen as a constrained versions of \emph{pair selection} where the agent has the constraint that $N$ consecutively sampled pairs should form a maximal matching instead of being chosen freely. Hence, before diving in this setting, we can wonder what would be the sample complexity to identify the best maximal matching, but by freely choosing the pairs, which is arguably a simpler problem than \emph{matching selection}. This can be done in two steps: {\bf 1)} identify the two best items using \RMono, {\bf 2)} sample all unranked items against them until the full best matching is identified. We refer to this two-step algorithm as \IBLearning. There again, the sample complexity of the algorithm is optimal up to a multiplicative constant, proofs are deferred to Appendix~\ref{IBLearningAppendix}.



\begin{restatable}{theorem}{PairSelect}\label{TH:UpperBoundIBLearning} For any $\delta \in (0,1)$, the sample complexity of the \IBLearning\ algorithm satisfies
$$
\tau_{\delta} \leq c_\mathfrak{u}A(u)\log(1/\delta)+c_\mathfrak{p}\log\log(1/\delta)
$$
with probability at least $1-\delta$, where 
$$
A(u)=\sum_{i\in [2N]: \Delta_i > 0} \frac{1}{(u_1 \Delta_{i})^2}\,,
$$
$$
\text{with}\ \Delta_{2,i} = u_{2i}-u_{2i+1}~~\text{and} ~~  \Delta_{2i-1} = u_{2i-2}-u_{2i-1}
$$
which is tight up to a multiplicative constant.
\end{restatable}

Surprisingly, identifying the full order of the items rather than simply the top two ones does not require $N$ times more samples. Rather the degradation appears in the gap, that are not anymore the gap to the second-best item, but rather the gap between consecutive matched pairs. 

In the following section, we investigate if similar guarantees hold in the \emph{matching selection} setting or if the constraints on the choice of consecutive pairs have a stronger impact.

\section{Matching selection problem}
\label{SE:Matching_Selection}

In this section, we present our results for the matching selection problem, where recall at each iteration step a maximal matching of items needs to be selected. We first consider the objective of pure exploration, and then consider the regret minimization objective. In this section, pair $i$ refers to the $i^{\rm th}$ pair of consecutive items $(u_{2i-1},u_{2i})$, the pair with the $i^{\rm th}$ highest expected reward in the optimal matching.

Contrarily to the pair selection setting, where the same algorithm has tight guarantees for both regret minimization and pure exploration, the matching selection setting requires different algorithms. This can be understood from a simple example where the smallest gap between consecutive items $i, i+1$ appears at the bottom of the ranking (between low quality items). From a pure exploration point of view, the fastest way to rank them is to compare them against the top ranked item. However, this is sub-optimal from a regret point of view as misranking $i$ and $i+1$ incurs a low regret, while playing $(1,i)$ or $(1, i+1)$ incurs a high regret. The pure exploration objective and the regret minimization objective are treated separately in the following as they require the use of different algorithms.

\subsection{Pure exploration for matching}
\label{SE:ExploMatching}


The algorithm for matching selection with an objective of pure exploration, referred to as \MatchingId, is based on this idea of comparing items with the top ones to rank them quickly (remember finding an optimal matching amounts to finding an order over the items). It makes use of two non-exclusive sets of items: $\cS\subseteq[2N]$ is the set of un-ranked items (their exact rank is unknown) and $\cB$ the set of items that are still potentially amongst the $|\cS|$ best items. The algorithm proceeds through iterations, in each iteration sampling matchings from $\cB\cup\cS$ such that all distinct pairs of items are sampled once. Items from $\cS$ are ranked by using the samples collected from matches with items in $\cB$. This procedure continues until the rank of all items is known.
\begin{algosketch}[ht]\label{MId}
\caption{\MatchingId}
\begin{algorithmic}
\STATE $\cS=[2N]$
\WHILE{$\cS\neq \emptyset$}
    \STATE Compute the set $\cS$ of unranked items
    \STATE Compute the set $\cB$ of candidate $|\cS|$ best items
    \STATE Sample each item in $\cB\cup \cS$ once against each other item in that set
    \STATE Update confidence intervals for the items in $\cS$ using observed outcomes of matches with items in $\cB$
\ENDWHILE
\end{algorithmic}
\end{algosketch}

We introduce the following notations
$$
\mu_{[2N]\setminus \{2k,2k+1\}}= \frac{1}{2N}\sum_{i\in[2N]\setminus\{2k,2k+1\}}u_i.
$$
Let $s$ and $h$ denote the indices of the smallest and the second smallest gap, i.e. $s =\arg\min_{k \in [2,N-1]}\Delta_{2k,2k+1}$ and $h=\arg\min_{k \in [N-1]\setminus \{s\} }\Delta_{2k,2k+1}\mu_{[2N]\setminus \{2k,2k+1\}}$. We also define
$$
\gamma_{\mathrm{min}} = \min_{k\in [N-1]} \{\mu_{[2N]\setminus\{2k,2k+1\}}\Delta_{2k,2k+1}\}.
$$
To simplify the exposition of the result, we assume that the smallest gap is not between the two best pairs (general version of the result is given in Appendix \ref{MatchingIdAppendix}).

\begin{restatable}{theorem}{Matchingidmaxbound}[upper bound]\label{matchingidmaxbound}
For  any $\delta > 0$, the sample complexity of the \MatchingId\ algorithm satisfies
$$
\tau_{\delta}\leq c_\mathfrak{u}\frac{1}{\gamma_{\mathrm{min}}^2}\log(1/\delta) +c_\mathfrak{p}
$$
with probability at least $1-\delta$. Moreover, by denoting, 
$$
\alpha:=\min\Big\{\frac{1}{2}\frac{(u_1+u_2)\Delta_{2s,2s+1}}{\mu_{[2N]\setminus\{2h,2h+1\}}\Delta_{2h,2h+1}}, 1\Big\},
$$
the following holds with probability at least $1-\delta$
$$
\tau_{\delta} \leq c_\mathfrak{u}\frac{1}{(1-\alpha)^2(u_1^2+u_2^2)\Delta_{2s,2s+1}^2}\log(1/\delta) + c_\mathfrak{p} .
$$
\end{restatable}

The proof of these upper-bounds is deferred to Appendix~\ref{MatchingIdAppendix}. These upper bounds are tight, up to multiplicative factors, for some interesting regimes of parameters, as stated below.


\begin{restatable}{theorem}{LowerBoundMatching}\label{TH:LowerBoundMatching}
Assume that stochastic rewards of item pairs have Gaussian distribution with unit variance. Then, for any $\delta$-PAC algorithm, we have
\begin{equation*}
\E[\tau_{\delta}]\geq c_\mathfrak{u}\hspace{-0.5cm} \sum_{i\in [2N]: \Delta_i > 0}\frac{1}{\sum_{j=1}^{2N} u_j^2}\frac{1}{\Delta_i^2}\log(1/\delta)
\end{equation*}
and
\begin{equation*}
\E[\tau_{\delta}]\geq c_\mathfrak{u} \frac{1}{u_1^2+u_2^2}\frac{1}{\Delta_{2s,2s+1}^2}\log(1/\delta).
\end{equation*}
\end{restatable}

In particular, if all gaps are equal, the first lower bound of Theorem \ref{TH:LowerBoundMatching} matches the first upper bound of Theorem \ref{matchingidmaxbound} up to a multiplicative constant. On the other hand, in the opposite regime where one gap is substantively  smaller than all others, then it is the second bounds that are equivalent.



These results with matching matching should be put into perspective with their pendant, Theorem \ref{TH:UpperBoundIBLearning}, for pair selection. In this case, the sample complexity of the latter is $N$ times bigger than the one of the former. This might seem surprising at first sight as pair selection is an ``easier'' problem (without constraints). The reason is that with matching selection, the algorithm gets to observe $N$ pairs at each iteration (and not just one). As a consequence, the overall number of pairs evaluation $X_{i,j,s}$ are actually of the same order.

This is quite surprising as selection matchings is much more constrained than selecting batches of $N$ arbitrary pairs (possibly with repetitions and/or single items sampled more than just once  in a batch). The main consequence is that the matching identification problem is \textsl{as difficult} with minimal than maximal matchings selection (and therefore in any intermediate case).

\subsection{Regret minimization}
\label{SE:REgretMatching}
We first describe and analyze a simplified matching divide and conquer algorithm, which is correct for problem instances satisfying a condition on model parameters introduced shortly. This simplified algorithm allows us to convey the key idea that underlies the design of a more complicated algorithm. Having described this simplified algorithm and shown the regret upper bound, we will remove the aforementioned condition and show a regret upper bound that holds for a 
matching divide and conquer algorithm.  In the following, we assume there is a unique optimal matching $m^*$.

\subsubsection{\SMDCname}

We consider problem instances under the following assumption on model parameters:

\begin{hyp}
\label{hyp:even} The model parameters $u_1, \ldots, u_{2N}$ are assumed to satisfy $u_{2i-1} = u_{2i}$, for all $i \in [N]$.
\end{hyp}

Under Assumption~\ref{hyp:even}, in the optimal matching every pair of matched items have equal parameter values. This assumption avoids some complications that arise due to uneven clusters in the divide and conquer procedure. 

The \SMDCname\ (\SMDC) successively partitions items into an ordered sequence of clusters (ranked clusters), such that all items in a  cluster have a higher rank than all items in any lower-ranked cluster with a high probability. Items in a cluster $\cS$ of size $|\cS|=2K$ are matched according to a round-robin tournament, which runs over $2K-1$ iterations. 

A cluster is split into two sub-clusters as soon as the upper bound for the reward of each item in one of the sub-clusters is smaller than the lower bound for the reward of each item in the other sub-cluster. Assumption~\ref{hyp:even} guarantees that, with high probability, at each iteration step, all clusters contain an \emph{even} number of items, so it is always possible to match all items within each cluster.

Functions {\tt sample\_matching} and {\tt conf\_bound} of Algorithm \ref{SMDCcode} are detailed in Appendix \ref{SMDCAppendix}. At the high level, {\tt sample\_matching} samples matchings that ensure that each $|\cS|-1$ iterations, any given item in $\cS$ has been matched once with any other item in $\cS$. {\tt conf\_bound} builds confidence intervals for the total reward of each item per match with items in the same cluster or in lower ranked cluster.

\begin{algorithm}
\caption{\SMDC}
\begin{algorithmic}\label{SMDCcode}
\STATE {\bfseries Input:} set of items $[2N]$ and horizon $T$
\STATE $t=0, C=X=\tilde{C}=\tilde{X}=[0]^{2N\times 2N}, \mathfrak{S}=\{[2N]\}$
\FOR{$t=1\ldots T$}
    \STATE $m_t \gets$ {\tt sample\_matching}$(\mathfrak{S},t)$
    \FOR{$(i,j)\in m_t$}
        \STATE $\tilde{X}(i,j) \gets \tilde{X}(i,j) + X_{i,j,t}$
        \STATE $\tilde{C}(i,j) \gets \tilde{C}(i,j) + 1$
    \ENDFOR
    \FOR{$\cS \in \mathfrak{S}$} 
        \IF{$\exists i \in \cS \text{ s.t. }\sum_{j\in S}\tilde{C}(i,j)=|\cS|-1$}
            \STATE $X([s],:),C([\cS],:) += \tilde{X}([\cS],:),\tilde{C}([\cS],:)$
            \STATE $\tilde{X}([\cS],:),\tilde{C}([\cS],:)=0$
        \ENDIF
    \ENDFOR 
    \STATE $Q_{+},Q_{-}  \gets$ {\tt conf\_bound}$(X,C,T,Q_{+},Q_{-},\mathfrak{S})$
    \FOR{$\cS \in \mathfrak{S}$}
        \STATE Order items in $\cS$ according to $Q_{+}$
        \FOR{$i \in \{2,\dots,|\cS|\}$} 
            \IF{$Q_{+}[i]<Q_{-}[i-1]$} 
            \STATE Split $\cS$ between $i-1$ and $i$
            \ENDIF
        \ENDFOR
    \ENDFOR
\ENDFOR
\end{algorithmic}
\end{algorithm}



The regret of \SMDC\ can be bounded by using the following additional notation 
$$
\Delta_{\mathrm{min}} = \min_{m\in \mathcal{M}: m \neq m^*}  \Big\{\sum_{(i,j)\in m^*}u_iu_j-\sum_{(i,j)\in m}u_iu_j\Big\}
$$ 
and the proof is again deferred to Appendix \ref{SMDCAppendix}.

\begin{restatable}{theorem}{smds}\label{thm:smds} The expected regret of  \SMDC\ satisfies, for any horizon  $T > 0$, 
$$
\E[R(T)] \leq c_\mathfrak{u} \frac{N\log(N)}{\Delta_{\mathrm{min}}}\log(T)+ c_\mathfrak{p}.$$
\end{restatable}


\subsubsection{\MDC\ algorithm}

In general, when Assumption~\ref{hyp:even} does not hold, some of the clusters may have odd sizes. In these odd-size clusters, uniform sampling within a cluster is infeasible, and we need to match items residing in different clusters. In order to deal with these complications, we use a new \MDC\ algorithm. 

\MDC\ is defined as an extension of \SMDC. It uses the same policy for splitting clusters. The \CMSampling\ procedure extends that of the previous algorithm to unevenly split clusters. This procedure ensures that any two mutually un-ranked items are matched similarly to other items, which guarantees that the expected rewards for those items are scaled similarly. At the high level, it defines a list of matchings that respect the desired item sampling proportion, then samples the matchings in this list in a round-robin.

A detailed description of \CMSampling, and the proof of the following result, are given in Appendix \ref{MDCAppendix}. 


\begin{restatable}{theorem}{mds} The expected regret of  \MDC\  is  bounded, for any  horizon $T > 0$, as  
$$
\E[R(T)] \leq c_\mathfrak{u} \frac{N\log(N)}{\Delta_{\mathrm{min}}}\log(T) + c_\mathfrak{p}.
$$
\end{restatable}

\subsubsection{Comparison with an exploration policy}


The \MDC\ algorithm matches high parameter value items together as soon as they are identified in order to exploit this for accruing reward. On the other hand, our algorithms, for the pair sampling problem and the pure exploration matching identification problem, used the detected high parameter value items to \emph{explore} the un-ranked ones. Without a comparison, it is unclear which of the two strategies will lead to a smaller regret in the end. For this reason, we consider an \emph{exploration-first} algorithm that matches identified high parameter value items with other items to speed up the learning of the rank of these other items, and compare it with \MDC. 


We do this under assumption that both algorithms are given as input the two best items, as well as a set of $2(N-1)$ un-ranked items. We chose as a comparison metric the upper bound on the total regret incurred by the two algorithms until the un-ranked items can be partitioned into two or more ranked clusters of items. We denote with $U_I$ the upper bound on the regret $R_I$ for the exploration-first strategy and with $U_D$ the upper bound on the regret $R_D$ for the \MDC\ algorithm. 

The following Lemma~\ref{strategyComparison} states that unless the second best item is sufficiently worse than the best item, the regret of the \MDC\ algorithm is at most of the same order as that of the exploration-first algorithm, and can be arbitrarily smaller depending on the problem parameters. If the ratio between the parameter values of the first and the second best item goes to zero, then the exploration-first algorithm becomes infinitely better than the \MDC\ algorithm. 

\begin{restatable}{lemma}{comparison}\label{strategyComparison}
   If $u_2/u_1>1/2$, then  $U_D/U_I\leq c_{\mathfrak{u}}N$, and it can be arbitrarily close to $0$ depending on the problem parameters. If $u_3/u_2>1/2$, then this ratio is smaller than a constant independent from the parameters of the problem. On the other hand, $\lim_{u_2/u_1 \to 0} R_D/R_I = +\infty$.
\end{restatable}

In summary, the lemma tells  that \MDC\ is essentially as good as the exploration-first strategy for any problem instance such that the parameter value of the second best item is at least a constant factor of the best item.

\section{Numerical results}
\label{SE:Simul}
\begin{figure}[t]
    \centering
    \includegraphics[width=0.35\textwidth]{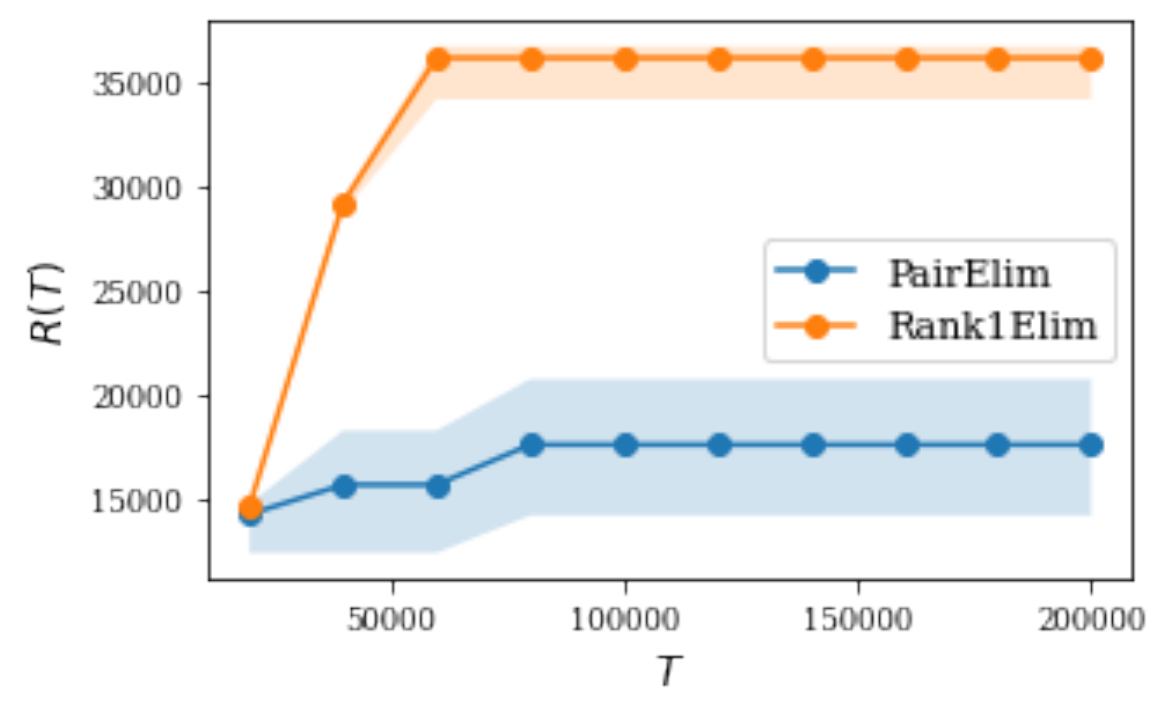}\\
    \includegraphics[width=0.35\textwidth]{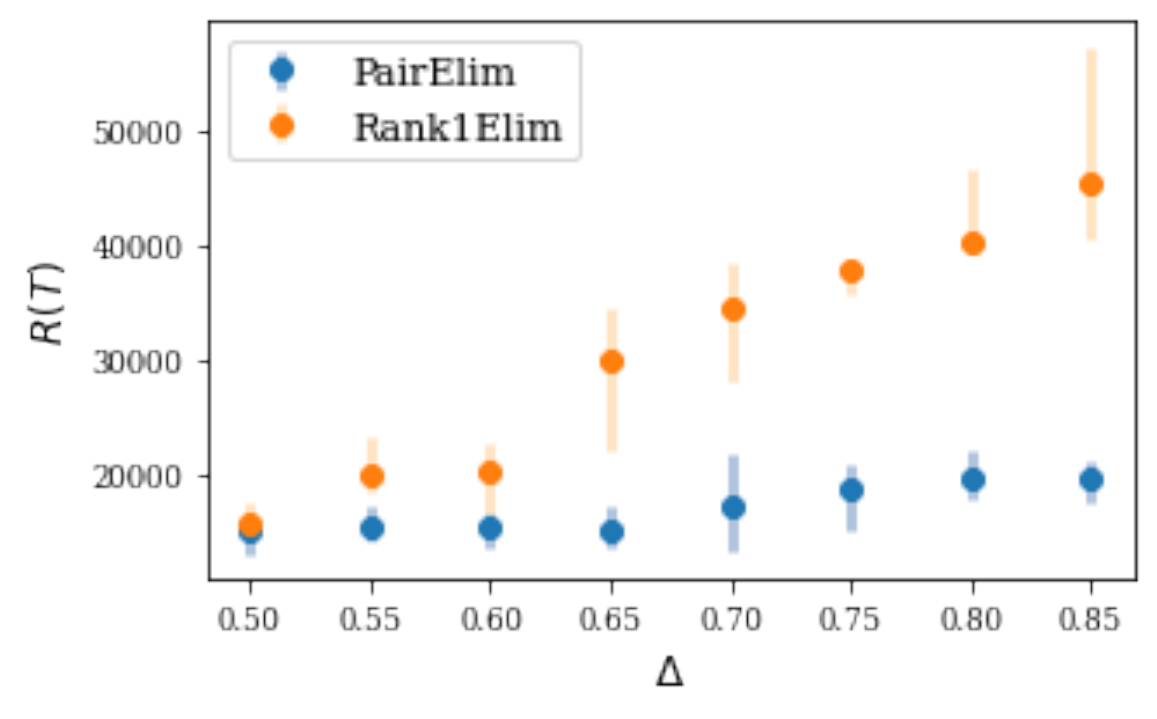}
    \caption{Regret comparison for \SRankElim\ vs \RankElim: (top) regret versus $T$ for fixed $\Delta = 0.75$ and (bottom) regret versus the gap parameter $\Delta$ for fixed $T = 2,000,000$. We used $20$ independent runs. The shaded areas show the range between the 5\% and 95\% percentile.
    }
    \label{fig:comparison}
\end{figure}

In this section we present numerical results, which demonstrate tightness of our theoretical bounds and compare our proposed algorithms with some state-of-the-art baseline algorithms. We first consider the pair selection problem and then the matching selection problem. In summary, our numerical results validate our theoretical results and demonstrate that significant performance gains can be achieved against some previously proposed algorithms.

All the code used for obtaining the results in this section is available from this public Gitlab repository: [anonymized] 

\subsection{Pair selection}

We consider \RankElim\ as a baseline for comparison. As noted in the introduction, \RankElim\ has a regret upper bound that is sub-optimal with respect to the problem parameters, which is in contrast to our algorithm, \SRankElim\ that has optimal regret bound up to a multiplicative constant. We demonstrate that significant performance gains that can be achieved by using \SRankElim\ versus \RankElim\ for some problem instances. 

In all experiments, the variables considered are Bernoulli variables. We consider the bipartite case with $N=M$. Each problem instance is defined by a tuple $(N, u_1, \Delta)$, where $0\leq \Delta\leq u_1 \leq 1$, and assuming that $u_1 = v_1$. 
The row parameter values $u_2, \ldots, u_N$ are defined as sorted values of independent random variables according to uniform distribution on $[0, 2(u_1 - \Delta)]$, where 
$\Delta$ is the expected gap between the value of the best item and the value of any other item.
Note that we have
$
\mu_U = u_1 - (1-1/N)\Delta.
$
For fixed value of parameter $u_1$ and increasing expected gap $\Delta$, we have problem instances with fixed maximum row parameter value and decreasing mean row parameter value $\mu_U$. We similarly define the column parameter values, and all the observations above made for row parameter values hold for column parameter values. According to our regret analysis, \SRankElim\ algorithm will outperform \RankElim\ when $\Delta$ is large for fixed $u_1$.
To confirm this claim, we ran the two algorithms on a set of problem instance with $N=8$ and $u_1=0.9$. The results are shown in Figure~ \ref{fig:comparison}. We have further evaluated the effects of varying $u_1$ which is discussed in Appendix~ \ref{appendix:exp:pair}.

\subsection{Matching selection}

In this section we evaluate the performance of our algorithm for the matching selection problem. Our goal is twofold. We first demonstrate numerical results according to which our proposed algorithm has expected regret that scales proportionally to $N\log(N)/\Delta_{\mathrm{min}}$ for a fixed horizon $T$. We then compare its performance with that of ESCB, which, as discussed in the introduction, has the expected regret bound $O(N^2\log^2(N)/\Delta_{\mathrm{min}})$ for fixed $T$. We demonstrate that our algorithm can achieve significant performance gains over ESCB. In our evaluations. We use our \SMDC\ algorithm, for problem instances such that there is a unique optimum matching with matched items having equal parameter values.

\begin{figure}[t]
    \centering
    \includegraphics[width=0.35\textwidth]{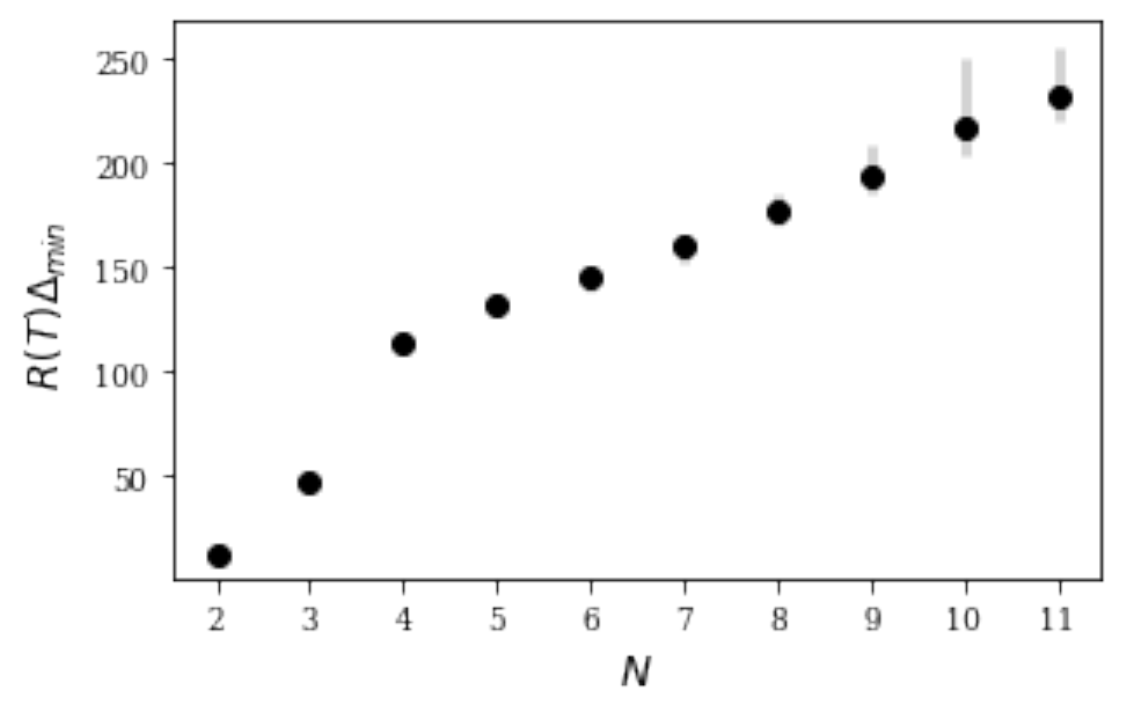}
    \caption{Normalized regret of \SMDC\ versus $N$ for $u_{2i-1} = u_{2i} = (N-i)\tilde{\Delta}$, for $i\in [N]$, 
    with $\Tilde{\Delta}=0.1$ and $T=200,000$.}
    \label{fig:matching_nlogn}
\end{figure}


We first show that the regret of our proposed algorithm scales in the order of $N\log(N)/\Delta_{\mathrm{min}}$.
We consider a set of problem instances, each is defined by a tuple $(N, \tilde{\Delta})$, where $0<(N-1)\tilde{\Delta}\leq 1$ and
$\tilde{\Delta}$ is the gap between parameter values of adjacent matched pairs in the optimum matching, so that $\tilde{\Delta} = \sqrt{\Delta_{\mathrm{min}}}$.
The parameter values are defined by $u_{2i-1}=u_{2i} = (N-i)\tilde{\Delta}$ for $i\in [N]$.
We run our algorithm on problem instances with $\tilde{\Delta}=0.1$.
The results shown in Figure~ \ref{fig:matching_nlogn} suggest that the cumulative regret of \SMDC\ scales as $(1/\Delta_{\mathrm{min}})N\log(N)$ as established in Theorem~\ref{thm:smds}. 

\begin{figure}[t]
    \centering
  \includegraphics[width=0.35\textwidth]{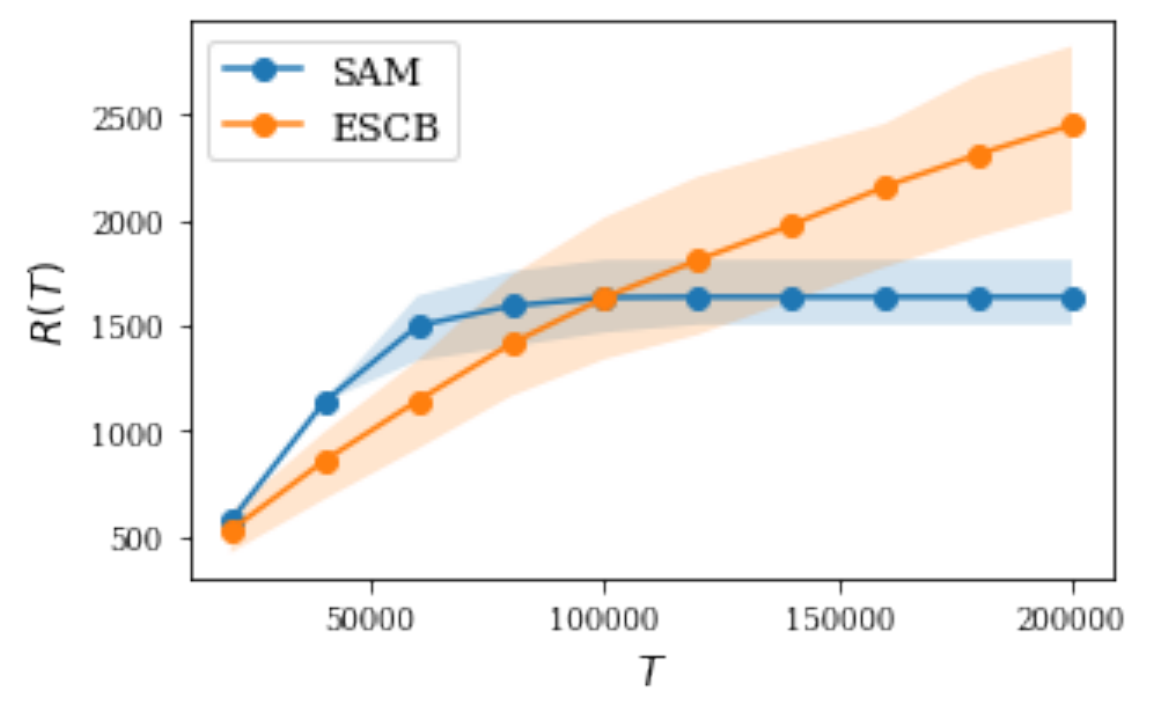}
    \includegraphics[width=0.35\textwidth]{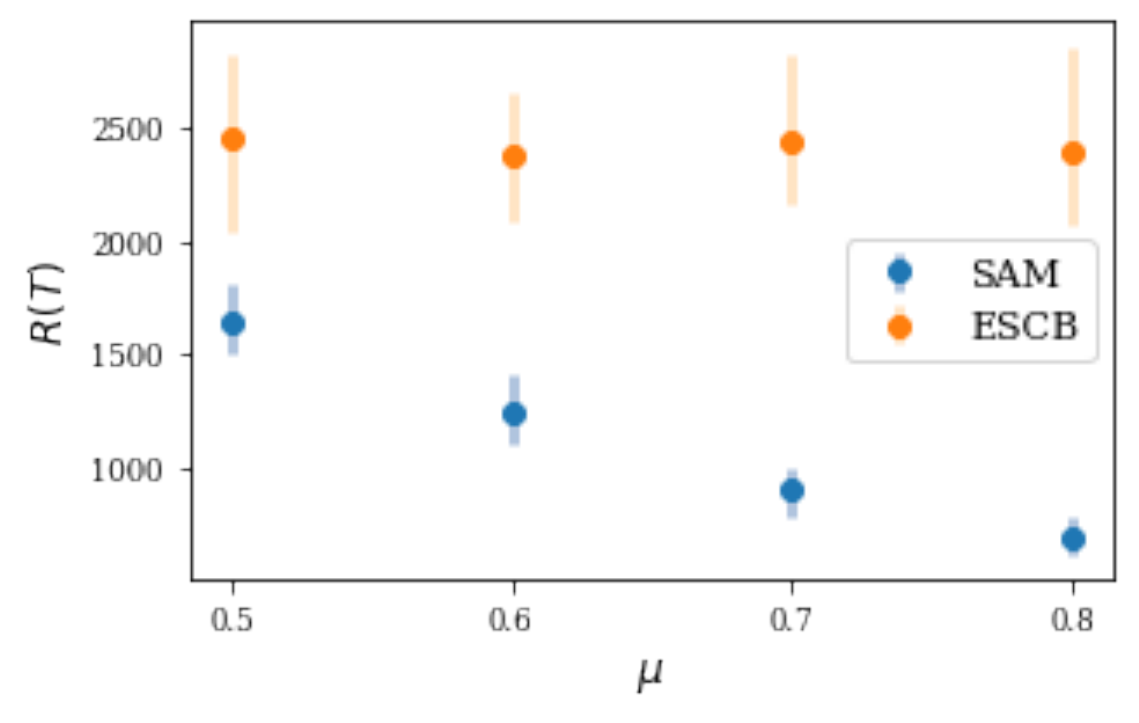}
    \caption{Regret comparison for \SMDC\ (SAM) vs ESCB: (top) regret versus $T$ for fixed $\mu = 0.5$ and (bottom) regret vs $\mu$ for fixed $T=200,000$.}
    \label{fig:matching_comparison}
\end{figure}

We next compare the performance of our algorithm and ESCB. We consider problem instances defined by a tuple $(N, \mu, \tilde{\Delta})$ where $\mu$ is the mean of parameter values and $\tilde{\Delta}$ is the gap between parameter values.
The values of item parameters are set as
$u_{2i-1}=u_{2i} = \mu + (N+1-2i)\tilde{\Delta}/2 \in [0, 1]$  for $i\in [N]$. Such problem instances allow us to vary $\mu$ while keeping other parameters fixed; we fix $N = 4$ and $\tilde{\Delta} = 0.1$. In Figure~\ref{fig:matching_comparison} (top) we show the regret versus the time horizon $T$ for fixed $\mu = 1/2$, which shows that our algorithm outperforms ESCB for large enough values of $T$. We expect our algorithm to perform better than ESCB as we increase $\mu$. The results in Figure~\ref{fig:matching_comparison} (bottom) confirm this claim. We have performed these experiments for a small value of $N$ because of the computation complexity of ESCB. ESCB requires solving an NP-hard problem in each iteration, 
and has overall computation complexity $O(|\mathcal{M}|T)$ 
where $\mathcal{M}$ is the set of all arms. For the matching selection problem, $|\mathcal{M}|$ scales as $\sqrt{2}(2N/e)^N$.  

\section*{Acknowledgements}
 V. Perchet acknowledges support from the ANR under grant number ANR-19-CE23-0026 as well as  the support grant as part of the Investissement d’avenir project, reference ANR-11-LABX-0056-LMH, LabEx LMH, in a joint call with Gaspard Monge Program for optimization, operations research and their interactions with data sciences. M. Vojnovi\'c was supported in part by the Criteo Faculty Research Award.


\bibliographystyle{icml2021}
\bibliography{references}

\begin{thebibliography}{21}
\providecommand{\natexlab}[1]{#1}
\providecommand{\url}[1]{\texttt{#1}}
\expandafter\ifx\csname urlstyle\endcsname\relax
  \providecommand{\doi}[1]{doi: #1}\else
  \providecommand{\doi}{doi: \begingroup \urlstyle{rm}\Url}\fi

\bibitem[Cesa-Bianchi \& Lugosi(2012)Cesa-Bianchi and
  Lugosi]{cesa2012combinatorial}
Cesa-Bianchi, N. and Lugosi, G.
\newblock Combinatorial bandits.
\newblock \emph{Journal of Computer and System Sciences}, 78\penalty0
  (5):\penalty0 1404--1422, 2012.

\bibitem[Chen et~al.(2014)Chen, Lin, King, Lyu, and Chen]{NIPS2014_5433}
Chen, S., Lin, T., King, I., Lyu, M.~R., and Chen, W.
\newblock Combinatorial pure exploration of multi-armed bandits.
\newblock In Ghahramani, Z., Welling, M., Cortes, C., Lawrence, N.~D., and
  Weinberger, K.~Q. (eds.), \emph{Advances in Neural Information Processing
  Systems 27}, pp.\  379--387. 2014.

\bibitem[Combes et~al.(2015)Combes, Shahi, Proutiere,
  et~al.]{combes2015combinatorial}
Combes, R., Shahi, M. S. T.~M., Proutiere, A., et~al.
\newblock Combinatorial bandits revisited.
\newblock In \emph{Advances in Neural Information Processing Systems}, pp.\
  2116--2124, 2015.

\bibitem[Cuvelier et~al.(2021)Cuvelier, Combes, and
  Gourdin]{cuvelier2021statistically}
Cuvelier, T., Combes, R., and Gourdin, E.
\newblock Statistically efficient, polynomial time algorithms for combinatorial
  semi bandits, 2021.

\bibitem[Degenne \& Perchet(2016)Degenne and Perchet]{degenne}
Degenne, R. and Perchet, V.
\newblock Combinatorial semi-bandit with known covariance.
\newblock In \emph{NIPS 2016 (Conference on Neural Information Processing
  Systems)}, 2016.

\bibitem[Garivier \& Kaufmann(2016)Garivier and Kaufmann]{garivier2016optimal1}
Garivier, A. and Kaufmann, E.
\newblock Optimal best arm identification with fixed confidence.
\newblock In \emph{29th Annual Conference on Learning Theory}, volume~49 of
  \emph{Proceedings of Machine Learning Research}, pp.\  998--1027, Columbia
  University, New York, New York, USA, 23--26 Jun 2016.

\bibitem[Johari et~al.(2018)Johari, Kamble, Krishnaswamy, and
  Li]{teamformation}
Johari, R., Kamble, V., Krishnaswamy, A.~K., and Li, H.
\newblock Exploration vs. exploitation in team formation.
\newblock \emph{CoRR}, abs/1809.06937, 2018.
\newblock URL \url{http://arxiv.org/abs/1809.06937}.

\bibitem[Katariya et~al.(2017{\natexlab{a}})Katariya, Kveton, Szepesv\'{a}ri,
  Vernade, and Wen]{bernoullirank1b}
Katariya, S., Kveton, B., Szepesv\'{a}ri, C., Vernade, C., and Wen, Z.
\newblock Bernoulli rank-1 bandits for click feedback.
\newblock IJCAI'17, pp.\  2001–2007. AAAI Press, 2017{\natexlab{a}}.

\bibitem[Katariya et~al.(2017{\natexlab{b}})Katariya, Kveton, Szepesvari,
  Vernade, and Wen]{pmlr-v54-katariya17a}
Katariya, S., Kveton, B., Szepesvari, C., Vernade, C., and Wen, Z.
\newblock {Stochastic Rank-1 Bandits}.
\newblock In \emph{Proceedings of the 20th International Conference on
  Artificial Intelligence and Statistics}, volume~54 of \emph{Proceedings of
  Machine Learning Research}, pp.\  392--401, Fort Lauderdale, FL, USA, 20--22
  Apr 2017{\natexlab{b}}.

\bibitem[Kaufmann et~al.(2016)Kaufmann, Capp{\'e}, and
  Garivier]{kaufmann2016complexity}
Kaufmann, E., Capp{\'e}, O., and Garivier, A.
\newblock On the complexity of best-arm identification in multi-armed bandit
  models.
\newblock \emph{The Journal of Machine Learning Research}, 17\penalty0
  (1):\penalty0 1--42, 2016.

\bibitem[Lov{\'a}sz \& Plummer(2009)Lov{\'a}sz and Plummer]{lovasz2009matching}
Lov{\'a}sz, L. and Plummer, M.~D.
\newblock \emph{Matching theory}, volume 367.
\newblock American Mathematical Soc., 2009.

\bibitem[Mehta(2012)]{mehta2012online}
Mehta, A.
\newblock Online matching and ad allocation.
\newblock \emph{Theoretical Computer Science}, 8\penalty0 (4):\penalty0
  265--368, 2012.

\bibitem[Perrault et~al.(2020)Perrault, Boursier, Perchet, and
  Valko]{perrault2021statistical}
Perrault, P., Boursier, E., Perchet, V., and Valko, M.
\newblock Statistical efficiency of thompson sampling for combinatorial
  semi-bandits.
\newblock In \emph{NeurIPS}, 2020.

\bibitem[Polyanskiy \& Wu(2014)Polyanskiy and Wu]{polyanskiy2014lecture}
Polyanskiy, Y. and Wu, Y.
\newblock Lecture notes on information theory.
\newblock \emph{Lecture Notes for ECE563 (UIUC) and}, 6\penalty0
  (2012-2016):\penalty0 7, 2014.

\bibitem[Rejwan \& Mansour(2020)Rejwan and Mansour]{pmlr-v117-rejwan20a}
Rejwan, I. and Mansour, Y.
\newblock Top-$k$ combinatorial bandits with full-bandit feedback.
\newblock In Kontorovich, A. and Neu, G. (eds.), \emph{Proceedings of the 31st
  International Conference on Algorithmic Learning Theory}, volume 117 of
  \emph{Proceedings of Machine Learning Research}, pp.\  752--776, San Diego,
  California, USA, 08 Feb--11 Feb 2020. PMLR.

\bibitem[Roth et~al.(2004)Roth, S{\"o}nmez, and {\"U}nver]{roth2004kidney}
Roth, A.~E., S{\"o}nmez, T., and {\"U}nver, M.~U.
\newblock Kidney exchange.
\newblock \emph{The Quarterly journal of economics}, 119\penalty0 (2):\penalty0
  457--488, 2004.

\bibitem[Trinh et~al.(2020)Trinh, Kaufmann, Vernade, and
  Combes]{trinh2019solving1}
Trinh, C., Kaufmann, E., Vernade, C., and Combes, R.
\newblock Solving {Bernoulli} rank-one bandits with unimodal {Thompson}
  sampling.
\newblock In \emph{Proceedings of the 31st International Conference on
  Algorithmic Learning Theory}, volume 117 of \emph{Proceedings of Machine
  Learning Research}, pp.\  862--889, San Diego, California, USA, 08 Feb--11
  Feb 2020.

\bibitem[Wang \& Chen(2021)Wang and Chen]{wang2021thompson}
Wang, S. and Chen, W.
\newblock Thompson sampling for combinatorial semi-bandits, 2021.

\bibitem[Wheaton(1990)]{wheaton1990vacancy}
Wheaton, W.~C.
\newblock Vacancy, search, and prices in a housing market matching model.
\newblock \emph{Journal of political Economy}, 98\penalty0 (6):\penalty0
  1270--1292, 1990.

\bibitem[Yue et~al.(2012)Yue, Broder, Kleinberg, and
  Joachims]{yue2012duelingbandit}
Yue, Y., Broder, J., Kleinberg, R., and Joachims, T.
\newblock The k-armed dueling bandits problem.
\newblock \emph{J. Comput. Syst. Sci.}, 78\penalty0 (5):\penalty0 1538–1556,
  September 2012.
\newblock ISSN 0022-0000.
\newblock \doi{10.1016/j.jcss.2011.12.028}.
\newblock URL \url{https://doi.org/10.1016/j.jcss.2011.12.028}.

\bibitem[Zoghi et~al.(2017)Zoghi, Tunys, Ghavamzadeh, Kveton, Szepesvari, and
  Wen]{pmlr-v70-zoghi17a}
Zoghi, M., Tunys, T., Ghavamzadeh, M., Kveton, B., Szepesvari, C., and Wen, Z.
\newblock Online learning to rank in stochastic click models.
\newblock In Precup, D. and Teh, Y.~W. (eds.), \emph{Proceedings of the 34th
  International Conference on Machine Learning}, volume~70 of \emph{Proceedings
  of Machine Learning Research}, pp.\  4199--4208, International Convention
  Centre, Sydney, Australia, 06--11 Aug 2017.

\end{thebibliography}

\newpage
\appendix

\onecolumn

\section{Additional numerical experiments}

In this section, we present some additional experiments for pair selection and matching selection problems.

\subsection{Pair selection}\label{appendix:exp:pair}

A notable improvement of \SRankElim\ compared to \RankElim\ is that the cumulative regret bound is inversely proportional to the \emph{maximums} of row and column item parameter values, instead of the \emph{means} of row and column item parameter values. For our problem instances, the former values correspond to $u_1$ and $v_1$ (with $u_1= v_1$), while the latter values correspond to $\mu_U$ and $\mu_V$ (with $\mu_U=\mu_V$). We thus expect that \SRankElim\ would outperform \RankElim\ for problem instances for which there is a significant gap between the maximum and mean values of the row and column item parameter values. We demonstrate this for problem instances defined as follows.

\begin{figure}[t]
    \centering
    \includegraphics[width=0.35\textwidth]{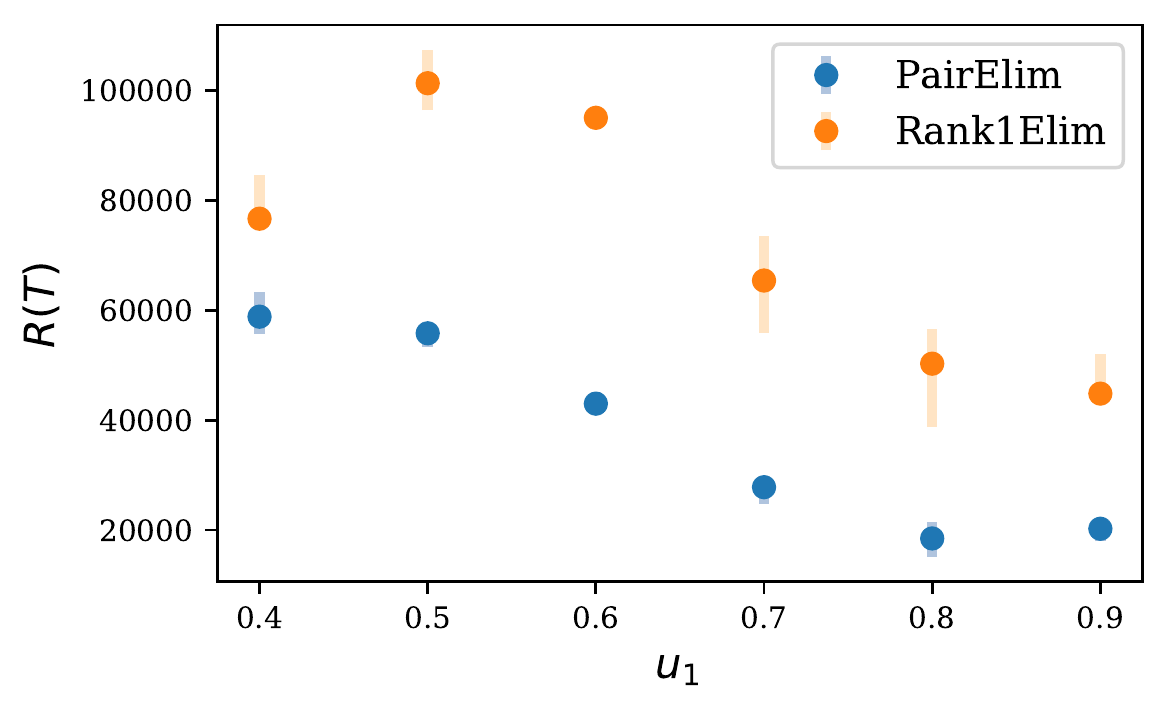}\\
    \hspace*{0.5cm}\includegraphics[width=0.33\textwidth]{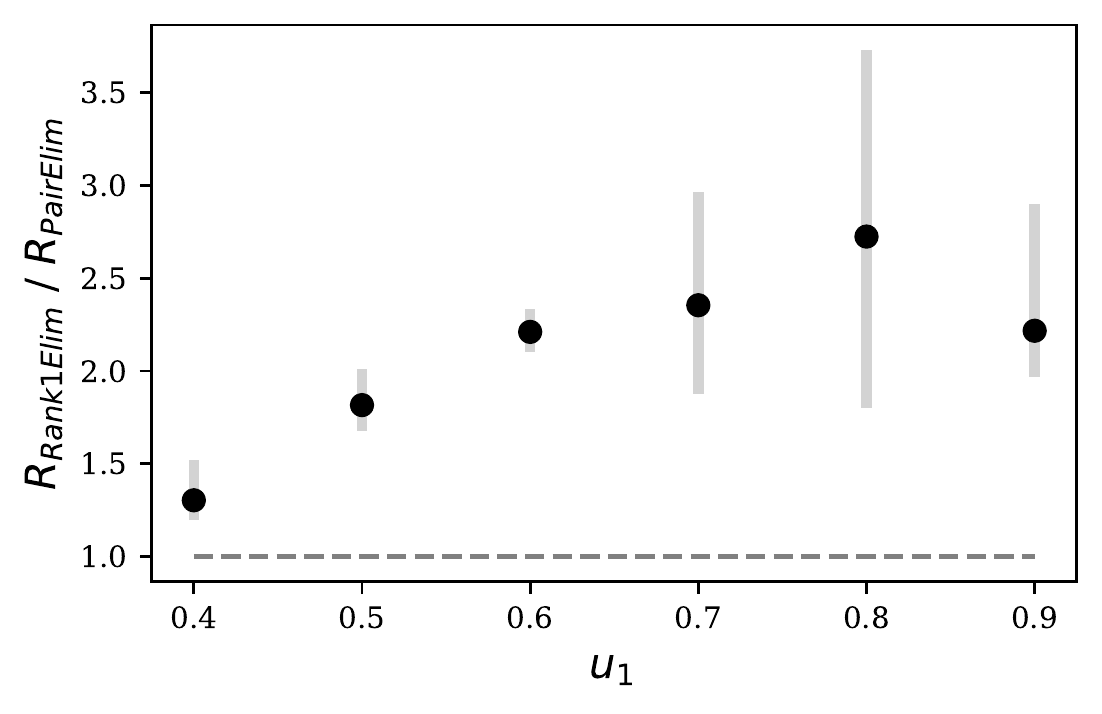}
    \caption{Regret comparison for \SRankElim\ vs \RankElim\ for different values of maximum item parameter value: (top cumulative regrets, and (bottom) ratio of the cumulative regrets. The parameter setting is $N=8$, $\mu_U = 0.2$, and we used $20$ independent runs.}
    \label{fig:u1}
\end{figure}

\begin{figure}[ht]
    \centering
    \includegraphics[width=0.35\textwidth]{delta_regret.pdf}\\
    \hspace*{0.5cm}\includegraphics[width=0.33\textwidth]{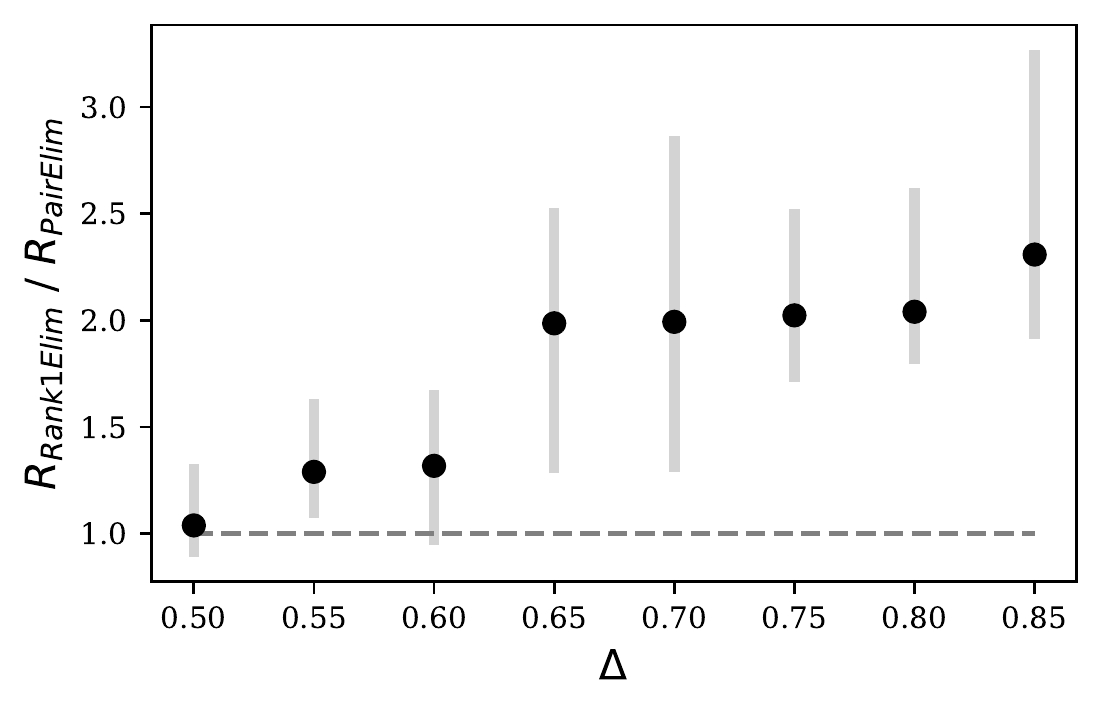}
    \caption{Regret comparison for \SRankElim\ vs \RankElim\ for different values of the gap parameter $\Delta$: (top) cumulative regrets, and (bottom) ratio of the cumulative regrets. The parameter setting is $N=8$, $u_1=0.9$, and we used $20$ independent runs.}
    \label{fig:ratio_delta}
\end{figure}

We consider the bipartite case with $N=M$. Each problem instance is defined by a tuple $(N, u_1, \Delta)$, where $0\leq \Delta\leq u_1 \leq 1$, and assuming that $u_1 = v_1$. 
Other row item parameter values $u_2, \ldots, u_N$ are sorted values of independent random variables according to uniform distribution over $[0, 2(u_1 - \Delta)]$, where 
$\Delta$ is the expected gap between the value of the best item and the value of any other item.
Note that we have
$
\mu_U = u_1 - (1-1/N)\Delta.
$
For fixed value of parameter $u_1$ and increasing expected gap $\Delta$, we have problem instances with fixed maximum row item parameter value and decreasing mean row item parameter value $\mu_U$. We similarly define the column item parameter values, and all the observations above made for row item parameter values hold for column item parameter values.

We first consider problem instances such that the mean values of row and column item parameters are fixed to value $\mu_U = 0.2$, and we vary the value of parameter $u_1$ in $[0.4, 0.9]$. The results are shown in Figure~\ref{fig:u1}. We observe that \SRankElim\ provides a significant performance gain over \RankElim, which for some problem instances is for as much as nearly a $1/3$ reduction of the cumulative regret. We also observe that there is a general trend of \SRankElim\ outperforming \RankElim\ more for larger gaps between the mean and maximum values of item parameter values.

We next consider problem instances that have the maximum row and column item parameter values fixed to value $u_1 = 0.9$ and varying gap parameter $\Delta$ taking values in $[0.5, 0.85]$. Note that with the value of the gap parameter $\Delta$ increasing, the means of row and column item parameter values decrease. As a consequence, the cumulative regret of \RankElim\ should increase. On the other hand, as the maximum values of row and column item parameter values are kept fixed, the cumulative regret of \SRankElim\ should remain roughly the same. These claims are confirmed in Figure~\ref{fig:ratio_delta} (top). In Figure~\ref{fig:ratio_delta} (bottom), we observe a trend of \SRankElim\ outperforming \RankElim\ more for larger values of $\Delta$, and that this can be for a significant amount.

\section{\SRankElim\ algorithm}\label{SRankElimAppendix}

\subsection{Algorithm description and pseudo-code}

In this section, we describe \SRankElim\ algorithm with the pseudo-code provided in Algorithm~\ref{alg:SRankElimAppendix}. 

The key feature of \SRankElim\ is the usage of two different time scales for elimination of rows and columns. The rows are eliminated by using an ETC policy with horizon $T$, and the columns are eliminated by using a similar ETC policy over time windows $w$ (referred to as ``blocks'') between steps $T_{w-1}$ and $T_w-1$. 

The ETC policy relies on maintaining confidence intervals, $Q^U_{+}$, $Q^U_{-}$ and $Q^V_{+}$, $Q^V_{-}$ for rows and columns, respectively. The confidence intervals for the rows are computed from all samples gathered since the beginning of the run, while the confidence intervals for the columns are limited to the samples gathered during the current time window. The target precision of the confidence intervals also varies: for the rows, it is set to optimize the regret over the horizon $T$, for the columns, it is less precise and only optimizes the regret over the current time window. 

A domination mapping $h^U$ is determined from the confidence intervals $Q^U_{+}$ and $Q^U_{-}$. It maps any dominated row to to one of those that dominates it. In the column exploration step, a random row is picked and all non-dominated rows are sampled against $h^U(i)$. $h^V$ is defined and used similarly in the row explorations step.

Note that this domination mapping is not necessary to obtain the proven upper bound on the regret of the algorithm. Sampling against any potentially optimal row or column would provide the same theoretical guarantee. However, we observed experimentally that using the domination mapping gave consistently better results.

When the algorithm is run in the regret minimization mode, the precision of the row confidence intervals is set to the required precision so that they hold until the horizon is reached. The precision of the column confidence intervals is set to the required precision so that they hold until the end of the running time window.

When the algorithm is run in the regret minimization mode, the precision of the row confidence intervals is set to the required precision so that they hold with probability ate least $1-\delta$. Until the best row is detected, the precision of the column confidence intervals is set to the required precision so that they hold until the end of the running time window. Once the best row is detected, it is set so that they hold with probability ate least $1-\delta$.

\begin{algorithm}[hbt!]
\caption{\SRankElim}
\label{alg:SRankElimAppendix}
\begin{algorithmic}
\STATE {\bfseries Input:} set of rows $U$, set of columns $V$, precision $\delta$ (or $\frac{1}{T}$) and $\text{ pure\_explore}$
\STATE $t=0, C=X=C_w=X_w=[0]^{N\times M}$,$\text{ column\_explore}=$True
\FOR{$t...$ }
    \FOR{window $w=0,1,2,\dots$}
        \STATE // columns
        \STATE $Q_{+}^V,Q_{-}^V \leftarrow$ {\tt confidence\_bound}($X_w,C_w,2^{2^w},Q_{+}^V,Q_{-}^V,\text{column\_explore}$)
        \STATE   $h^{V}\leftarrow$ {\tt domination\_map}($Q_{+}^V,Q_{-}^V$)\;
        \STATE  $J \leftarrow \bigcup_{j\in V}\left\{h^{V}(j)\right\}$
        \STATE  // rows
        \STATE $Q_{+}^U,Q_{-}^U \leftarrow$ {\tt confidence\_bound}($X,C,T,Q_{+}^U,Q_{-}^U,\text{ pure\_explore}$) 
        \STATE   $h^{U}\leftarrow$ {\tt domination\_map}($Q_{+}^U,Q_{-}^U$)\;
        \STATE  $I \leftarrow \bigcup_{i\in U}\left\{h^{U}(i)\right\}$
        \STATE   $j\leftarrow h^{V}\left(\text{Unif}(V)\right)$
        \FOR{$i\in I$}
            \STATE $X(i,j) \gets X(i,j) + x_{i,j,t}$
            \STATE $C(i,j) \gets C(i,j) + 1$
            \STATE $s \gets s + 1$, $t \gets t + 1$
        \ENDFOR
        \STATE $i\leftarrow h^{U}\left(\text{Unif}(U)\right)$
        \FOR{$j\in J$}
            \STATE $X_w(i,j) \gets X_w(i,j) + x_{i,j,t}$
            \STATE $C_w(i,j) \gets C_w(i,j) + 1$
            \STATE $s \gets s + 1$, $t \gets t + 1$
        \ENDFOR
        \IF{$s>2^{2^w}$ and $\text{row\_explore}=$True}
            \STATE // Change time window
            \STATE $s=0,C_w=X_w=[0]^{N\times M}$
        \ENDIF
        \IF{$\text{column\_explore}=$False, $|I|$=1 and $\text{ pure\_explore}$}
            \STATE // Start looking for the best column with high probability
            \STATE $w=\log_2\log_2(T)$
            \STATE $\text{column\_explore}=$True
        \ENDIF
        \IF{$\text{column\_explore}=$True, $|J|$=1 and $\text{ pure\_explore}$}
            \STATE // Return best entry
            \STATE return $(I,J)$
        \ENDIF
    \ENDFOR
\ENDFOR
\end{algorithmic}
\end{algorithm}

The algorithm uses function {\tt domination\_map} that defines $h^U$ by using $Q^U_{+}$ and $Q^U_{-}$ through the following equation:
$$
h^U(i) = i \mathds{1}_{ \{\exists k \in [0,M] \text{ s.t. }\forall j \in [N], Q_{+}(k,i)>Q_{-}(k,j)\}} +\max_{j \in [N]} j \mathds{1}_{\{ \exists k \in [0,M] \text{ s.t. }Q_{+}(k,i)<Q_{-}(k,j)\}},\ \forall i \in U.
$$

The algorithm uses function {\tt confidence\_bound}, with input arguments $X,C,h,Q_{+},Q_{-}$, and $p$,  
and outputs $Q_{+}$ and $Q_{-}$ defined as follows. 
For each $(k,i) \in U\times V$, if $C(k,i) = k_l$, with $k_l :=\lceil 4^{l+1} \log(\beta_l)\rceil$ for some integer $l>0$, then:
$$
Q_{-}(k,i)= \frac{X(k,i)}{C(k,i)}-\sqrt{\frac{\log(\beta_l)}{k_l}} \hbox{ and } Q_{+}(k,i)= \frac{X(k,i)}{C(k,i)}+\sqrt{\frac{\log(\beta_l)}{k_l}}.
$$
For each $i\in U$, if $\sum_{k=1}^M C(k,i)=k_l$ for some $k_l$,then:
$$
Q_{-}(0,i)= \frac{\sum_{k=1}^M X(k,i)}{\sum_{k=1}^M C(k,i)}-\sqrt{\frac{\log(\beta_l)}{k_l}} \hbox{ and } Q_{+}(0,i)= \frac{\sum_{k=1}^M X(k,i)}{\sum_{k=1}^M C(k,i)}+\sqrt{\frac{\log(\beta_l)}{k_l}}.
$$

In all other cases, values in $Q_{+}$ and $Q_{-}$ remain unchanged.
 The value used for parameter $\beta_l$ depends on $h$ and $p$. In the case where $p$ is false, $\beta_l=h$. When $p$ is true $\beta_l=\pi \sqrt{(N+1)(M+1)h/3}\cdot l$.
 

\subsection{Proof of the upper bound}

In this section, we prove Theorems \ref{Regret2S} and \ref{UBIB-L}. The proof starts by a lemma that bounds the probability that any of the built confidence interval fails. It is given in a slightly more general form than needed, as it will be used in other proofs.

Let $x_1,...,x_{k}$ be a sequence of $k$ independent samples from some distributions, and $\hat{x}$ denote the empirical mean
$$
\hat{x} = \frac{1}{k}\sum_{s=1}^{k} x_s.
$$

Let us denote $x^* = \mathds{E}[\hat{x}]$. The lower and upper bounds for $x^*$ are defined for $k = k_l$, with $k_l =\lceil4 \Delta_{l}^{-2} \log(\beta_l)\rceil$ where $\Delta_{l}= 1/2^l$ and $\beta_l$ is a parameter, as follows
$$
L_l(\hat{x}) = \hat{x}-\sqrt{\frac{\log(\beta_l)}{k_l}} \hbox{ and } U_l(\hat{x}) = \hat{x}+\sqrt{\frac{\log(\beta_l)}{k_l}}.
$$

Let $\mathcal{E}_l(\hat{x})$ be the event that $x^*$ lies within the interval $[L_l(\hat{x}), U_l(\hat{x})]$, i.e.
$$
\mathcal{E}_l(\hat{x}) = \{x^* \in [L_l(\hat{x}),U_l(\hat{x})]\}.
$$

\begin{lemma}\label{BoundProbaFail} The event $\mathcal{E}_l(\hat{x})$ holds with probability at least $1-2/\beta_l^2$.
\end{lemma}

The proof of the lemma follows by direct application of Hoeffding's inequality and is omitted. \hfill \(\Box\)

We also state a general purpose lemma that bounds the number of samples before two sets of samples can be ranked. 

Let $x_1,\ldots,x_k$ and $y_1,\ldots,y_k$ be two sequences of sampled values, and $\hat{x}$ and $\hat{y}$ be their respective empirical means. Let $\Delta(\hat{x},\hat{y}) = x^*-y^*>0$.

\begin{lemma}\label{maxsample}
If $\Delta_l< \Delta(\hat{x},\hat{y})/2$, then
$$
U_l(\hat{y})<L_l(\hat{x}).
$$
\end{lemma}

\textbf{Proof of Theorem \ref{Regret2S} (Regret)}: We first restate the theorem here:
\PairElimRegret*

The proof proceeds as follows. Let $\mathcal{A}$ be the "good event" that all row confidence intervals hold. Lemma \ref{BoundProbaFail} together with a union bound on the $N(M+1)$ tracked parameters and all confidence interval's updates up to horizon $T$ gives:
$$
\Pb[\overline{\mathcal{A}}]\leq 2\frac{N(M+1)}{T}.
$$



A first bound on the regret follows 
\begin{align*}
    R(T) &\leq \E[R(T)\mid \mathcal{A}]+T\Pb[\overline{\mathcal{A}}]\\
    & \leq \E[R(T)\mid \mathcal{A}] + 2N(M+1). \numberthis \label{eqn:rowprobafail}
\end{align*}

Let $\mathcal{B}_w$  be the "good event $w$" that all column related confidence intervals hold during time window $w$. Repeating the same arguments, we have
$$
2^{2^w}\Pb[\overline{\mathcal{B}}_w] \leq 2(N+1)M.
$$

Note that $w$ ranges from $0$ to $w_{\mathrm{max}}$, where $w_{\max}$ is the largest integer $k$ such that $2^{2^{k-1}} \leq T$. Hence, we have
$$
w_{\mathrm{max}} \leq \log_2( \log_2 (T))+1.
$$

Which gives the following bound:

\begin{equation}\label{eqn:columnprobafail}
  \sum_{w=0}^{w_{\mathrm{max}}}2^{2^w}\Pb[\overline{\mathcal{B}}_w]\leq 2(N+1)M(\log_2( \log_2 (T)) +2).  
\end{equation}

Under event $\mathcal{B}_w$, optimal column $1$ is never eliminated in time window $w$. Let $n_{1,j,w}$ denote the number of times column $j$ is sampled against the optimal row $1$ during time window $w$. Note that this number is equal for all non-eliminated columns. This, together with Lemma \ref{maxsample}, gives the following bound
$$
n_{1,j,w} \leq \left\lceil\frac{64\log(2^{2^w})}{(u_1\Delta_j^V)^2} \right\rceil.
$$

Let $R_{1,j,w}$ be the total regret incurred by the sampling of pair $(1,j)$, at a sampling step where $1$ is the picked active row with which columns are sampled against, during time window $w$. We show
\begin{equation}\label{eqn:rowexplore}
\sum_{w=0}^{w_{\text{max}}}R_{1,j,w} \leq \frac{512}{u_1\Delta_j^V}\log(T)+2\log_2(\log_2(T))+4.
\end{equation}

This follows from the following inequalities

\begin{align*}
    \sum_{w=0}^{w_{\text{max}}}R_{1,j,w} &\leq 2u_1 \Delta_j^V \sum_{w=0}^{w_{\text{max}}}n_{1,j,w} \\
    &\leq\frac{128}{u_1\Delta_j^V}\left(\sum_{w=0}^{w_{\text{max}}}2^w\right)\log(2)+2w_{\text{max}}+2\\
    &\leq\frac{512}{u_1\Delta_j^V}\log(T)+2\log_2(\log_2(T))+4.\\
\end{align*}
The factor $2$ in the first inequality accounts for the possibility that active column $j$ is the column that active rows are sampled against at a given sampling step where $1$ is the picked row.

Similarly, under event $\mathcal{A}$, the number of times row $i$ is sampled against the optimal column $1$, denoted as $n_{i,1}$, is bounded as
$$
n_{i,1} \leq \left\lceil\frac{64\log(T)}{(v_1\Delta_i^U)^2} \right\rceil.
$$

Thus, we have
\begin{equation}\label{eqn:columnexplore}
  \sum_{w=0}^{w_{\text{max}}}R_{i,1,w} \leq 2v_1 \Delta_i^U n_{i,1} \leq\frac{128}{v_1 \Delta_i^U}\log(T)+1. 
\end{equation}

Under event $\mathcal{A}$, when row $i$ is eliminated, it is mapped by $h^U$ to an item with higher parameter. Thus, the following property always holds
$$
\sum_{i \in [N]} u_{h^U(i)} \geq \sum_{i\in[N]} u_i=N\mu_U.
$$

We let $n_{j,w}$ denote the number of sampling steps in time window $w$ before suboptimal column $j$ is eliminated. Thus, under event $\mathcal{B}_w$, according to Lemma~\ref{maxsample}, we have
$$
n_{j,w} \leq \left\lceil\frac{64\log(2^{2^w})}{(\mu_U \Delta_j^V)^2} \right\rceil.
$$

If the stochastic rewards are Bernoulli random variables, arms including column $j$ such that $v_j=0$ return a deterministic reward of value $0$. Thus, those columns are eliminated before any other. We define $v_{\mathrm{min}}= \min\{v_j: j\in [C], v_j>0\}$. If the stochastic rewards are not Bernoulli random variables, then $v_{\mathrm{min}}=v_N$.

The following property always holds
$$
\sum_{j \in V} v_{h^V(j)} \geq M\min\{v_{\mathrm{min}},\mu_V\}.
$$

We let $\tilde{v}_{\mathrm{min}}=\min\{v_{\mathrm{min}},\mu_V\}$ and $n_i$ be the number of sampling steps before row $i$ is eliminated. The previous inequality  together with Lemma~\ref{maxsample} gives that under event $\mathcal{A}$, 
$$
n_i = \left\lceil \frac{64\log(T)}{(\tilde{v}_{\mathrm{min}}\Delta_i^U)^2}\right \rceil.
$$

This means that row $i$ stops being sampled by the end of the first time window $w_i$ such that $2^{2^{w_i}}>(N+M)n_i>2^{2^{w_i-1}}$. Thus, we have
$$
2^{w_i+1}\leq 4\log_2(n_i)+4\log_2(N+M).
$$

Let $R_{i,j,w}$ be the regret incurred by the sampling of pair $(i,j)$ at a sampling step where row $i$ and/or column $j$ are the active rows/columns other entries are sampled against. In the special case where both row $i$ and column $j$ are picked, $R_{i,j,w}$ also account for the "useless" samples $(i,1)$ and $(1,j)$, useless in the sense that they are not used to update the confidence intervals related to items $i$ and $j$.
The following relations hold:
\begin{align*}
    \sum_{w=0}^{w_{\text{max}}}R_{i,j,w}
    \leq& 4(u_1v_1-u_iv_j)\sum_{w=0}^{w_i}n_{j,w}\\
    \leq& 256\frac{\Delta_i^U+\Delta_j^V}{(\mu_U \Delta_j^V)^2}\log(2)\sum_{w=0}^{w_i}2^w+w_i+1\\
    \leq& 256\frac{\Delta_i^U+\Delta_j^V}{(\mu_U\Delta_j^V)^2}\log(2)2^{w_i+1}+4w_i+4\\
    \leq& 1024\frac{\Delta_i^U+\Delta_j^V}{(\mu_U\Delta_j^V)^2}\log\left( \frac{64\log(T)}{(\tilde{v}_{\mathrm{min}}\Delta_i^U )^2}+1\right)+4\log_2\left(4\log\left(\frac{64\log(T)}{(\tilde{v}_{\mathrm{min}}\Delta_i^U)^2}+1\right)+1\right) \numberthis \label{eqn:doublysuboptimal}\\
    &+1024\frac{\Delta_i^U+\Delta_j^V}{(\mu_U\Delta_j^V)^2}\log\left(N+M\right)+4\log_2(4\log_2(N+M))+4.
\end{align*}

Summing up the obtained terms (\ref{eqn:rowprobafail}), (\ref{eqn:columnprobafail}), (\ref{eqn:columnexplore}), (\ref{eqn:rowexplore}) and (\ref{eqn:doublysuboptimal}), completes the proof of the theorem. \hfill \(\Box\)

\textbf{Proof of Theorem \ref{UBIB-L} (Pure exploration)}: We first restate the theorem:
\PairElimExplore*

The proof goes as follows. Let $\mathcal{A}$ be the "good event" that all high probability confidence intervals hold. Lemma \ref{BoundProbaFail} together with a union bound on the $N(M+1)+M$ tracked parameters and all confidence interval's updates gives:
$$
\Pb[\overline{\mathcal{A}}]\leq \delta.
$$

Note that the algorithm runs in two phases: the first phase where the best row is identified, which is followed by the second phase in which the best column is identified, using sampling against the best row identified in the first phase.

At every sampling step, the expected values of the column active rows are sampled against is at least $v_{\text{min}}$. Lemma \ref{maxsample} implies that rows $1$ and $i$ are guaranteed to be relatively ranked by the first sampling step where $\sum_{j=1}^{M}C(i,j)=k_l$  with:
$$\frac{1}{2^{l+1}}<\sqrt{\frac{\log(\beta_l)}{k_l}}<\frac{1}{4}v_{\text{min}}\Delta^U_i.$$
This implies
\begin{align*}
    \frac{1}{4}v_{\text{min}}\Delta^U_i&\leq \frac{1}{2^l}.
\end{align*}
Thus, we have
$$
k_l \leq   \frac{32}{(v_{\text{min}}\Delta^U_i)^2}\log\left(\frac{1}{\delta}\right)+\frac{32}{(v_{\text{min}}\Delta^U_i)^2}\log\left(\frac{\pi^2N(M+1)}{3}\right) 
+64\log\left(\log_2\left(\frac{1}{v_{\text{min}}\Delta^U_i}\right)+2\right).
$$

Let $\tau^U$ be the number of sampling steps after which the best row is detected with probability at least $1-\delta$, i.e. $\tau^U$ is the length of the first phase. The previous inequality implies:
\begin{eqnarray}\label{eqn:tau_u}
\tau^U & \leq & \frac{32}{(v_{\text{min}}\Delta^U_2)^2}\log\left(\frac{1}{\delta}\right)+\frac{32}{(v_{\text{min}}\Delta^U_2)^2}\log\left(\frac{\pi^2N(M+1)}{3}\right) 
+64\log\left(\log_2\left(\frac{1}{v_{\text{min}}\Delta^U_2}\right)+2\right).
\end{eqnarray}

We start by bounding the number of "bad" samples, that is samples $(i,j)$ with $j>1$. Repeating the computations used to prove equation \ref{eqn:columnprobafail}, the  expected number of "bad" samples due to the failures of a confidence interval during any time window $w$ is bounded as :
\begin{equation}\label{eqn:badsampleconffail}
    \sum_{w=0}^{w_{\mathrm{max}}}2^{2^w}\Pb[\overline{\mathcal{B}}_w]\leq 2(N+1)M(\log_2( \log_2 ((N+M)\tau^U)) +2).
\end{equation}

At every sampling step, the expected values of the column active rows are sampled against is at least $\mu_U$. Repeating the computations used to obtain equation \ref{eqn:columnexplore}, under the good event the confidence intervals hold during the successive time windows, the number of sampling steps where sub-optimal column $j$ is active during the first phase, $\tau_{j}^V(1) $, is bounded as:
\begin{equation}\label{eqn:badsample}
\tau_{j}^V(1) \leq 256\frac{1}{(\mu_U\Delta_j^V)^2}\log\left((N+M)\tau^U\right)+\log_2\left(4\log((N+M)\tau^U))+1\right).
\end{equation}

In a sampling step where column $j$ is active, it is sampled at most $N+1$ times. Thus the number of "bad" samples involving column $j$ is less than $(N+1)\tau_{j}^V(1)$.

The number of sampling steps where $1$ is the picked column active rows are sampled against before sub-optimal entry $i$ is eliminated, $\tau^U_i$, is bounded as:
\begin{eqnarray*}
\tau_i^{U}(1) & \leq & \frac{32}{(v_1\Delta^U_i)^2}\log\left(\frac{1}{\delta}\right)+\frac{32}{(v_1\Delta^U_i)^2}\log\left(\frac{\pi^2N(M+1)}{3}\right) 
+64\log\left(\log_2\left(\frac{1}{v_1\Delta^U_i}\right)+2\right).
\end{eqnarray*}

The best row $1$ is sampled at every sampling step until all other rows have been eliminated. Thus, we have
\begin{eqnarray*}
\tau_1^{U}(1) & \leq & \frac{32}{(v_1\Delta^U_2)^2}\log\left(\frac{1}{\delta}\right)+\frac{32}{(v_1\Delta^U_2)^2}\log\left(\frac{\pi^2N(M+1)}{3}\right)
+64\log\left(\log_2\left(\frac{1}{v_1\Delta^U_2}\right)+2\right).
\end{eqnarray*}

Also, in those sampling steps, column $1$ is sampled against an active row. The number of "good" samples $(i,1)$ is thus bounded by 
\begin{equation}\label{eqn:samplei1}
3\tau_1^{U}(1)+\sum_{i=2}^N\tau_i^{U}(1).
\end{equation}
Similarly, under the good event that the confidence intervals hold, during the second phase of the algorithm, column $j>1$ is eliminated after $\tau_j^V(2)$ samples, with
\begin{align*}
\tau_j^{V}(2)  \leq & \frac{32}{(u_1\Delta^V_j)^2}\log\left(\frac{1}{\delta}\right)+\frac{32}{(u_1\Delta^V_i)^2}\log\left(\frac{\pi^2M}{3}\right)\numberthis \label{eqn:phase2}
+64\log\left(\log_2\left(\frac{1}{u_1\Delta^V_j}\right)+2\right).
\end{align*}

The best column $1$ is sampled at every sampling step until all other columns have been eliminated. Thus, we have
\begin{align*}
\tau_1^{V}(2) \leq & \frac{32}{(u_1\Delta^V_2)^2}\log\left(\frac{1}{\delta}\right)+\frac{32}{(u_1\Delta^V_2)^2}\log\left(\frac{\pi^2M)}{3}\right)\numberthis \label{eqn:phase21} 
+64\log\left(\log_2\left(\frac{1}{u_1\Delta^V_2}\right)+2\right).
\end{align*}

Summing up equations (\ref{eqn:badsampleconffail}), $(N+1)$(\ref{eqn:badsample}), (\ref{eqn:samplei1}), (\ref{eqn:phase2}) and (\ref{eqn:phase21}) gives the result.
\hfill \(\Box\)

\subsection{Proof of the lower bound} 

Let $d(x, y)=x \log (x / y)+(1-x) \log ((1-x) /(1-y))$ denote the binary relative entropy.
 Let $F_{\mu}$ denote the distribution of the stochastic reward of a row-column pair with the product of the parameters equal to $\mu$. We admit the following assumption. 
 \begin{hyp}\label{hyp:kl} For every $p,q \in [0,1]$, and $\kappa \geq 1$ such that $\kappa p, \kappa q \in [0,1]$, it holds 
 $$
 \mathrm{D_{KL}}(F_{p}||F_{q}) \leq \mathrm{D_{KL}}(F_{\kappa p}|| F_{\kappa q}).
 $$
 \end{hyp}
 
 The assumption holds for normal distributions and Bernoulli distributions as explained next. Let $F_{\mu_1}$ and $F_{\mu_2}$ be two normal distribution with means $\mu_1$ and $\mu_2$, and variances $\sigma_1^2$ and $\sigma_2^2$. Then, we have $\mathrm{D_{KL}}(F_{\mu_1}|| F_{\mu_2}) = (1/(2\sigma_2^2))(\mu_1-\mu_2)^2 + \sigma_1^2/(2\sigma_2^2) + \log(\sigma_1/\sigma_2) - 1$. From this, it is readily observed that Assumption~\ref{hyp:kl} holds. Now, assume that $F_{\mu_1}$ and $F_{\mu_2}$ are two Bernoulli distributions with means $\mu_1$ and $\mu_2$, then by the data processing inequality in Theorem 2.2 and Proposition 6 in \cite{polyanskiy2014lecture}, it follows that Assumption~\ref{hyp:kl} holds.

We next prove the following theorem:

 \begin{theorem}[lower bound]
\label{LB-pairs}
Assume that stochastic rewards of row-column pairs satisfy Assumption~\ref{hyp:kl}. Then, for any $\delta$-PAC algorithm, the expected sampling complexity is lower bounded as
$$
\E[\tau_{\delta}] \geq A_F(u,v) d(\delta, 1-\delta)
$$
where 
\begin{eqnarray*}
A_F(u,v) &:=&\frac{1}{2}\left( \sum_{i=2}^N \frac{1}{\mathrm{D_{KL}}(F_{u_iv_1}|| F_{(u_i+\Delta^U_i)v_1})} +\sum_{j=2}^M
\frac{1}{\mathrm{D_{KL}}(F_{u_1 v_j}||F_{u_1(v_j+\Delta^V_j)})} \right).
\end{eqnarray*}
\end{theorem}

\textbf{Proof}: Let $N_{i,j}(t)$ denote the number of time steps in which pair $(i,j)$ is sampled until step $t$, i.e.
$$
N_{i,j}(t) =\sum_{s=1}^{t} \mathds{1}_{\left\{(i_t,j_t)=(i,j)\right\}}.
$$

We use the following lemma of \cite{kaufmann2016complexity} adapted to our setting.

\begin{lemma}\label{kaufmann}. Consider two problem instances with parameters $A=(u,v)$ and $A'=(u',v')$. For any almost-surely finite stopping time $\sigma$ with respect to the filtration $\mathcal{F}_{t}$, we have
\[
\sum_{i<j}\mathds{E}_{A}\left[N_{i,j}(\sigma)\right] \mathrm{D_{KL}}(F_{u_iv_j}|| F_{u_i^{\prime}v_j^{\prime}}) \geq \sup _{\mathcal{E} \in \mathcal{F}_{\sigma}} d\left(\mathds{P}_{A}(\mathcal{E}), \mathds{P}_{A^{\prime}}(\mathcal{E})\right)
\]
\end{lemma}

For any given $u = (u_1, \ldots, u_N)$ and $\alpha>0$, we define the following model alternatives
$$
U_{\alpha} = \{u_{\alpha,i}= (u_1,..,u_i',...,u_N): u_i'= u_i + \Delta_i^U+\alpha),  1<i\leq N\}
$$

and we define $V_{\alpha}$ analogously.

Let $\mathcal{E}$ denotes the event that the algorithm outputs pair $(1,1)$. Let $A_{\alpha,i} = (u_{\alpha,i}, v)$. For any $\delta$-PAC algorithm, $\mathds{P}_{A}(\mathcal{E})>1-\delta$ and $\mathds{P}_{A_{\alpha,i}}(\mathcal{E})<\delta$ (similarly for $(u,v_{\alpha,i})$). This, together with Lemma~\ref{kaufmann} gives that the solution to the following linear program is a lower bound on the sample complexity of any $\delta$-PAC algorithm:
\begin{align*} 
\mathbf{LP}_\alpha: & \\
\text{ minimize } & \sum_{i,j} x_{i,j}\\
\text { subject to } & \sum_{j=1}^{M} x_{i,j}\mathrm{D_{KL}}(F_{u_iv_j}|| F_{u_{\alpha,i}v_j}) \geq d(\delta,1-\delta), \hbox{ for } i \in [N]\\
& \sum_{i=1}^{N} x_{i,j} \mathrm{D_{KL}}(F_{u_iv_j}|| F_{u_iv_{\alpha,j}}) \geq d(\delta,1-\delta), \text{ for } j \in [M].
\end{align*}

Let $x^* = (x^*_{i,j}: (i,j)\in [N]\times [M])$ be an optimal solution to $\mathbf{LP}_\alpha$, and let $\hat{x} = (\hat{x}_{i,j}: (i,j)\in [N]\times [M])$ be defined as follows
$$
\hat{x}_{i,j} = \left\{
\begin{array}{ll}
\sum_{j'=1}^M x^*_{1,j} +\sum_{i'=1}^N x^*_{i',1} & \hbox{ for } i = 1 \hbox{ and } j = 1\\
\sum_{i'=1}^N x^*_{j,i'} & \hbox{ for } i = 1 \hbox{ and } 1<j\leq M\\
\sum_{j'=1}^M x^*_{i,j'} & \hbox{ for } 1<i\leq N \hbox{ and } j = 1\\
0 & \hbox{ otherwise}.
\end{array}
\right .
$$


The vectors $x^*$ and $\hat{x}$ satisfy the following equation
$$
\sum_{i,j} \hat{x}_{i,j} = 2 \sum_{i,j} x^*_{i,j}.
$$

By Assumption~\ref{hyp:kl}, it follows that $\hat{x}$ is an admissible solution to the following linear program: 

\begin{align*} 
\mathbf{LP}_\alpha': & \\
\text{ minimize } & \frac{1}{2}\sum_{i=1}^{N} x_{i,1}+\frac{1}{2}\sum_{j=1}^{M} x_{1,j}\\
\text { subject to } &  x_{i,1} \geq \frac{1}{\mathrm{D_{KL}}(F_{u_iv_1} || F_{u_{\alpha,i}v_1}) } d(\delta,1-\delta), \hbox{ for } i \in [N]\\
 &  x_{1,j} \geq \frac{1}{\mathrm{D_{KL}}(F_{u_1v_j}|| F_{u_1 v_{\alpha,j}}) } d(\delta,1-\delta), \hbox{ for } j \in [M].
\end{align*}

Thus, the optimal solution to $\mathbf{LP}_\alpha'$ is a lower bound on the sample complexity. This lower bound holds for any $\alpha >0$. Letting $\alpha$ go to zero establishes the proof of Theorem~\ref{LB-pairs}. \hfill \(\Box\)

We next show lower bounds on the sampling complexity for stochastic rewards according to normal distributions, and then after according to Bernoulli distributions.

\begin{theorem}\label{unitgaussian} Assume that stochastic rewards of item pairs have normal distributions with unit variance. Then, we have
$$
\E[\tau_{\delta}] \geq A(u,v) \left(\log\left(\frac{1}{\delta}\right)-1\right)
$$
where
$$
A(u,v) = \sum_{i\in [N]: \Delta_i^U > 0} \frac{1}{(v_1 \Delta^U_i)^2}+\sum_{j\in [M]: \Delta_j^V>0}
\frac{1}{(u_1\Delta^V_j)^2}.
$$

\label{cor:1}
\end{theorem}

{\bf Proof:} For stochastic rewards of item pairs according to normal distributions with unit variances, we have $\mathrm{D_{KL}}(F_{\mu_1}\mid \mid F_{\mu_2}) = (\mu_1-\mu_2)^2/2$. This combined with the fact,
$$
d(\delta, 1-\delta) = (1-2\delta) \log\left(\frac{1-\delta}{\delta}\right) \geq \log\left(\frac{1}{\delta}\right)-1
$$
yields the statement of the theorem.
\hfill\(\Box\)

\begin{theorem} Assume that stochastic rewards of item pairs are Bernoulli random variables. Then, we have
\begin{equation*}\label{eq:1}
\E[\tau_{\delta}] \geq \frac{\min\{u_1 v_1, 1-u_1 v_1\}}{4} A(u,v) \left(\log\left(\frac{1}{\delta}\right)-1\right).
\end{equation*}
\label{cor:2}
\end{theorem}

\textbf{Proof}: If for some $\alpha \in (0,1/2]$, $\alpha \leq q \leq 1-\alpha$, then
$$
d(p,q) \leq \frac{2}{\alpha}(p-q)^2.
$$


Let us first consider the term $\mathrm{D_{KL}}(F_{u_iv_1}|| F_{(u_i+\Delta^U_i)v_1})$. Let $p = u_i v_1$ and $q = (u_i+\Delta^U_i)v_1$. Note that $q = u_1 v_1$. We can apply the above upper bound for the KL divergence by taking $\alpha = \min\{u_1 v_1, 1-u_1 v_1\}$, and note that $(p-q)^2 = (v_1\Delta^U_i)^2$. It follows 
$$
\mathrm{D_{KL}}(F_{u_iv_1}|| F_{(u_i+\Delta^U_i)v_1}) \leq \frac{2(v_1 \Delta^U_i)^2}{\min\{u_1 v_1, 1-u_1 v_1\}} .
$$
By the same arguments, we have
$$
\mathrm{D_{KL}}(F_{u_1v_j}|| F_{u_1(v_j+\Delta^V_j)}) \leq \frac{2(u_1 \Delta^V_j)^2}{\min\{u_1 v_1, 1-u_1 v_1\}} .
$$\hfill \(\Box\)

\section{\RMono\ algorithm}\label{RMonoAppendix}

\subsection{Algorithm description and pseudo-code}

The \RMono\ algorithm is similar to the \SRankElim\ algorithm. The main differences are in the definition of the confidence intervals and the set of active pairs. The pseudo-code of the algorithm is shown in Algorithm~\ref{alg:SRankElimMono}.

\begin{algorithm}[H]
\caption{\RMono\ }
\label{alg:SRankElimMono}
\begin{algorithmic}
\STATE {\bfseries Input:} set of items $U$, precision $\delta$ (or $\frac{1}{T}$) and $\text{ pure\_explore}$
\STATE $t=0, C=X=C_w=X_w=[0]^{2N\times 2N}, S=\{(i,j)|i\in[2N], j\in[2N],i\neq j\}$
\WHILE{$t \leq T$ }
    \FOR{window $w=0,1,2,\dots$}
        \STATE $Q_{+}^V,Q_{-}^V \leftarrow$ {\tt confidence\_bound}($X_w,C_w,2^{2^w},Q_{+}^V,Q_{-}^V,\text{False}$)
        \STATE $Q_{+}^U,Q_{-}^U \leftarrow$ {\tt confidence\_bound}($X,C,T,Q_{+}^U,Q_{-}^U,\text{ pure\_explore}$)
        \STATE   $S \leftarrow$ {\tt active\_entries}($Q_{+}^U,Q_{-}^U,Q_{+}^V,Q_{-}^V$,$S$)\;
        \FOR{$(i,j)\in S$}
            \STATE $X(i,j) \gets X(i,j) + x_{i,j,t}$
            \STATE $C(i,j) \gets C(i,j) + 1$
            \STATE $X_w(j,i) \gets X_w(j,i) + x_{i,j,t+1}$
            \STATE $C_w(j,i) \gets C_w(j,i) + 1$
            \STATE $s \gets s + 2$, $t \gets t + 2$
        \ENDFOR
        \IF{$s>2^{2^w}$}
            \STATE // Change time window
            \STATE $s=0,C_w=X_w=[0]^{N\times M}$
        \ENDIF
        \IF{\text{ pure\_explore} and {\tt optimal\_pair}($Q_{+}^U,Q_{-}^U$)}
            \STATE return optimal pair
        \ENDIF
    \ENDFOR
\ENDWHILE
\end{algorithmic}
\end{algorithm}

The function {\tt confidence\_bound}($X,C,h,Q_{+},Q_{-},p$) updates $Q_{+}$ and $Q_{-}$ as follows. For each $(i,j) \in U\times U$, if $C(i,j) = k_l$, with $k_l :=\lceil 4^{l+1} \log(h)\rceil$ for some integer $l>0$, then:
$$
Q_{-}(i,j)= \frac{X(i,j)}{C(i,j)}-\sqrt{\frac{\log(h)}{k_l}} \hbox{ and } Q_{+}(i,j)= \frac{X(i,j)}{C(i,j)}+\sqrt{\frac{\log(h)}{k_l}}.
$$
If $C'(i,j)+\sum_{k=1, k\neq i,j}^{2N} C(i,k)=k_l$ for some $k_l$, then:
$$
Q_{-}(i,2N+j)= \frac{\sum_{k=1, k\neq i,j}^{2N}X(i,k)}{\sum_{k=1, k\neq i,j}^{2N} C(i,k)}-\sqrt{\frac{\log(h)}{k_l}} 
$$
and
$$
Q_{+}(i,2N+j)= \frac{\sum_{k=1, k\neq i,j}^{2N} X(i,k)}{\sum_{k=1, k\neq i,j}^{2N} C(i,k)}+\sqrt{\frac{\log(h)}{k_l}}.
$$
The value used for parameter $\beta_l$ depends on $h$ and $p$. 
In the case where $p$ is false, $\beta_l=h$. When $p$ is true $\beta_l=\pi \sqrt{4N(2N-1)h/3}\cdot l$

\bigskip
The function {\tt active\_entries} updates $S$ as follows. For any item $i$:
\begin{itemize}
    \item if there exist two entries $j$ and $j'$ s.t. for some $k$, $Q_{+}^U(i,k)<Q_{-}^U(j,k)$ and for some $k'$, $Q_{+}^U(i,k')<Q^U_{-}(j',k')$, then all entries $(i,l)$ and $(l,i)$ are removed from $S$. 
    \item if there exist two entries $j$ and $j'$ s.t. for some $k$, $Q_{+}^V(i,k)<Q^V_{-}(j,k)$ and for some $k'$, $Q^V_{+}(i,k')<Q^V_{-}(j',k')$, then all entries $(l,i)$ are removed from $S$. 
    \item if there exist an entry $j$ s.t. for some $k$, $Q_{+}^U(i,k)<Q_{-}^U(j,k)$ then all entries $(i,l)$ and $(l,i)$ are removed from $S$ except for $(i,j)$ and $(j,i)$.
    \item if there exist an entry $j$ s.t. for some $k$, $Q_{+}^V(i,k)<Q_{-}^V(j,k)$ then all entries $(l,i)$ are removed from $S$ except for $(j,i)$. 
\end{itemize}

\subsection{Proof of the upper bounds}

Many computational steps of the two following proofs are similar to those given in Appendix\ref{SRankElimAppendix} and are not repeated here.

\textbf{Proof of Theorem \ref{monopartite} (Regret)} We first provide the proof of the following theorem.
\PairElimMonoregret*
The proof goes as follows. \RMono\ algorithm tracks $2N(N-1)$ row parameters, and there are less than $T$ confidence interval updates. Thus, Lemma~\ref{BoundProbaFail} together with a union bound yield the following bound
\begin{equation}\label{eqn:rowprobafailMono}
\Pb[\overline{\mathcal{A}}]\leq 4\frac{N(N-1)}{T}.
\end{equation}
Similarly, we have
\begin{equation}\label{eqn:columnprobafailMono}
  \sum_{w=0}^{w_{\mathrm{max}}}2^{2^w}\Pb[\overline{\mathcal{B}}_w]\leq 4N(N-1)(\log_2( \log_2 (T)) +2).  
\end{equation}

From Lemma \ref{maxsample}, under event $\mathcal{A}$ and $\mathcal{B}_w$, the number of sampling steps where column $j$ is sampled against row $1 $ or row $2$, $n_{1,j,w}$ and $n_{2,j,w}$, during time window $w$ are bounded as
$$
n_{1,j,w} \leq \left\lceil\frac{64\log(2^{2^w})}{(u_1\Delta_{2,j})^2} \right \rceil \hbox{ and } n_{2,j,w} \leq \left\lceil\frac{64\log(2^{2^w})}{(u_2\Delta_{1,j})^2} \right\rceil .
$$

Thus, the total regret for sampling of pair $(1,j)$, $R_{1,j}(T)$,  is bounded as
\begin{align}\label{eqn:regret1j}
    R_{1,j}(T) &\leq\frac{512}{u_1\Delta_{2,j}}\log(T)+2\log_2(\log_2(T))+4
\end{align}

and the fact $u_1\Delta_{2,i} < u_2\Delta_{1,i}$ gives
\begin{equation}\label{eqn:regret2j}
   R_{2,j}(T)  \leq R_{1,j}(T). 
\end{equation}

Similarly, we have
\begin{align}\label{eqn:regreti1}
    R_{i,1}(T) &\leq\frac{128}{u_1 \Delta_{2,i}}\log(T)+1
\end{align}
and
\begin{equation}\label{eqn:regreti2}
R_{i,2}(T) \leq R_{i,1}(T).
\end{equation}

Under assumption that event $\mathcal{A}$ occurs, rows $1$ and $2$ are active at every iteration. Thus, at every sampling step, column $j$ is sampled against the two optimal rows and is definitely eliminated at most once it is found smaller than column $2$, which gives
$$
n_{j,w} \leq \left\lceil\frac{64\log(2^{2^w})}{(u_1 \Delta_{2j})^2} \right\rceil .
$$

Consider a doubly sub-optimal pair $(i,j)$ with $i>j$. Doubly sub-optimal pair $(i,j)$ definitely stops being sampled as soon as $i$ has been deemed smaller than any other item $\neq j$. Before this happens, $i$ can be compared with $1$ or $2$ against at least one of the active columns. The number of sampling steps before doubly sub-optimal pair $(i,j)$ is definitely eliminated is thus upper bounded as

$$
n_{ij} = \left \lceil \frac{128\log(T)}{(\tilde{u}_{\mathrm{min}}\Delta_{2,i})^2}\right\rceil .
$$
with $\tilde{u}_{\mathrm{min}}=\min\{u_{\mathrm{min}},\mu_U\}$, $u_{\mathrm{min}}= \min\{u_j: j\in [C], u_j>0\}$.

Consider a doubly sub-optimal pair $(i,j)$ with $i<j$. It may happen that the order $i<j$ is discovered before any other. In that case, $i$ is only sampled against $j$ and can no longer be compared with either $1$ or $2$ at every sampling step, since $j$ might get eliminated as a column. However, doubly sub-optimal pair $(i,j)$  is still definitely eliminated as soon as $j$ is deemed smaller than any other entry, and number of sampling steps before doubly sub-optimal entry  $(i,j)$ is definitely eliminated is thus upper bounded as
$$
n_{ij} = \left \lceil \frac{128\log(T)}{(\tilde{u}_{\mathrm{min}}\Delta_{2,j})^2}\right\rceil .
$$
Thus, the regret for the sampling of a doubly sub-optimal entry $(i,j)$ is bounded as
\begin{align*}
    \sum_{w=0}^{w_{\text{max}}}R_{i,j,w} 
    \leq& 512\frac{\Delta_{2,i}+\Delta_{2,j}}{(u_1\Delta_{2,j})^2}\log\left( \frac{128}{(\tilde{u}_{\mathrm{min}}\min\{\Delta_{2,i},\Delta_{2,j}\} )^2}\log(T)+1\right)\\
    & +\log_2\left(4\log\left(\frac{128\log(T)}{(\tilde{u}_{\mathrm{min}}\min\{\Delta_{2,i},\Delta_{2,j}\})^2}+1\right)+1\right)\numberthis \label{eqn:doublysuboptMono}\\
    &+512\frac{\Delta_i^U+\Delta_j^V}{u_1\Delta_{2,j}^2}\log\left(4N(2N-1)\right)+4\log_2(4\log_2(4N(2N-1)))+4.
\end{align*} 

The results follows from equations (\ref{eqn:rowprobafailMono}), (\ref{eqn:columnprobafailMono}), (\ref{eqn:regret1j}), (\ref{eqn:regret2j}), (\ref{eqn:regreti1}), (\ref{eqn:regreti2}) and (\ref{eqn:doublysuboptMono}).\hfill \(\Box\)

\textbf{Proof of Theorem \ref{TH:mono_id} (Pure Exploration)}: We next prove the following theorem.
\PairElimMonoexplore*
The proof goes as follows. Let $\mathcal{A}$ be the "good event" that all high probability confidence intervals hold. Lemma \ref{BoundProbaFail} together with a union bound on the $4N(2N-1)$ tracked parameters and all confidence interval updates gives
$$
\Pb[\overline{\mathcal{A}}]\leq \delta.
$$

Let $\tau^U$ be the number of sampling steps after which the best pair is detected with probability at least $1-\delta$. The following holds:
\begin{eqnarray*}
\tau^{U} & \leq & \frac{64}{(\tilde{u}_{\mathrm{min}}\Delta_{23})^2}\log\left(\frac{1}{\delta}\right)+\frac{64}{(\tilde{u}_{\mathrm{min}}\Delta_{23})^2}\log\left(\frac{\pi^2N(M+1)}{3}\right)\\
&& +128\log_2\left(\log_2\left(\frac{1}{\tilde{u}_{\mathrm{min}}\Delta_{23}}\right)+2\right)
\end{eqnarray*}

We call bad samples the sampled pairs $(i,j)$ with $j>2$. The expected number of "bad" samples due to the failures of a confidence interval during any time window $w$ is bounded as
$$
\sum_{w=0}^{w_{\mathrm{max}}}2^{2^w}\Pb[\overline{\mathcal{B}}_w]\leq 2N(2N-1)(\log_2( \log_2 (2N(2N-1)\tau^U)) +2).
$$

Under the good event the confidence intervals hold, the number of sampling steps where column $j$ is sampled, $\tau_j$, $j>2$, is bounded as
$$
\tau_{j} \leq \frac{256}{(u_1\Delta_{2j})^2}\log\left(2N(2N-1)\tau^U\right)+\log_2\left(4\log(2N(2N-1)\tau^U))+1\right).
$$
In each of those sampling steps, column $j$ is sampled less than $2(2N-1)$ times.

Let $\tau_2$ be the number of samples to detect $1$ when sampling against $2$. The following holds
\begin{eqnarray*}
\tau_2 & \leq & 64\left(\sum_{i=3}^N\frac{1}{(u_2\Delta_{1,i})^2}\right)\log\left(\frac{1}{\delta}\right)\\
&& +128\left(\sum_{i=2}^N\frac{1}{(u_2\Delta_{1,i})^2}\left(\frac{1}{2}\log\left(\frac{\pi^22N(2N-1)}{3}\right)+\log_2\left(\log_2\left(\frac{1}{u_2\Delta_{1,i}}\right)+2\right)\right)\right).
\end{eqnarray*}

Let $\tau_1$ be the necessary number of samples to detect $2$ when sampling against $1$. The following holds
\begin{eqnarray*}
\tau_1 & \leq & 64\left(\sum_{i=3}^N\frac{1}{(u_1\Delta_{2,i})^2}\right)\log\left(\frac{1}{\delta}\right)\\
&& +128\left(\sum_{i=3}^N\frac{1}{(u_1\Delta_{2,i})^2}\left(\frac{1}{2}\log\left(\frac{\pi^22N(2N-1)}{3}\right)+\log_2\left(\log_2\left(\frac{1}{u_1\Delta_{2,i}}\right)+2\right)\right)\right).
\end{eqnarray*}
The inequality $u_1\Delta_{2,i}\leq u_2\Delta_{1,i}$ gives $\tau_2\leq\tau_1$. 

Under event $\mathcal{A}$, the total number of samples is less than $4\tau_1 +2(2N-1)\tau_{j}+ \sum_{w=0}^{w_{\mathrm{max}}}2^{2^w}\Pb[\overline{\mathcal{B}}_w]$, which gives the result. \hfill \(\Box\)

\subsubsection{Proof of the lower bound}

As in Appendix \ref{SRankElimAppendix}, we admit Assumption \ref{hyp:kl}. We prove the following theorem.

\begin{theorem}\label{LBRMono}
For any $\delta$-PAC algorithm:
\begin{equation}\label{eq:LBRMono1}
\E[\tau_{\delta}] \geq \frac{1}{2}\left( \sum_{i=2}^n \frac{1}{d\left(\theta_i\theta_1, (\theta_i + \Delta_{1,i}+\alpha)\theta_1\right)} \right)d(\delta,1-\delta)
\end{equation}
and
\begin{equation}\label{eq:2}
  \E[\tau_{\delta}] \geq \frac{1}{2}\left( \sum_{i=3}^n \frac{1}{\tilde{d}\left(\theta_i\theta_2, \theta_1\theta_2\right)} \right)d(\delta,1-\delta)
\end{equation}
where $\tilde{d}\left(u_iu_2, u_1u_2\right) = \min\{d\left(u_iu_2, u_1u_2\right),d\left(u_1u_2, u_iu_2\right)\}$.
\end{theorem}

We prove (\ref{eq:LBRMono1}) by considering the classes of alternative models: $$U_{\alpha} = \{\boldsymbol{u}_i^{\alpha}= (u_1,..,u_i',...,u_n) \text{ with } u_i^{\alpha} = u_i \pm (\Delta_i+\alpha)\mid i \in [2,N]\}.$$

Let $\mathcal{E}_{\mathcal{M}_{u}}$ denote the event that the algorithm outputs pair $(1,2)$. For any $\delta$-PAC algorithm, $\mathds{P}_{u}(\mathcal{E}_{\mathcal{M}_{u}})\geq1-\delta$ and $\mathds{P}_{\boldsymbol{u}_i^{\alpha}}(\mathcal{E}_{\mathcal{M}_{u}})\leq \delta$. This together with Lemma \ref{kaufmann} gives that the solution to the following linear program is a lower bound on the sample complexity of any $\delta-PAC$ algorithm:

\begin{align*}
\mathbf{LP}_\alpha: & \\
\text{ minimize } & \sum_{i<j} x_{i,j}\\
\text { subject to } & \sum_{j=1}^{i-1} x_{j,i}d\left(u_iu_j, u_i'u_j\right)+\sum_{j=i+1}^{2N} x_{i,j} d\left(u_iu_j, u_i^{\alpha}u_j\right) \geq d(\delta,1-\delta), \text{ }i \in \{2,\ldots,2N\}\\
\end{align*}

Let $x^* = (x^*_{i,j}: (i,j)\in [2N]\times [2N])$ be an optimal solution to $\mathbf{LP}_\alpha$, and let $\hat{x} = (\hat{x}_{i,j}: (i,j)\in [2N]\times [2N])$ be defined as follows
$$
\hat{x}_{i,j} = \left\{
\begin{array}{ll}
\hat{x}_{1,i} = \sum_{j=1}^{i-1} x^*_{j,i} + \sum_{j=i+1}^{N} x^*_{i,j} & \hbox{ for } i \in \{2,\dots,2N\}\\
\hat{x}_{l,i} = 0 & \hbox{ for } l>1,\ i \in \{2,\dots,2N\}\ \\
0 & \hbox{ otherwise}.
\end{array}
\right .
$$

$x^*$ and $\hat{x}$ satisfy the following equation
$$
\sum_{i,j} \hat{x}_{i,j} = 2 \sum_{i,j} x^*_{i,j}. 
$$

Thus, the solution of the following linear program is a lower bound on the sample complexity:

\begin{align*} 
\mathbf{LP}_{\alpha}': & \\
\text{ minimize } & \frac{1}{2}\sum_{i=2}^{n} x_{i,1}\\
\text { subject to } &  x_{i,1} \geq \frac{1}{d\left(u_iu_1, u_i^{\alpha}u_1\right) } d(\delta,1-\delta), \text{ }i \in \{2,\ldots,2N\}\\
\end{align*}

Equation (\ref{eq:2}) is obtained by considering the class of alternative models: $$u_{\text{switch}} = \{\boldsymbol{u}_i= (u_i,..,u_1,...,u_n) \mid i \in \{3,\ldots,2N\}\}.$$

Considering the event $\mathcal{E}_{\mathcal{M}_{u}} $ and Lemma \ref{kaufmann} , we get that the solution of the following linear program is a lower bound on the sample complexity of any $\delta-PAC$ algorithm:

\begin{align*} 
\mathbf{LP}_{\text{switch}}: & \\
\text{ minimize } & \sum_{i<j} x_{i,j}\\
\text { subject to } & \sum_{j=2, j\neq i}^{n}x_{1,j}d\left(u_1u_j, u_iu_j\right) +
\sum_{j=2, j\neq i}^{n}x_{j,i} d\left(u_i u_j, u_1u_j\right) \geq d(\delta,1-\delta), \text{ }i \in \{3,\ldots,2N\}\\
\end{align*}

Using the same technique, from an optimal solution $x^*$ we can build an alternative solution that satisfies the constraints.

\begin{align*}
     \forall i \in \{1\}\cup[3,N]:\text{ } &
    \hat{x}_{2,i} = \sum_{j=2}^{i-1} x_{j,i}  +\sum_{j=i+1}^{N} x_{j,i}\\
    &\hat{x}_{i,j} = 0, j \in \{3,\ldots,2N\}\\
\end{align*}

As before, we also have
$$
\sum_{i,j} \hat{x}_{i,j} = 2 \sum_{i,j} x^*_{i,j}.
$$

Thus, the optimal solution to the following linear program is a lower bound on the sampling complexity:
\begin{align*} 
\text{ minimize } & \frac{1}{2}\left(x_{1,2}+\sum_{i} x_{2,i}\right)\\
\text { subject to } & x_{i,2} d\left(u_iu_2, u_1u_2\right) +x_{1,2}d\left(u_1u_2, u_iu_2\right)\geq d(\delta,1-\delta), \text{ }i \in \{3,\ldots,2N\}\\
\end{align*}

which gives Equation~(\ref{eq:2}).

As in Appendix \ref{SRankElimAppendix}, specific lower bounds can be obtained for Bernoulli and unit-variance Gaussian random variables.

\section{\IBLearning\ algorithm}\label{IBLearningAppendix}

\subsection{Algorithm description and pseudo-code}

\begin{algorithm}[H]
\caption{\RMono\ }
\label{alg:SRankElim}
\begin{algorithmic}
\STATE {\bfseries Input:} set of items $U$, precision $\delta$ 
\STATE Detect items $1,2$ with \RMono
\STATE Sample unranked items against items $1,2$ 
\end{algorithmic}
\end{algorithm}

\subsection{Proof of the upper bound} We prove the following theorem.

\PairSelect*

The proof goes as follows. Recall the following definitions:
$$
\Delta_{2,i} = u_{2i}-u_{2i+1}~~\text{and} ~~  \Delta_{2i-1} = u_{2i-2}-u_{2i-1}.
$$
with the convention $u_{2N+1}=u_{2N-2}$.

During the second phase of the algorithm, at least half of the gathered samples involving item $i$ are pairs $(1,i)$. Repeating the computations used to obtain Equation~(\ref{eqn:tau_u}), the number of samples gathered during the second phase of the algorithm, $\tau_m$, is bounded as:
\begin{eqnarray*}
\tau_m & \leq & 64\left(\sum_{i=3}^N\frac{1}{(u_1\Delta_{i})^2}\right)\log\left(\frac{1}{\delta}\right)\\
&& +128\left(\sum_{i=3}^N\frac{1}{(u_1\Delta_{i})^2}\left(\frac{1}{2}\log\left(\frac{\pi^22N(2N-1)}{3}\right)+\log_2\left(\log_2\left(\frac{1}{u_1\Delta_{i}}\right)+2\right)\right)\right).
\end{eqnarray*}

The upper bound on the number of samples to detect $1$ and $2$, is the same as the upper bound on the sample complexity of \RMono. Noticing that $u_1\Delta_{i}\leq u_1\Delta_{2,i}$ and summing the bounds on the number of samples during each of the two phase completes the proof. \hfill \(\Box\)

\subsection{Proof of the lower bound}

Note that the lower bound of Theorem~\ref{LBRMono} still holds. 

To ease the notations, we will note 
$$
u_{2i} \pm \Delta_{2,i} = u_{2i+1}~~\text{and} ~~  u_{2i-1}\pm \Delta_{2i-1} = u_{2i-2}.
$$ 

\begin{theorem}
For any $\delta$-PAC algorithm, we have
\begin{equation}\label{eq:LBMono}
\E[\tau_{\delta}] \geq \frac{1}{2}\left( \sum_{i=2}^n \frac{1}{d\left(u_iu_1, (u_i \pm \Delta_i)u_1\right)} \right)d(\delta,1-\delta)
\end{equation}
\end{theorem}

We prove the theorem by considering the following classes of alternative models: $$u_{\alpha} = \{\boldsymbol{u}_i^{\alpha}= (u_1,..,u_i',...,u_{2N}) \hbox{ with } u_i^{\alpha} = u_i \pm (\Delta_i+\alpha) \mid  i \in \{2,\ldots,2N\}\}.$$

Let $\mathcal{E}_{\mathcal{M}_{u}}$ denote the event that the algorithm returns the optimal matching. For any $\delta$-PAC algorithm, $\mathds{P}_{u}(\mathcal{E}_{\mathcal{M}_{u}})\geq 1-\delta$ and $\mathds{P}_{\boldsymbol{u}_i^{\alpha}}(\mathcal{E}_{\mathcal{M}_{u}})\leq \delta$. Using this with \ref{kaufmann} we obtain that the solution to the following linear program is a lower bound on the sample complexity of any $\delta$-PAC algorithm:

\begin{align*}
\mathbf{LP}_{\alpha}: & \\
\text{ minimize } & \sum_{i<j} x_{i,j}\\
\text { subject to } & \sum_{j=1}^{i-1} x_{j,i}d\left(u_iu_j, u_i'u_j\right)+\sum_{j=i+1}^{2N} x_{i,j} d\left(u_iu_j, u_i^{\alpha}u_j\right) \geq d(\delta,1-\delta), \text{ }i \in \{2,\ldots,2N\}\\
\end{align*}

Let $x^* = (x^*_{i,j})$ be an optimal solution to $\mathbf{LP}_{\alpha}$.

We can build an alternative solution satisfying the constraints of the linear program:
\begin{align*}
     \forall i \in \{2,\ldots,2N\}: \text{ } &\hat{x}_{1,i} = \sum_{j=1}^{i-1} x^*_{j,i} + \sum_{j=i+1}^{2N} x^*_{i,j}  \\
    &\hat{x}_{l,i} = 0, \forall l>1.
\end{align*}

We have
$$
\sum_{i,j} \hat{x}_{i,j} = 2 \sum_{i,j} x^*_{i,j}.
$$

Thus, the solution of the following linear program is a lower bound on the sample complexity:

\begin{align*} 
\mathbf{LP}_{\alpha}' & \\
\text{ minimize } & \frac{1}{2}\sum_{i=2}^{2N} x_{i,1}\\
\text { subject to } &  x_{i,1} \geq \frac{1}{d\left(u_iu_1, u_i^{\alpha}u_1\right) } d(\delta,1-\delta), \text{ }i \in \{2,\dots,2N\}\\
\end{align*}

This lower bound holds for any $\alpha >0$, so we can let $\alpha$ go to zero and get Equation~(\ref{eq:1}).

As in Appendix \ref{SRankElimAppendix}, specific lower bounds can be obtained for Bernoulli and unit-variance Gaussian random variables.

\section{\SMDC\ algorithm}\label{SMDCAppendix}
\subsection{Algorithm description and pseudo-code}

\begin{algorithm}
\caption{\SMDC}
\begin{algorithmic}
\STATE {\bfseries Input:} set of items $[2N]$ and horizon $T$
\STATE $t=0, C=X=\tilde{C}=\tilde{X}=[0]^{2N\times 2N}, \mathfrak{S}=\{[2N]\}$
\FOR{$t=1\ldots T$}
    \STATE $m_t \gets$ {\tt sample\_matching}$(S,t)$
    \FOR{$(i,j)\in m_t$}
        \STATE $\tilde{X}(i,j) \gets \tilde{X}(i,j) + x_{i,j,t}$
        \STATE $\tilde{C}(i,j) \gets \tilde{C}(i,j) + 1$
    \ENDFOR
    \FOR{$S \in \mathfrak{S}$} 
        \IF{$\exists i \in S \text{ s.t. }\sum_{j\in S}\tilde{C}(i,j)=|S|-1$}
            \STATE // the notation $X([S],:)$ is a vectorized numpy index notation
            \STATE $X([S],:),C([S],:) += \tilde{X}([S],:),\tilde{C}([S],:)$
            \STATE $\tilde{X}([S],:),\tilde{C}([S],:)=0$
        \ENDIF
    \ENDFOR 
    \STATE $Q_{+},Q_{-}  \gets$ {\tt confidence\_bound}$(X,C,T,Q_{+},Q_{-},S)$
    \FOR{$S \in \mathfrak{S}$}
        \STATE Order items in $s$ according to $Q_{+}$
        \FOR{$i \in \{2,\dots,|S|\}$} 
            \IF{$Q_{+}[i]<Q_{-}[i-1]$}
            \STATE Split $S$ between $i$ and $i-1$
            \ENDIF
        \ENDFOR
    \ENDFOR
\ENDFOR
\end{algorithmic}
\end{algorithm}

The items are split into even clusters with high probability. The {\tt sample\_matching} procedure samples the items in the clusters following a round-robin tournament. For a cluster $S$, the round-robin tournaments guarantees that each item has been matched once with any other item in the cluster each $|S|-1$ iterations.

The goal of the {\tt sample\_matching} procedure is to ensure that at every iteration, any two items $i,i'$ in the same cluster $S$ have been matched the same number of times with any other item.  This means that, for any item $j \in [2N]\setminus\{i,i'\}$, $C(i,j)=C(i',j)$, which guarantees: 
\begin{equation}\label{eqn:samenumsapmle}
\frac{\E[X(i,j)-X(i',j)]}{C(i,j)}=u_j\Delta_{i,i'}.
\end{equation}
This implies that it is possible to compare to items within the same cluster by comparing the total reward received for both of those items.

We define an ordered sequence of clusters as a sequence $S_k,\ldots,S_{k'}$ such that, for any $l\in \{k,\ldots,k'-1\}$
$$
\forall (i,j) \in S_l \times S_{l+1}: \ i<j.
$$
At every iteration, $\mathfrak{S}$ is an ordered sequence of clusters. We note $\mathfrak{S}[k]$ the $k^{th}$ cluster of the sequence.

The function {\tt confidence\_bound} computes confidence bounds for the total reward per item when matched with an item within the same cluster or in a lower ranked cluster. Consider any item $i \in [2N]$ and $k$ such that $i\in \mathfrak{S}[k]$. If $\sum_{j=1}^{2N}C(i,j)\mathds{1}_{\{j \in \mathfrak{S}[k']|k'\leq k\}}=k_l$ for some $l$, then
$$
Q_{-}(i)= \frac{\sum_{j=1}^{2N}X(i,j)\mathds{1}_{\{j \in \mathfrak{S}[k']|k'\leq k\}}}{k_l}-\sqrt{\frac{\log(T)}{k_l}}
$$
and
$$ Q_{+}(i)= \frac{\sum_{j=1}^{2N}X(i,j)\mathds{1}_{\{j \in \mathfrak{S}[k']|k'\leq k\}}}{k_l}+\sqrt{\frac{\log(T)}{k_l}}.
$$

Equation \ref{eqn:samenumsapmle} implies that for any two items $i,i'$ ins the same cluster $\mathfrak{S}[k]$:

\begin{align*}
   \E[\frac{\sum_{j=1}^{2N}(X(i,j)-X(i,j))\mathds{1}_{\{j \in S[k']|k'\leq k\}}}{k_l}]&=\Delta_{i,i'}\frac{\sum_{j=1}^{2N}u_jC(i,j)\mathds{1}_{\{j \in S[k']|k'\leq k\}}}{k_l}\\
   &\geq \Delta_{i,i'}\frac{1}{|\mathfrak{S}[k]|-1}\sum_{j\in \mathfrak{S}[k]\setminus\{i,i'\}}u_j \numberthis \label{eqn:meancompare}
\end{align*}
\subsection{Proof of the upper bound} We prove the following theorem.
\smds*
The proof goes as follows. The \SMDC\ algorithm tracks the expected rewards per item. Lemma \ref{BoundProbaFail} together with a union bound on the $2N$ parameters and the confidence interval updates implies
$$
\E[R(T)] \leq  \E[R(T) \mid \mathcal{A}] +T\Pb(\bar{\mathcal{A}})
\leq \sum_{i=1}^{N}\sum_{j=i+1}^{N}R_{i,j}(T)+4N.
$$

We call inter-pair match between pair $i$ and pair $j$ the match of an item of pair $i$ with an item of pair $j$. The per iteration average regret of a tournament on a cluster $S$ can be decomposed into the sum of regrets of inter-pair matches. 




\begin{lemma}\label{isolated} Assume that items in $S=\{2k-1,\ldots,2l\}$, for $1\leq k < l \leq K$, such that $|S| = 2K$, are matched according to a round-robin tournament, so that every pair $(i,j)$ with $i\neq j$ is matched exactly once over $2K-1$ iterations. Then, the expected average regret, denoted as $R_{S}$, is given as
$$
R_{S} = \frac{1}{2K-1}\sum_{i=1}^{K} \sum_{j=i+1}^{K} r_{i,j}
$$
where
$$
r_{i,j} = (u_{2i}-u_{2j-1})(u_{2i-1}-u_{2j})+(u_{2i-1}-u_{2j-1})(u_{2i}-u_{2j}).
$$
\end{lemma}
This proof of Lemma~\ref{isolated} is given in Appendix~\ref{app:isolated}.

\begin{lemma}\label{ns1} The total number of sampled matchings $T_S$ before the set of items $[2K]$ is separated is upper bounded as
$$
T_S \leq \frac{64}{(\tilde{\mu}_{-\{k,k+1\}}\Delta_{k,k+1})^2}\log(T) +2K-1
$$
for all $k\in [2K-1], $where
\begin{equation}
\tilde{\mu}_{-\{k,k+1\}} = \frac{1}{2K-1}\left(\sum_{i=1}^{2K}u_i-u_{k}-u_{k+1}\right).
\label{equ:tildemu}
\end{equation}
\end{lemma}
The last lemma follows from Lemma \ref{maxsample} and equation \ref{eqn:meancompare}.

Let $R_{i,j}(T)$ denote the total regret incurred due to interactions between pairs $i$ and $j$, written
\begin{eqnarray*}
R_{i,j}(T) &=& \sum_{t=1}^T \sum_{\{k,l\} \in \{2i-1,2i\}\times\{2j-1,2j\}}\mathds{1}_{\{(k,l)\in m_t\}} \left(\frac{1}{2}(u_{2i-1}u_{2i}+u_{2j-1}u_{2j})-u_{l}u_{k}\right).
\end{eqnarray*}

Let $\mathcal{E}_{i,j}(S)$ denote the event that all inter-pair matches between pairs $i$ and $j$ are sampled before $S$ is separated. Note that 
$$
\E[R_{i,j}(T)\mid \mathcal{E}_{i,j}(S)] \leq \frac{T_S}{|S|-1}r_{i,j}
$$
which with some computations yield the following lemma.

\begin{lemma}\label{pairwiseRbound} Assume that all inter-pair matches between pairs $i$ and $j$ are sampled before $S$ is separated, for a set $S$ such that $|S|=2K$ and $K\geq 3$. Then under event $\mathcal{A}$, the total regret due to inter-pair matches between pairs $i$ and $j$ is upper bounded as
$$
\E[R_{i,j}(T)\mid \mathcal{E}_{i,j}(S)]  \leq \frac{640}{K-2}\frac{1}{\Delta_{\mathrm{min}}}\log(T)+2.
$$
If $K=2$, the bound is
$$
\E[R_{i,j}(T)\mid \mathcal{E}_{i,j}(S)] \leq 384\frac{1}{\Delta_{\mathrm{min}}}\log(T)+2.
$$
\end{lemma}
The proof of Lemma~\ref{pairwiseRbound} is given in Appendix~\ref{app:pairwiseRbound}.

The total regret incurred for inter-pair matches between pairs $i$ and $j$ is bounded as
$$
\E[R_{i,j}(T)]\leq \max_{S\subseteq [2N]: (i,j)\in S}\E[R_{i,j}(T)\mid \mathcal{E}_{i,j}(S)]. 
$$

Adjacent pairs can interact in a cluster $S$ with $|S|=2K$ and either $2<K\leq N$ or $K=2$. The total regret for interactions between pair $i$ and pair $i+1$ is thus upper bounded by Lemma~\ref{pairwiseRbound} as

\begin{align*}
R_{i,i+1}(T)& \leq \max_{S\subset[2N]: (i,i+1)\in S}\E[R_{i,i+1}(T)\mid \mathcal{E}_{i,i+1}(S)] \\
&\leq \max\left\{384, \max_{2<K\leq N}\frac{640}{K-2}\right\}\frac{\log(T)}{\Delta_{\mathrm{min}}}+2\\
&\leq \frac{640}{\Delta_{\mathrm{min}}}\log(T)+2.
\end{align*}

All non-adjacent items can only $i$ and $j$ interact in a cluster s.t. $K\geq j-i+1\geq 3$. Thus, Lemma \ref{pairwiseRbound} gives
$$
R_{i,j}(T) \leq \frac{640}{(j-i-1)\Delta_{\mathrm{min}}}\log(T)+2.
$$
Putting the pieces together, we have
\begin{eqnarray*}
R(T) & \leq & \left(N+
    \sum_{i=1}^{N}\sum_{j=i+2}^{N}\frac{1}{j-i-1}\right)\frac{640\log(T)}{\Delta_{\mathrm{min}}}+N^2\\
    & \leq & \left(N+
    \sum_{i=1}^{N}\sum_{k=1}^{N-i-1}\frac{1}{k}\right)\frac{640\log(T)}{\Delta_{\mathrm{min}}}+N^2\\
    & \leq & \left(2N+
    \sum_{i=1}^{N}\log(N-i-1)\right)\frac{640\log(T)}{\Delta_{\mathrm{min}}}+N^2\\
    & \leq & 640\frac{N(\log(N)+2)}{\Delta_{\mathrm{min}}}\log(T)+N^2.
\end{eqnarray*}

\subsubsection{Proof of Lemma~\ref{isolated}}
\label{app:isolated}

The items in set $S=\{2k-1,\ldots,2l\}$ are matched according to a round-robin tournament. This means that each pair $(i,j)$ such that $i\neq j$ is matched exactly once over $2K-1$ iterations. Therefore, the total expected reward over $2K-1$ iterations is:
$$
\sum_{i=2k-1}^{2l}\sum_{j=i+1}^{2l} u_iu_{j}.
$$

And the optimal way to match items in set $S$ gives the expected reward
$$
\sum_{i=k}^l u_{2i-1}u_{2i}.
$$
It follows that the expected regret per iteration over $2K-1$ iterations is
\begin{align*}
    R_S &= \sum_{i=k}^l u_{2i-1}u_{2i} -\frac{1}{2(2K-1)} \sum_{i=2k-1}^{2l}\sum_{j=2k-1, j\neq i}^{2l} u_iu_{j}\\
    &= \frac{1}{2(2K-1)}\sum_{i=l}^{p}\sum_{j=k, j\neq i}^{l}[u_{2i-1}(u_{2i}-u_{2j-1})+u_{2i-1}(u_{2i}-u_{2j})\\
    &\qquad \qquad \qquad \qquad \qquad+u_{2i}(u_{2i-1}-u_{2j-1})+u_{2i}(u_{2i-1}-u_{2j})]\\
    &=\frac{1}{2K+1}\sum_{i=k}^{l} \sum_{j=i+1}^{K-1} r_{i,j}.
\end{align*}

\subsubsection{Proof of Lemma~\ref{pairwiseRbound}}
\label{app:pairwiseRbound}

We first compute the bound for a cluster $S$ of size $2K$ where $K > 2$, and then consider the case $K=2$. Without loss of generality, we enumerate the items as $1, \ldots, 2K$.

\paragraph{Case: $K\geq3$} Consider any two pairs $i$ and $j$ cluster $S$.  The total regret for interactions between the two pairs until time $T_S$ is bounded as follows:
\begin{align*}
    \E[R_{i,j}(T)\mid \mathcal{E}_{i,j}(S)]&= \sum_{t=1}^{T_S} \sum_{\{k,l\} \in \{2i-1,2i\}\times\{2j-1,2j\}}\mathds{1}_{\{\{u_k,u_l\}\in m_t\}}\left(\frac{1}{2}(u_{2i-1}u_{2i}+u_{2j-1}u_{2j})-u_{l}u_{k}\right)\\
    &\leq \frac{T_S}{2K-1}r_{i,j}\\
    &\leq \frac{T_S}{2K-1}r_{1,K}\\
    &\leq \frac{T_S}{2K-1}\left[(u_1-u_{2K})(u_2-u_{2K-1})+(u_1-u_{2K-1})(u_2-u_{2K})\right]\\
    &\leq \frac{T_S}{2K-1}2\Delta_{2,2K-1}^2
\end{align*}

where the last inequality holds by Assumption~\ref{hyp:even}.

Let $m= \argmax \{\Delta_{k,k+1} : 1 < k \leq 2K-1\}$ and $\gamma = \min_{i \in [K-2]} \Delta_{2i,2i+1}$. Note that 
$$
\Delta_{2,2K-1}\leq (K-1)\Delta_{m,m+1} \hbox{ and } \Delta_{\mathrm{min}} \leq \gamma^2
$$
and
$$
u_k \geq u_{2N}+\sum_{i=\lceil k/2\rceil}^{N-1}\Delta_{2i,2i+1}.
$$

Recall the definition of $\mu_{-m,m+1}$ in Equation (\ref{equ:tildemu}). We have 
\begin{align*}
    (2K-1)\tilde{\mu}_{-\{m,m+1\}}&= \sum_{k=1}^{2K}u_k-(u_{m}+u_{m+1})\\
    &\geq \sum_{k=3}^{2K}u_k\geq (2K-2)u_{2K}+\sum_{k=3}^{2K}\sum_{i=\lceil\frac{k}{2}\rceil}^{N-1}\gamma\\
    &\geq (2K-2)u_{2K-1}+(K-1)(K-2)\gamma\\
    & \geq (K-1)(K-2)\gamma
\end{align*}

Combining with Lemma~\ref{ns1}, we have
\begin{align*}
    T_S \leq \frac{64(2K-1)^2}{(K-1)^2(K-2)^2\gamma^2\Delta_{m,m+1}^2}\log(T)+2K-1.
\end{align*}


Therefore, we have
\begin{align*}
    R_{i,j}(T) &\leq 128\frac{(K-1)^2\Delta_{m,m+1}^2(2K-1)^2}{(2K-1)(K-1)^2(K-2)^2\gamma^2\Delta_{m,m+1}^2}\log(T)+2\\
    &\leq 128\frac{(2K-1)}{(K-2)^2\gamma^2}\log(T)+2\\
    &\leq \frac{640}{(K-2)\Delta_{\mathrm{min}}}\log(T)+2
\end{align*}

which completes the proof.

\paragraph{Case: $K=2$} Under Assumption~\ref{hyp:even}, the regret can be bounded as follows
\begin{align*}
    R_{0,1}(T) &\leq 64 \frac{2\cdot 3\log(T)(u_2-u_3)^2}{(u_1+u_4)^2(u_1-u_2)^2}+2\\
    &\leq \frac{384\log(T)}{(u_1+u_4)^2}+2\leq \frac{384\log(T)}{(u_2-u_3)^2}+2\\
    &\leq \frac{384\log(T)}{\Delta_{\mathrm{min}}}+2.
\end{align*}

\section{\MDC\ algorithm}\label{MDCAppendix}

\subsection{Pseudo-code}

The main difficulty when removing Assumption \ref{hyp:even}, is that there is no reason to believe the created clusters will contain an even number of items. For instance, set of items $\{1,\dots,6\}$ may be split into $\{1,2,3\},\{4,5,6\}$. In that case, items can no longer be matched within the cluster they belong to alone. To tackle this difficulty, we introduce the \CMSampling\ procedure. As in the simplified setting, this procedure guarantees that two items within a cluster are matched the same number of time with any other item. 

The rest of the algorithm is similar to the \SMDC algorithm.

\begin{algorithm}[hbt!]
\caption{\MDC}
\begin{algorithmic}\label{MDC}
\STATE {\bfseries Input:} set of items $[2N]$ and horizon $T$
\STATE $t=0, C=X=\tilde{C}=\tilde{X}=[0]^{2N\times 2N}, \mathfrak{S}=\{[2N]\}$
\FOR{$t=1\ldots T$}
    \STATE $m_t \gets$ \CMSampling$(\mathfrak{S},t)$
    \FOR{$(i,j)\in m_t$}
        \STATE $\tilde{X}(i,j) \gets \tilde{X}(i,j) + x_{i,j,t}$
        \STATE $\tilde{C}(i,j) \gets \tilde{C}(i,j) + 1$
        \STATE $\tilde{X}(j,i) \gets \tilde{X}(j,i) + x_{i,j,t}$
        \STATE $\tilde{C}(j,i) \gets \tilde{C}(j,i) + 1$
    \ENDFOR
    \FOR{$S \in \mathfrak{S}$} 
        \STATE // $r(S)$and $L(S)$ defined in the description of \CMSampling
        \IF{$\exists i \in S \text{ s.t. }\sum_{j\in S}\tilde{C}(i,j)=r(S)|L(S)|$}
            \STATE $X([S],:),C([S],:) += \tilde{X}([S],:),\tilde{C}([S],:)$ 
            \STATE $\tilde{X}([S],:),\tilde{C}([S],:)=0$
        \ENDIF
    \ENDFOR 
    \STATE $Q_{+},Q_{-}  \gets$ {\tt confidence\_bound}$(X,C,T,Q_{+},Q_{-},\mathfrak{S})$
    \FOR{$S \in \mathfrak{S}$}
        \STATE Order items in $S$ according to $Q_{+}$
        \FOR{$i \in [2,|s|]$} 
            \IF{$Q_{+}[s(i)]<Q_{-}[s(i-1)]$} 
                \STATE Split $S$ between $S(i)$ and $S(i-1)$
                \IF{$S$ is part of a chain of clusters $H$}
                    \STATE $\tilde{X}([H],:),\tilde{C}([H],:)=0$
                \ENDIF
            \ENDIF
        \ENDFOR
    \ENDFOR
\ENDFOR
\end{algorithmic}
\end{algorithm}

As in \SMDC, the function {\tt confidence\_bound} computes confidence bounds for the total reward per item when matched with an item within the same cluster or in a lower ranked cluster. Consider any item $i \in [2N]$ and $k$ such that $i\in \mathfrak{S}[k]$. If $\sum_{j=1}^{2N}C(i,j)\mathds{1}_{\{j \in \mathfrak{S}[k']|k'\leq k\}}=k_l$ for some $l$, then
$$
Q_{-}(i)= \frac{\sum_{j=1}^{2N}X(i,j)\mathds{1}_{\{j \in \mathfrak{S}[k']|k'\leq k\}}}{k_l}-\sqrt{\frac{\log(T)}{k_l}}
$$
and
$$ Q_{+}(i)= \frac{\sum_{j=1}^{2N}X(i,j)\mathds{1}_{\{j \in \mathfrak{S}[k']|k'\leq k\}}}{k_l}+\sqrt{\frac{\log(T)}{k_l}}.
$$

The main difference between the \SMDC\ and the \MDC\ algorithms is the \CMSampling\ procedure, which is detailed in subsection \ref{cmsampling}.

\subsection{Cluster properties}\label{Terminology}

Before presenting the \CMSampling\ procedure and the analysis of the algorithm, we introduce some needed terminology.

Recall that an ordered sequence of clusters is a sequence $S_k,\ldots,S_{k'}$ such that, for any $l\in \{k,\ldots,k'-1\}$
$$
\forall (i,j) \in S_l \times S_{l+1}: \ i<j.
$$
In the optimal matching, the items are matched in decreasing order. Hence, only the lowest and highest ranked items of a cluster $S_l$ can be optimally matched outside the cluster.

\begin{itemize}
\item {\bf Isolated cluster}: A cluster is said to be an \emph{isolated cluster} if all items in this cluster have their optimal matching partners within this cluster.

\item {\bf Chain of clusters}: A \emph{chain of clusters} is an ordered sequence of clusters such that every cluster in the chain contains an item whose optimal match is in the following cluster (except for the last one). Also, each item and its optimal matching partner either belong to the same cluster or to adjacent clusters in the chain.

\item {\bf Trivial chain of clusters}: A chain is referred to be a \emph{trivial chain} if it is of the form  $\{2k-1\},\{2k,\ldots,2l-1\},\{2l\}$, for any $1\leq k<l\leq N$.
\end{itemize}

Note that if cluster $S_i$ and cluster $S_{i+1}$ are adjacent clusters in a chain, the highest ranked item of cluster $S_i$ is optimally matched with the lowest ranked item of cluster $S_{i+1}$. 

Note that within any cluster $S$, the items are not sorted in monotonic order. Therefore, knowing the highest -or lowest- ranked item of $S$ does not provide any information about which item it should be matched with, as it could be any of the other un-ranked items. Also, the sampling policy and confidence interval's updates remain unchanged after splitting $S$ between the top -or bottom- item and the rest of the items. Thus, for simplicity, even if the highest or the lowest element of the cluster (alone) is discovered, in the analysis, we consider that cluster $S$ is not yet split.
\subsection{\CMSampling\ procedure}\label{cmsampling}

At a high level, \CMSampling\ is a procedure that constructs a list of matchings $\tilde{\mathcal{M}}$, then iteratively samples each matching in this ensemble. For each pair $(i,j)$, we denote with $p_{i,j}$ the proportion of draws of pair $(i,j)$ in the list $\tilde{\mathcal{M}}$:
$$
p_{i,j} =\frac{1}{|\tilde{\mathcal{M}}|} \sum_{m \in \mathcal{M}} \mathds{1}_{\{(i,j)\in \mathcal{M}\}}
$$
\paragraph{Sampling guarantees}The \CMSampling\ procedure guarantees the following properties:
\begin{itemize}
\item Items residing in an isolated cluster $S$ are uniformly matched among themselves:
$$
p_{i,j} =\frac{1}{|S|-1},\ \forall (i,j) \in S^2,\ i\neq j. 
$$
\item For every chain of clusters $S_k,\ldots,S_{k'}$, items residing within each cluster $S_i$ of the chain, are uniformly matched among themselves
$$
p_{i,j} =\frac{|S_l|-2}{|S_l|(|S_l|-1)},\ \forall (i,j) \in S_l^2, \ i\neq j, l \in \{k+1,\dots,k'-1\},
$$
$$
p_{i,j} =\frac{1}{|S_l|},\ \forall (i,j) \in S_l^2, \ i\neq j, l=k \text{ or } l=k'.
$$
\item All potential pairs $(i,j) \in S_{l}\times S_{l+1}$, for $l\in\{k,\dots,k'-1\}$, are sampled uniformly;
$$
p_{i,j} =\frac{1}{|S_l||S_{l+1}|},\ \forall (i,j) \in S_{l}\times S_{l+1}, l\in[k,k'-1].
$$
\end{itemize}

Let $\nu(S)$ be the sampling frequency of cluster $S$, which has values as follows: 

\begin{equation}\label{nuS}
    \nu(S) = \left\{
\begin{array}{ll}
|S|-1 & \hbox{ if } S \hbox{ is an isolated cluster}\\
|S| & \hbox{ if } S \hbox{ is the first or the last cluster of a chain}\\
|S|+1 & \hbox{ if } S \hbox{ is an intermediate cluster of a chain}.
\end{array}
\right .
\end{equation}

Let $i$ and $i'$ be to items in cluster $\mathfrak{S}[k]$. If $\mathfrak{S}[k]$ is an isolated or the first cluster of a chain, then:
\begin{align*}
   \E[\frac{\sum_{j=1}^{2N}(X(i,j)-X(i,j))\mathds{1}_{\{j \in \mathfrak{S}[k']|k'\leq k\}}}{k_l}]
   &\geq \Delta_{i,i'}\frac{1}{\nu(\mathfrak{S}[k])}\sum_{j\in \mathfrak{S}[k]\setminus\{i,i'\}}u_j \numberthis \label{eqn:meancompare2}
\end{align*}

If $\mathfrak{S}[k]$ is an within or the last cluster of a chain, then:
\begin{align*}
   \E[\frac{\sum_{j=1}^{2N}(X(i,j)-X(i,j))\mathds{1}_{\{j \in \mathfrak{S}[k']|k'\leq k\}}}{k_l}]
   &\geq \Delta_{i,i'}\frac{1}{\nu(\mathfrak{S}[k])}\left[\sum_{j\in \mathfrak{S}[k]\setminus\{i,i'\}}u_j +\frac{1}{|\mathfrak{S}[k-1]|}\sum_{j\in \mathfrak{S}[k-1]}u_j \right]\numberthis \label{eqn:meancompare3}
\end{align*}

Note that in the case of isolated clusters, equations (\ref{eqn:meancompare}) and (\ref{eqn:meancompare2}) are equivalent.

\paragraph{General procedure} The  \CMSampling\ procedure constructs ensemble $\tilde{\mathcal{M}}$ as follows. Each cluster $S= [k,l]$ is associated with a list of \textit{matching schemes} $L(S) =(l_1, \dots, l_r)$. Each \textit{matching schemes}  $l_i$ indicates which items should be paired together. 
For example, $l_i = \{(a,b), \ldots, (y,z)\}$ indicates that items $a$ and $b$, and items $y$ and $z$ should be matched together. 
If $S$ is part of a chain, \textit{matching schemes} $l_i$ may contain at most one pair $(a, \text{next})$ to indicate that item $i$ should be matched with an item of the next cluster. It may also contain at most one pair $(a, \text{previous})$) to indicate a match with the previous cluster.  
Constructing the list of \textit{matching schemes} therefore depends on the type of the cluster: whether the cluster is an isolated cluster, the first, an intermediate, or the last cluster of the chain. 

Each cluster $S$ is associated with sampling rate $r_S$, that depends on the position of the cluster in the chain. For each cluster $S$, the \textit{matching schemes}  in $L_S$ are chosen according to a deterministic round-robin schedule, with each matching being chosen $r_S$ times before moving on to the next one. At iteration $t$, the \CMSampling\ procedure builds a full matching by aggregating the \textit{matching schemes}  chosen for individual clusters.

\paragraph{Construction of a list of \textit{matching schemes} }

We discuss separately how lists of \textit{matching schemes}  are constructed for each type of the cluster. 

\begin{itemize}
    \item \textbf{Isolated cluster:} The list of \textit{matching schemes}  $L(S)$ is the scheduling list of an all-meet-all tournament over items in $S$.
    
    Note that in this case we have $|L(S)|=|S|-1$.
    
    \item \textbf{First or last cluster in a chain:} A virtual item "next" or "previous" is added to cluster $S$, and, then, $L(S)$ is built using the isolated clusters' method.
    
    Note that in this case we have $|L(S)|=|S|$.
    
    \item \textbf{Intermediate cluster in a chain:} An auxiliary list of matchings $\tilde{L}(S)=(l_1,\ldots,l_{|S|-1})$ is built using the isolated clusters' method. The list $L(S)$ is iteratively constructed from $\tilde{L}(S)$ using Algorithm~\ref{interlist}. 
    
    Note that $|L(S)|=(|S|-1)|S|$. For each item $a$ in $S$, pairs $(a,\text{prev})$ and $(a,\text{next})$ each appear in $|S|-1$ \textit{matching schemes} . Each pair $(a,b)\in S^2,\ a\neq b$ appears in $|S|-2$ \textit{matching schemes}. Also note that the \textit{matching schemes}  in $L(S)$ are ordered so that all pairs $(a,\text{prev})$ and $(a,\text{next})$, $a \in S$, appear once in each sub-list $(l_{n|S|+1},\dots,l_{(n+1)|S|})$, $n \in \{0,\dots,|S|-2\}$.
\end{itemize}

\begin{algorithm}[hbt!]
\caption{InterList}
\begin{algorithmic}\label{interlist}
\STATE $L(S)=\emptyset$
\STATE $\tilde{L}(S)=(l_1,\dots,l_{|S|-1})$
\FOR{$l_i \in \tilde{L}(S)$}
    \FOR{$(a,b) \in l_i$}
    \STATE $l_i \leftarrow l_i\setminus(a,b)$
    \STATE $L(S)\leftarrow L(S)\cup \left(l_i,(a,\text{next}),(b,\text{prev})\right)$
    \STATE $L(S)\leftarrow L(S)\cup \left(l_i,(b,\text{next}),(a,\text{prev})\right)$
    \ENDFOR
\ENDFOR
\end{algorithmic}
\end{algorithm}

\paragraph{Sampling rates and synchronization:}

If $(a,\text{next})$ and $(\text{prev},b)$ appear in the matchings of adjacent clusters $S_1$ and $S_2$, pair $(a,b)$ is sampled. A synchronization method is used in \CMSampling\ to ensure all pairs $(a,b)\in S_1 \times S_2$ are sampled uniformly.

Note that their is no need for such a synchronization for an isolated cluster $S$: the sampling rate $r_S$ is set to $1$, and \CMSampling\ iterates over $L(S)$. Note that \CMSampling\ and {\tt sample\_matching} are equivalent on an isolated cluster.

The following paragraph explains, without loss of generality, the synchronisation method of \CMSampling\  for specific chain $S_1,\ldots,S_K$. 

Each cluster $S_i$ in the chain is associated with a sampling rate $r_{S_i}$ that depends on its position in the chain and the size of the adjacent clusters.
\begin{itemize}
    \item If $i$ is odd, $r_{S_i}=1$.
    \item If $i$ is even, $r_{S_i} =\mathrm{LCM}(|S_{i-1}|,|S_{i+1}|)$, with the convention $|S_{K+1}|=1$, and where $\mathrm{LCM}(a,b)$ denotes the least common multiplier of $a$ and $b$.
\end{itemize}

\CMSampling\ selects the same \textit{matching schemes} for cluster $S$, $r_S$ times, before switching to the next one. Suppose \CMSampling\ starts on the chain $S_1,\ldots,S_K$ at iteration $t_{\text{init}}$. The following properties hold:
\begin{itemize}
    \item If $i$ is odd, all items in cluster $S_{i}$ are sampled against the same items between iterations $t_{\text{init}}$ and $t_{\text{init}}+n|L(S)|$, for every integer $n$.
    \item If $i$ is even, all items in cluster $S_{i}$ are sampled against the same items between iterations $t_{\text{init}}$ and $t_{\text{init}}+nr(S)|L(S)|$, for every integer $n$.
\end{itemize}

Consider the event that a cluster $S_i$ is separated at iteration $t_{split}$. \CMSampling\ determines the list of \textit{matching schemes} and the sampling rates associated with the newly created clusters, and from iteration $t_{split}+1$, the matchings are sampled following those new list of \textit{matching schemes} and the sampling rates.

It may happen that at iteration $t_{split}$, \CMSampling\ is stopped in the middle of the round on $L(S)$ for some cluster $S$. Thus, some couple of items in the cluster have not been uniformly sampled against the same other items. 
To correct this, all samples gathered after iteration $(t_{split}-[(t_{split}-t_{\text{init}})\mod{r(S)|L(S)|}])$ are dropped for the items in $S$. 

\begin{lemma}\label{regretdropped}
The sum of the regret for all dropped samples is at most $32N^6$. 
\end{lemma}
\textbf{Proof}: Note that for any cluster $S$, it always holds that $r(S)|L(S)|<(2N)^4$. Thus, the samples of at most $(2N)^4$ iterations are not yet stored in $X$ and $C$ at any given split. Those iterations all cost at most $N$ in terms of regret, and there are at most $2N$ splits. \hfill \(\Box\)

\subsection{Proof of the upper bound} In this section we prove the following theorem.
\mds*

The proof goes as follows. The \MDC\ algorithm tracks estimators of the expected value of the total reward received for each of the $2N$ parameters. Lemma \ref{BoundProbaFail}, together with a union bound and the fact that the regret per iteration is always smaller than $N$ imply
$$
\E[R(T)] \leq  \E[R(T) \mid \mathcal{A}] +T\Pb(\bar{\mathcal{A}}) \leq \E[R(T) \mid \mathcal{A}]+4N^2.
$$



We denote with $R_{S_k,\ldots,S_{k'}}$ the average regret of the \CMSampling\ procedure on the chain of clusters $S_k,\ldots,S_{k'}$. 
The following lemma states that the average regret of \CMSampling\ on any chain of clusters can be decomposed into the sum of average regrets for isolated clusters and trivial chains of clusters. 

\begin{lemma}[Chain regret]\label{ensemblecluster}
The expected average regret of the \CMSampling\ procedure on a chain of clusters $S_k,\ldots,S_{k'}$ is equal to
\begin{align*}
    R_{S_k,\ldots,S_{k'}}
    &= R_{S_k\cup\{v(S_{k+1})\}}+\sum_{i=k+1}^{k'-1}R_{\{l_{i-1}\},S_i,\{v(S_{i+1})\}}+R_{\{l_{k'-1}\}\cup S_{k'}}
\end{align*}
where $v(S_{i})$ is a virtual item with parameter value $\bar{u}_{S_i}= \frac{1}{|S_i|}\sum_{k \in S_i}u_k$, and $l_i$ is the highest ranked item in $S_i$.
\end{lemma}

To transform the problem on chains of cluster onto a problem on isolated clusters and trivial chains of clusters, we introduce some additional notations. We denote with $S_t(i)$ the cluster to which item $i$ belongs to at iteration $t$, and we let $\mathcal{E}_i(t)$ be the event $\{S_t(2i)=S_t(2i-1)\}$. Let $h_t(j)$ be a virtual item such that
$$
h_t(j) = \mathds{1}_{\mathcal{E}_{j}(t)}j+\mathds{1}_{\overline{\mathcal{E}}_{j}(t)}v(S_t(j)).
$$

The \emph{virtual match event} $\mathcal{V}_t(l,j)$, is defined as follows:
$$
\mathcal{V}_t(l,j)= (\{l\in S(j)\}\cap \{(l,j)\in m_t\}) \cup(\{l\in S(n^*(j))\}\cap \overline{\mathcal{E}}_j(t)\cap \{m_t(l)\in S(j)\}).
$$
This means that an item $l$ is "virtually" matched with item $j$ when either item $j$ belongs to the same cluster as $l$ and $l$ is matched with $j$ at iteration $t$, or $l$ belongs to the same cluster as $j$'s optimal neighbor $n^*(j)$, $j$ belongs to another cluster $S$ and $l$ is matched with an item in $S$ at iteration $t$.

We define $\tilde{R}_{i,j}(T)$ as follows:

\begin{eqnarray*}
\tilde{R}_{i,j}(T) &=& \sum_{t=1}^T \sum_{k \in \{2i-1,2i\}}\mathds{1}_{\mathcal{V}_t(2j-1,k)} \left(\frac{1}{2}(u_{2i-1}u_{2i}+u_{2j-1}u_{2j})-u_{l}u_{2j-1}\right)\\
&&+\sum_{t=1}^T \sum_{k \in \{2i-1,2i\}}\mathds{1}_{\mathcal{V}_t(k,2j)} \left(\frac{1}{2}(u_{2i-1}u_{2i}+u_{2j-1}u_{2j})-u_{l}u_{h_t(2j)}\right).\\
\end{eqnarray*}

Note that under Assumption \ref{hyp:even}, $\tilde{R}_{i,j}(T)$ and $R_{i,j}(T)$ are equivalent, since pairs of items are never split into different clusters.

According to Lemma \ref{ensemblecluster}, the following equation holds:
$$
\sum_{i=1}^N \sum_{j=i+1}^N R_{i,j}(T) = \sum_{i=1}^N \sum_{j=i+1}^N  \tilde{R}_{i,j}(T).
$$

It remains to bound $\tilde{R}_{i,j}(T)$.

\begin{lemma}[Trivial chain regret]\label{bookended}
The average regret of the \CMSampling\ procedure on a trivial chain of clusters is upper bounded as
\begin{align*}
    R_{\{2k-1\},[2k,2l-1],\{2l\}}
    \leq&\frac{2}{2K+1}(u_{2k-1}-u_{2l})(u_{2k}-u_{2l-1})\\
    & + \frac{1}{2K+1}\sum_{j=k+1}^{l-1}\sum_{j'=j+1}^{l-1} r_{j,j'}
    +\frac{1}{2K+1}\sum_{j =k+1}^{p-1}r_{j,l}+r_{k,j}
\end{align*}
where $K =l-k$.
\end{lemma}

Note that this is exactly the regret of uniform matching over items $\{2k-1,\ldots,2l\}$, except that all inter-pair matches between pairs $k$ and $l$ are the most favorable ones, namely $(2k-1,2l-1)$ and $(2k,2l)$. Recall that an inter-pair match between a pair $i$ and a pair $j$ is the match of an item of pair $i$ with an item of pair $j$.

Recall the definition of $\nu(S)$ in equation (\ref{nuS}). Note that $\tilde{R}_{i,j}(T)$ has the following properties:
\begin{itemize}
    \item If $S_t(2i)\neq S_t(2j-1)$:
    $$
    \tilde{R}_{i,j}(t+1)-\tilde{R}_{i,j}(t)=0.
    $$
    \item If pairs $i$ and $j$ belong to the same cluster $S$ between iterations $t$ and $t+\nu(S)$, then 
    $$
    \tilde{R}_{i,j}(t+\nu(S))-\tilde{R}_{i,j}(t)\leq r_{i,j}
    $$
    which follows from Lemmas \ref{bookended} and  \ref{isolated}.
    \item If at least three items of pairs $i$ and $j$ belong to the cluster $S$, the last property holds with $r_{i,j}$ replaced with
    $$
    r_{i,j,t} = (u_{2i}-u_{2j-1})(u_{2i-1}-u_{h_t(2j)})+(u_{2i-1}-u_{2j-1})(u_{2i}-u_{h_t(2j)})
    $$
    \item If only items $2i$ and $2j-1$ belong to the same cluster $S$ between iterations $t$ and $t+\nu(S)$, we have
    $$
    \tilde{R}_{i,j}(t+\nu(S))-\tilde{R}_{i,j}(t)\leq 2(u_{2i}-u_{2j-1})(u_{2i-1}-u_{h_t(2j)})
    $$
    which follows from Lemma~\ref{bookended}. Notice that in that case, $S$ is necessarily an intermediate cluster in a chain.
\end{itemize}

The following equality holds:
\begin{align*}
    r_{i,j}&=(u_{2i}-u_{2j-1})(u_{2i-1}-u_{2j})+(u_{2i-1}-u_{2j-1})(u_{2i}-u_{2j})\\
    &=2(u_{2i}-u_{2j-1})(u_{2i-1}-u_{2j})+\Delta_{2i-1,2i}\Delta_{2j-1,2j}.
\end{align*}

Let us define 
$$
\bar{u}_j(t) = \frac{1}{t}\sum_{s=1}^tu_{h_t(j)}
$$
and
$\mathcal{E}_{i,j}(S)$ is redefined to be the event that cluster $S$ is the last cluster to contain  both items $2i$ and $2j-1$. And $\tilde{\mathcal{E}}_{i,j}(S')$ is the event that cluster $S'$ is the last cluster to contain at least three items of pairs $i$ and $j$. For a cluster $S$, $T_S$ is the minimal number s.t. if for some $i\in S$, $\sum_{j\in[2N]}C(i,j)>T_S$, then $S$ is separated. 
The following relation holds
\begin{align*}
\E[\tilde{R}_{i,j}(T)\mid \mathcal{E}_{i,j}(S)\cap \tilde{\mathcal{E}}_{i,j}(S')]\leq &\frac{T_S}{\nu(S)}2(u_{2i}-u_{2j-1})(u_{2i-1}-\bar{u}_{2j}(T_S))\\
&+\frac{T_{S'}}{\nu({S'})}\Delta_{2i-1,2i}(u_{2j-1}-\bar{u}_{2j}(T_{S'}))\\
&+\underbrace{h(i,j)+\frac{r(S')|L(S')|}{\nu(S')}+\frac{r(S)|L(S)|}{\nu(S)}}_{:=b(i,j)}
\end{align*}

where $h(i,j)$ accounts for the regret due to dropped samples. According to lemma \ref{regretdropped}:
$$
\sum_{(i,j)}h(i,j)\leq 32N^6.
$$

Note that for any cluster $S$, it always holds that $|L(S)|/\nu(S)<N$. Thus:
$$
\frac{r(S')|L(S')|}{\nu(S')}+\frac{r(S)|L(S)|}{\nu(S)}\leq 2N^3
$$

\begin{lemma}\label{maxsample2}
Consider an isolated cluster $S=\{2l-1,\dots,2l'\}$, under event $\mathcal{A}$, the minimal number $T_S$ s.t. if for some $i\in S$, $\sum_{j\in[2N]}C(i,j)>T_S$, then $S$ is separated, is upper bounded as
$$
T_S \leq 64\frac{1}{(\tilde{\mu}_{-\{k,k+1\}}\Delta_{k,k+1})^2}\log(T),\ \hbox{ for all } k\in \{2l,\dots,2l'-2\}
$$
and
$$
T_S \leq 64\max\left\{\frac{1}{(\tilde{\mu}_{-\{2l-1,2l\}}\Delta_{2l-1,2l})^2},\frac{1}{(\tilde{\mu}_{-\{2l'-1,2l'\}}\Delta_{2l'-1,2l'})^2}\right\}\log(T).
$$
where
$$
\tilde{\mu}_{-\{k,k+1\}} = \frac{1}{\nu(S)}\left(\sum_{i=2l-1}^{2l'}u_i-u_{k}-u_{k+1}\right).
$$
If cluster $S=\{l,\ldots,l'\}$ belongs to a chain and is ranked between clusters $S_{-1}$ and $S_{+1}$ (those clusters may be empty if $S$ is the first or last element of the chain), under event $\mathcal{A}$, $T_S$ is upper bounded as
$$
T_S \leq 64\frac{1}{(\tilde{\mu}_{-\{k,k+1\}}\Delta_{k,k+1})^2}\log(T),\ \hbox{ for all } k\in \{l,\ldots,l'-1\}
$$
where
$$
\tilde{\mu}_{-\{k,k+1\}} = \frac{1}{\nu(S)}\left(\bar{u}_{S_{-1}}+\left(\sum_{i=l}^{l'}u_i\right)-u_{k}-u_{k+1}\right).
$$
\end{lemma}

This lemma is a consequence of Lemma \ref{maxsample} and equations (\ref{eqn:meancompare2}),(\ref{eqn:meancompare3}).

The following lemma bounds the regret for interactions between pair $i$ and $j$.

\begin{lemma}\label{pairwiseRbound2} Under event $\mathcal{A}$, the total regret for interaction between pairs $i$ and $j$ such that $j-i\geq 2$ is upper bounded as follows
$$
\tilde{R}_{i,j}(T) - b(i,j)\leq \frac{c_{\mathfrak{u}}}{(j-i-1)}\frac{1}{\Delta_{\mathrm{min}}}\log(T).
$$
If $j-i=1$, the following bound holds
$$
\tilde{R}_{i,j}(T)-b(i,j) \leq \frac{c_{\mathfrak{u}}}{\Delta_{\mathrm{min}}}\log(T).
$$
\end{lemma}
The proof of Lemma~\ref{pairwiseRbound2} is given in Appendix~\ref{pro:pairwiseRbound2}.

It remains to sum the bounds over all possible pairs:
\begin{eqnarray*}
R(T) & \leq & c_{\mathfrak{u}} \left(N+
    \sum_{i=1}^{N}\sum_{j=i+2}^{N}\frac{1}{j-i-1}\right)\frac{\log(T)}{\Delta_{\mathrm{min}}}+32N^6+N^5+4N^2\\
    & \leq & c_{\mathfrak{u}}\left(N+
    \sum_{i=1}^{N}\sum_{k=1}^{N-i-1}\frac{1}{k}\right)\frac{\log(T)}{\Delta_{\mathrm{min}}}+34N^6\\
    & \leq & c_{\mathfrak{u}}\left(2N+
    \sum_{i=1}^{N}\log(N-i-1)\right)\frac{\log(T)}{\Delta_{\mathrm{min}}}+34N^6\\
    & \leq & c_{\mathfrak{u}}\frac{N(\log(N)+2)}{\Delta_{\mathrm{min}}}\log(T)+34N^6.
\end{eqnarray*}

\hfill \(\Box\)
\subsubsection{Proof of Lemma \ref{bookended}}

Let $p_{i,j}$ be the proportion of draws of pair $(i,j)$ by \CMSampling\  on the chain of clusters $\{2k-1\},S,\{2l\}$. Let $S=\{2k\} \cup A \cup \{2l-1\}$ so that $A$ is the set of items in $S$ not in pair $k$ or $l$.
As noted in the analysis of \CMSampling\, $p_{i,j}$  is defined as

$$
p_{i,j} = \left\{
\begin{array}{cl}
\frac{1}{|S|} & \hbox{ if }i \in S, j = 2k-1 \hbox{ or } j = 2l\\
0 & \hbox{ if } i = 2k-1 \hbox{ and } j = 2l\\
\frac{1}{|S|}\frac{|S|-2}{|S|-1} & \hbox{ if } (i,j)\in S^2,i\neq j
\end{array}
\right . .
$$

By definition, we have
\begin{align*}
    R_{\{2l-1\},S,\{2p\}}= \frac{1}{2}\sum_{i,j \in \{u_{2l-1}\} \cup S \cup \{u_{2p}\}}  p_{i,j} u_i(u_{m^*(i)}-u_j).
\end{align*}

The terms on the right hand side can be grouped to obtain the desired expression.

The first group of terms corresponds to a scaled-down uniform matching on set $A$,

\begin{align*}
    R_1 &=\frac{|S|-2}{|S|(|S|-1)}\sum_{j=l+1}^{p-1}\sum_{j'=j+1}^{p-1} r_{j,j'}.
\end{align*}

The second group of terms corresponds to matches $(2l,2p)$ and $(2l-1,2p-1)$, with value
\begin{align*}
    R_2 = \frac{1}{|S|}(u_{2l-1}-u_{2p})(u_{2l}-u_{2p-1}).
\end{align*}

The third group of terms corresponds to matches $(2l-1,j)$, $(2l,m^*(j))$, and $(2p-1,j)$, $(2p,m^*(j))\}$, with respective values
\begin{align*}
    R_3 = \frac{|S|-2}{|S|(|S|-1)}\sum_{j=l+1}^{p-1}r_{l,j} \hbox{ and } R_3' = \frac{|S|-2}{|S|(|S|-1)}\sum_{j=l+1}^{p-1}r_{j,p}.
\end{align*}

 The last group of terms corresponds to matches $(2l-1,j)$, $(2p,m^*(j))$, and $(2p-1,2l)$, with value 
\begin{align*}
   R_4 &\leq \frac{|S|-2}{|S|(|S|-1)} (u_{2l-1}-u_{2p})(u_{2l}-u_{2p-1})
\end{align*}

where the inequality comes from the fact that matches $(2l-1,2p-1)$, $(j,m^*(j))$, and $(2l,2p)$ have a smaller expected reward than matches $(2l-1,j)$, $(2p,m^*(j))$, and $(2p-1,2l)$.

Summing $R_1+R_2+R_3+R_3'+R_4$ gives the result of the lemma.

\subsubsection{Proof of Lemma \ref{ensemblecluster}}\label{Proensemblecluster}

Let $S_j,\ldots,S_{j'}$ be a chain of clusters, and $l_i$ denote the highest-ranked item in cluster $i$. Let $p_{k,k'}$ denote the proportion of draws of pair $(k,k')$ by \CMSampling\  on $S_j,\ldots,S_{j'}$. We use the convention $S_{j-1}=S_{j'+1}=\emptyset$. Let $R_{S_j,...,S_{j'}}$ denote the expected average regret of \CMSampling\ on $S_j,\ldots,S_{j'}$.

We have the following relations

\begin{align*}
    2R_{S_j,\ldots,S_{j'}} =& \sum_{i=j}^{j'}\sum_{k\in S_i} u_{k}u_{m^*(k)} -\sum_{i=j}^{j'}\sum_{k,k'\in S_i}p_{k,k'}u_{k}u_{k'}\\
    &-2\sum_{i=j}^{j'}\sum_{k\in S_i}\sum_{k'\in S_{i-1}}p_{k,k'}\left(u_{k}u_{k'}-u_ku_{l_{i-1}}+u_ku_{l_{i-1}}\right)\\
    &+2\sum_{i=j}^{j'}\sum_{k\in S_i}\sum_{k'\in S_{i+1}}p_{k,k'}\left(u_{k}u_{k'}-u_{k'}u_{l_i}+u_{k'}u_{l_i}\right)\\
    =& \sum_{i=j}^{j'}\left(2u_{l_i}\sum_{k \in S_i,k'\in S_{i+1}}p_{k,k'}u_{k'}+\sum_{k\in S_i\cup\{l_{i-1}\}\setminus\{l_i\}}u_{k}u_{m^*(k)}\right)\\
    &-\sum_{i=j}^{j'}\left(\sum_{k,k'\in S_i}p_{k,k'}u_k u_{k'}+2\sum_{k\in S_i,k'\in S_{i+1}}p_{k,k'}u_ku_{k'}+2\sum_{k'\in S_{i-1},k \in S_i}p_{k,k'}u_ku_{l_{i-1}}\right).
\end{align*}

Let us define
$$
\bar{u}_{S} = \frac{1}{|S|}\sum_{k \in S}u_k
$$
and let us use the following convention $u_{l_{j-1}}=0$ and $\bar{u}_{S_{p+1}}=0$.

Note that the following equations hold:
\begin{eqnarray*}
\sum_{k \in S_i,k'\in S_{i+1}}p_{k,k'}u_{k'} &=& \bar{u}_{S_{i+1}}\\ 
\sum_{k\in S_i,k'\in S_{i+1}}p_{k,k'}u_ku_{k'} &=&  \sum_{k\in S_i}\frac{1}{|S_i|}u_k\bar{u}_{S_{i+1}}\\
\sum_{k'\in S_{i-1},k \in S_i}p_{k,k'}u_ku_{l_{i-1}} &=& \sum_{k\in S_i}\frac{1}{|S_i|}u_ku_{l_{i-1}}.
\end{eqnarray*}

The equation above for $R_{S_j,\ldots,S_{j'}}$ can be simplified to
\begin{align*}
    2R_{S_j,\ldots,S_{j'}}
    = & \sum_{i=j}^{j'}\left(2u_{l_i}\bar{u}_{S_{i+1}}+\sum_{k\in S_i\cup\{l_{i-1}\}\setminus\{l_i\}}u_{k}u_{m^*(k)}\right)\\
    &-\sum_{i=j}^{j'}\left(\sum_{k,k'\in S_i}p_{k,k'}u_ku_k'+2\sum_{k\in S_i}\frac{1}{|S_i|}u_k\bar{u}_{S_{i+1}}+2\sum_{k\in S_i}\frac{1}{|S_i|}u_ku_{l_{i-1}}\right)
\end{align*}

which can be written as
\begin{align*}
    R_{S_j,\ldots,S_{j'}}
    &= R_{S_j\cup\{v_{j+1}\}}+\sum_{i=j+1}^{j'-1}R_{\{l_{i-1}\},S_i,\{v_{i+1}\}}+R_{\{l_{j'-1}\}\cup S_{j'}}
\end{align*}
where $v_k$ is a virtual item with parameter value $\bar{u}_{S_{k}}$. 

\subsubsection{Proof of Lemma~\ref{pairwiseRbound2}}\label{pro:pairwiseRbound2}

For any two clusters $S$ and $S'$, the following holds:
\begin{align*}
\E[\tilde{R}_{i,j}(T)\mid \mathcal{E}_{i,j}(S)\cap \tilde{\mathcal{E}}_{i,j}(S')]-b(i,j)\leq &\frac{T_S}{\nu(S)}2(u_{2i}-u_{2j-1})(u_{2i-1}-\bar{u}_{2j}(T_S))\\
&+\frac{T_{S'}}{\nu({S'})}\Delta_{2i-1,2i}(u_{2j-1}-\bar{u}_{2j}(T_{S'})).
\end{align*}

To simplify the proof, we first assume that $S'=\{2i-1,\dots,2j\}$ and $S=\{2i,\dots,2j-1\}$ or $S=\{2i-1,\dots,2j\}$, and then argue that the obtained result holds for arbitrary $S$ and $S'$. Note that this implies that $S\subset S'$, hence $S$ is either isolated or in the trivial chain $\{2i-1\},S,\{2j\}$.

The section starts with the presentations  and proofs of the two following lemmas. The proof of Lemma \ref{pairwiseRbound2} is provided afterwards.

\begin{lemma}[Intermediate result]\label{Intermediate}
Assume that $S'=\{2i-1,\dots,2j\}$ or $S'=\{2i,\dots,2j-1\}$  and $S=\{2i,\dots,2j-1\}$, with $j-i\geq 2$. Then, we have
$$
\E[\tilde{R}_{i,j}(T)\mid \mathcal{E}_{i,j}(S)\cap \tilde{\mathcal{E}}_{i,j}(S')]-b(i,j)\leq \frac{c_{\mathfrak{u}}}{(j-i-1)\Delta_{\mathrm{min}}}\log(T).
$$
\end{lemma}

\begin{lemma}[Neighboring pairs]\label{neighbor}
Suppose $S'=\{2i-1,\dots,2j\}$ with $j-i=1$. Then, we have
$$
\E[\tilde{R}_{i,j}(T)\mid \mathcal{E}_{i,j}(S)\cap \tilde{\mathcal{E}}_{i,j}(S')]-b(i,j)\leq \frac{c_{\mathfrak{u}}}{\Delta_{\mathrm{min}}}\log(T).
$$
\end{lemma}

\paragraph{Proof of Lemma \ref{Intermediate}}

Let $m = \argmax_{k \in \{2i,\dots,2j-1\}}\Delta_{k,k+1}$. The following inequality holds
\begin{equation}\label{gapupperbound}
(u_{2i}-u_{2j-1})(u_{2i-1}-u_{2j})\leq (2(j-i)-1)^2\Delta_{m,m+1}^2+\Delta_{m,m+1}(\Delta_{2i-1,2i}+\Delta_{2-1,2j}).
\end{equation}

Let $\gamma = \min_{k \in [i,j-1]} \Delta_{2i-1,2i+2}$. We have the following inequalities
$$
\Delta_{\mathrm{min}}\leq \gamma^2 \hbox{ and } \Delta_{\mathrm{min}}\leq \gamma \Delta_{m,m+1}.
$$

For all $k\in[2i-1,2j-1]$,
$$
u_k\geq u_{2j}+\frac{1}{2}\Delta_{2j-1,2j}+\frac{1}{2}\sum_{l=\lceil\frac{k+1}{2}\rceil}^{j-1}\gamma.
$$
Hence, it follows
\begin{align*}
    (2(j-i)+1)\mu_{-m,m+1} \geq&\sum_{k=2i-1}^{2j-1}u_k-u_{m}-u_{m+1} \numberthis \label{muLB}\\
    \geq&u_{2i-1}+\sum_{k=2i+2}^{2j-1}u_k\\
    \geq&\frac{j-i}{2}\Delta_{2j-1,2j}+\Delta_{2i-1,2i}+\frac{1}{2}\left(\sum_{k=2i+1}^{2j-1}\left(j-\left\lceil\frac{k+1}{2}\right\rceil\right)\right)\gamma\\
    \geq&\frac{j-i}{2}\Delta_{2j-1,2j}+\Delta_{2i-1,2i}+\frac{(j-i-1)^2}{2}\gamma.
\end{align*}

Combining with Equation~(\ref{gapupperbound}), we have
\begin{align*}
    A_{i,j}(S) &:=\frac{T_S}{2(j-i)+1}2(u_{2i}-u_{2j-1})(u_{2i-1}-u_{2j}(T_S))\\
    &\leq 64\frac{2(j-i)+1}{\Delta_{m,m+1}^2}\frac{(2(j-i)-1)^2\Delta_{m,m+1}^2+\Delta_{m,m+1}(\Delta_{2i-1,2i}+\Delta_{2j-1,2j})}{(\frac{j-i}{2}\Delta_{2j-1,2j}+\Delta_{2i-1,2i}+\frac{(j-i-1)^2}{2}\gamma)^2}\log(T).
\end{align*}

The following holds

\begin{align*}
    a_1&:=\frac{2(j-i)+1}{\Delta_{m,m+1}^2}\frac{(2(j-i)-1)^2\Delta_{m,m+1}^2}{(\frac{j-i}{2}\Delta_{2j-1,2j}+\Delta_{2i-1,2i}+\frac{(j-i-1)^2}{2}\gamma)^2}\\
    &\leq \frac{4(2(j-i)+1)(2(j-i)-1)^2}{(j-i-1)^4}\frac{1}{\gamma^2}\\
    &\leq \frac{180}{j-i-1}\frac{1}{\Delta_{\mathrm{min}}}
\end{align*}

and
\begin{align*}
    a_2&:= \frac{2(j-i)+1}{\Delta_{m,m+1}^2}\frac{\Delta_{m,m+1}(\Delta_{2i-1,2i}+\Delta_{2j-1,2j})}{(\frac{j-i}{2}\Delta_{2j-1,2j}+\Delta_{2i-1,2i}+\frac{(j-i-1)^2}{2}\gamma)^2}\\
    &\leq \frac{2(j-i)+1}{\Delta_{m,m+1}\gamma}\frac{\Delta_{m,m+1}(\Delta_{2i-1,2i}+\Delta_{2j-1,2j})}{(\Delta_{2j-1,2j}+\Delta_{2i-1,2i})(j-i-1)^2\Delta_{m,m+1}}\\
    &\leq \frac{5}{j-i-1}\frac{1}{\Delta_{\mathrm{min}}}.
\end{align*}

Putting the bounds on $a_1$ and $a_2$ together gives
$$
A_{i,j}(S) \leq  \frac{c_{\mathfrak{u}}}{j-i-1}\frac{1}{\Delta_{\mathrm{min}}}.
$$

We note
$$
B_{i,j}(S') :=  \frac{T_{S'}}{\nu({S'})}\Delta_{2i-1,2i}\Delta_{2j-1,2j}.
$$

We separately consider two different cases as follows.

\begin{itemize}
    \item {\bf Case $\Delta_{2i-1,2i}\leq \Delta_{m,m+1}$ or $\Delta_{2j-1,2j}\leq \Delta_{m,m+1}$.}
    By definition of $S$ and $S'$, $S'$ is a cluster containing $S$. Hence, we have
$$
\frac{T_{S'}}{\nu({S'})}\leq \frac{T_{S}}{\nu({S})}.
$$

Then, we have
$$
\frac{T_{S'}}{\nu({S'})}\Delta_{2i-1,2i}\Delta_{2j-1,2j} \leq \frac{T_{S}}{\nu({S})}\Delta_{m,m+1}(\Delta_{2i-1,2i}+\Delta_{2j-1,2j})=64a_2
$$
which implies, according to the bound on $a_2$,
$$
B_{i,j}(S')  \leq  \frac{320}{j-i-1}\frac{1}{\Delta_{\mathrm{min}}}\log(T).
$$
\item {\bf Case $\Delta_{2i-1,2i}> \Delta_{m,m+1}$ and $\Delta_{2j-1,2j}> \Delta_{m,m+1}$}. According to Lemma \ref{maxsample2}, we have
$$
B_{i,j}(S') \leq \max\left\{\frac{64}{\Delta_{2i-1,2i}^2\mu_{-2i-1,2i}^2},\frac{64}{\Delta_{2j-1,2i}^2\mu_{-2j-1,2j}^2} \right\}\frac{\Delta_{2i-1,2i}\Delta_{2j-1,2j}}{2(j-i)+1}\log(T).
$$
Repeating the computations in Equation~(\ref{muLB}) gives
$$
(2(j-i)+1)\mu_{-2j-1,2j}\geq \Delta_{2i-1,2i}+\frac{(j-i-1)^2}{2}\gamma.
$$
Hence, we have
\begin{align*}
\frac{\Delta_{2i-1,2i}\Delta_{2j-1,2j}}{(2(j-i)+1)\Delta_{2j-1,2j}^2\mu_{-2j-1,2j}^2} &\leq \frac{2(j-i)+1}{(j-i-1)^2\gamma\Delta_{2j-1,2j}}\\
&\leq \frac{5}{(j-i-1)\Delta_{\mathrm{min}}}.
\end{align*}
Similarly, we obtain
$$
(2(j-i)+1)\mu_{-2i-1,2i}\geq \Delta_{2j-1,2j}+\frac{(j-i-1)^2}{2}\gamma.
$$
This implies that the following holds
$$
B_{ij}(S') \leq \frac{320}{(j-i-1)\Delta_{\mathrm{min}}}\log(T).
$$
\end{itemize}

Combining the bounds on $A_{i,j}(S)$ and $B_{i,j}(S')$ gives the following bound
\begin{equation}\label{endresult}
\E[\tilde{R}_{i,j}(T)\mid \mathcal{E}_{i,j}(S)\cap\tilde{\mathcal{E}}_{i,j}(S')] \leq \frac{c_{\mathfrak{u}}}{j-i-1}\frac{1}{\Delta_{\mathrm{min}}}\log(T)
\end{equation}
when $S'=\{2i-1,\dots,2j\}$ and $S=\{2i,\dots,2j-1\}$, $j-i\geq2$. \hfill \(\Box\)

\paragraph{Proof of Lemma \ref{neighbor}} The following inequality holds
$$
\Delta_{\mathrm{min}}\leq u_{2i-1}\Delta_{2i,2j-1}.
$$
Hence, we have
\begin{align*}
\frac{T_S}{\nu(S)}2(u_{2i}-u_{2j-1})(u_{2i-1}-u_{2j})&\leq 64 \cdot 6\frac{(u_{2i}-u_{2j-1})}{\Delta_{2i,2j-1}u_{2i}^2}\log(T)\\
&\leq \frac{384}{\Delta_{2i,2j-1}u_{2i}}\log(T)\\
&\leq\frac{384}{\Delta_{\mathrm{min}}}\log(T).
\end{align*}

If $\Delta_{2i-1,2i}\leq \Delta_{2i,2j-1}$ or $\Delta_{2j-1,2j}\leq \Delta_{2i,2j-1}$, then
$$
\frac{T_{S'}}{\nu(S')}\Delta_{2i-1,2i}\Delta_{2j-1,2j}\leq \frac{T_{S}}{\nu(S)}(u_{2i}-u_{2j-1})(u_{2i-1}-u_{2j}).
$$
If $\Delta_{2i-1,2i}> \Delta_{2i,2j-1}$ and $\Delta_{2j-1,2j}> \Delta_{2i,2j-1}$, then
\begin{align*}
 \frac{T_{S'}}{\nu(S')}\Delta_{2i-1,2i}\Delta_{2j-1,2j}&\leq64 \max\left\{\frac{3\Delta_{2i-1,2i}\Delta_{2j-1,2j}}{\Delta_{2i-1,2i}^2(u_{2j-1}+u_{2j})^2},\frac{3\Delta_{2i-1,2i}\Delta_{2j-1,2j}}{\Delta_{2j-1,2j}^2(u_{2i-1}+u_{2i})^2}\right\}\log(T)\\  
 &\leq 192 \max\left\{\frac{1}{\Delta_{2i-1,2i}u_{2j-1}},\frac{1}{\Delta_{2j-1,2j}u_{2i-1}}\right\}\log(T)\\
 &\leq \frac{192}{\Delta_{\mathrm{min}}}\log(T).
\end{align*}
\hfill \(\Box\)

\paragraph{Proof of Lemma \ref{pairwiseRbound2}}

Up to now, $S'$ was an isolated cluster. Consider now it is the first element of a chain $S'=\{2i-1,\dots,2j-1\},S_2,\ldots$. All previous computations hold with $\bar{u}_{2j}(T_S')$ instead of $u_{2j}$, hence the result remains true.

If $S'$ is the last element of a chain, 
$u_{2i-1}$ needs to be replaced by $\bar{u}_{2i-1}(T_S')$ in all expressions for $\mu_{-k,k+1}$. Since $u_{S_{-1}}\geq u_{2i-1}$, it holds that $\bar{u}_{2i-1}(T_S')\geq u_{2i-1}$, so all previous results remain true.

Therefore, for any cluster $S',$ such that $S'=\{2i,\dots,2j\}$,$S'=\{2i-1,\dots,2j-1\}$ or $S'=\{2i-1,\dots,2j\}$, and $S=S'$ or $S=\{2i,\dots,2j-1\}$:
\begin{align*}
    \E[\tilde{R}_{i,j}(T)|\mathcal{E}_{i,j}(S),\tilde{\mathcal{E}}_{i,j}(S')]-b(i,j)
    &\leq c_{\mathfrak{u}} \min(1,\frac{1}{j-i-1})\frac{1}{\Delta_{\mathrm{min}}}\log(T)
\end{align*}

For any $k\leq i< j < l$ or $k< i< j \leq l$, 
$$
r_{i,j}(t) \leq 2(u_{2k}-u_{2l-1})(u_{2k-1}-\bar{u}_{2l}(t)).
$$
Thus, for any clusters such that $S=[a,b]$ with $a\in (2l-1,2l)$ and $b\in(2k-1,2k)$, and any subset $S'$, we have
\begin{align*}
    \E[\tilde{R}_{i,j}(T)|\mathcal{E}_{i,j}(S),\tilde{\mathcal{E}}_{i,j}(S')]-h(i,j)&\leq\frac{T_S}{\nu(S)} 2(u_{2k}-u_{2l-1})(u_{2k-1}-\bar{u}_{2l}(t))\\
    &\leq c_{\mathfrak{u}} \min(1,\frac{1}{l-k-1})\frac{1}{\Delta_{\mathrm{min}}}\log(T)\\
    &\leq c_{\mathfrak{u}}\min(1,\frac{1}{j-i-1})\frac{1}{\Delta_{\mathrm{min}}}\log(T)
\end{align*}
where the second inequality comes from the computations of the proof of Lemma \ref{Intermediate}. This proves that the result still holds if the clusters $S$ and $S'$ are larger (in the sense of inclusion) than those studied in Lemma \ref{Intermediate}, and completes the proof. 
\hfill \(\Box\)

\section{Comparison of \MDC\ with an exploration policy}\label{comparisonAppendix}

\subsection{Proof of Lemma \ref{strategyComparison}}\label{proofStratcomp}
\comparison*

Consider a matching bandit problem on the ordered sequence of cluster
$$
\mathfrak{S}= \{1,2\}, \{3,...,2N\}
$$
where the two best items are identified.

We consider an \textit{information first strategy} that matches the items in $\mathfrak{S}$ following a round-robin tournament.  The \MDC\ algorithm matches $1$ with $2$ and items in set $\{3,...,2N\}$ following a round-robin tournament. We denote with $R_I$ and $R_D$ the regrets incurred by these two strategies, respectively, before they can rank any two items $m$ and $m+1$, $m\in \{3,\dots,2N\}$. According to Lemmas \ref{maxsample} and \ref{isolated}, the following bounds hold
\begin{equation}
R_D \leq U_D:=\frac{2N-3}{\Delta_{m,m+1}^2}\frac{\sum_{i=2}^{N}\sum_{j=i+1}^{N} r_{i,j}}{(\sum_{i=3}^{2N}u_i-u_m-u_{m+1})^2}
\label{equ:RD}
\end{equation}
and
\begin{equation}
R_I \leq U_I :=\frac{2N-2}{\Delta_{m,m+1}^2}\frac{\sum_{j=2}^{N}r_{1,j}+\sum_{i=2}^{N}\sum_{j=i+1}^{N} r_{i,j}}{(\sum_{i=1}^{2N}u_i-u_m-u_{m+1})^2}.
\label{equ:RI}
\end{equation}


Let us define $\rho_1, \ldots, \rho_{2N}$ such that
$$
u_i = \prod_{k=1}^i \rho_k, \hbox{ for } i\in [2N].
$$

\begin{align*}
    r_{i,j} &= 2(u_{2i-1}-u_{2j})(u_{2i}-u_{2j-1})+(u_{2i-1}-u_{2i})(u_{2j-1}-u_{2j}) \\
    &= 2\left(\prod_{k=1}^{2i-1}\rho_k \right)^2 \rho_{2i} \left(1-\prod_{k=2i+1}^{2j}\rho_k\right)\left(1-\prod_{k=2i}^{2j} \rho_k\right)+\prod_{k=1}^{2i-1}\rho_k\prod_{k=1}^{2j-1}\rho_k(1-\rho_{2i})(1-\rho_{2j})\\
    &= \left(\prod_{k=1}^{2i-1}\rho_k \right)^2 \rho_{2i}\underbrace{\left(2 \left(1-\prod_{k=2i+1}^{2j-1}\rho_k\right)\left(1-\prod_{k=2i}^{2j-1}\rho_k\right)+\prod_{k=2i+1}^{2j-1}\rho_k(1-\rho_{2i})(1-\rho_{2j})\right)}_{a_{i,j}}\\
    &= \left(\prod_{k=1}^{2i-1}\rho_k \right)^2\rho_{2i} a_{i,j}
\end{align*}

The expressions for $U_D$ and $U_I$ simplify to

\begin{align*}
U_D &= \frac{2N-3}{\Delta_{m,m+1}^2} \frac{\sum_{i=2}^{N}\sum_{j=i+1}^{N} \left(\prod_{k=1}^{2i-1}\rho_k \right)^2\rho_{2i} a_{i,j}}{(\sum_{i=3,i\neq m,m+1}^{2N}\prod_{k=1}^{i}\rho_k)^2}\\
&= \frac{2N-3}{\Delta_{m,m+1}^2}\rho_4 \frac{\sum_{j=3}^{N}a_{2,j} +\sum_{i=3}^{N}\sum_{j=i+1}^{N} \left(\prod_{k=5}^{2i-1}\rho_k \right)^2\rho_4\rho_{2i} a_{i,j}}{(1+\sum_{i=4,i\neq m,m+1}^{2N}\prod_{k=4}^{i}\rho_k)^2}\\
&=  \frac{2N-3}{\Delta_{m,m+1}^2}\frac{A_2}{(1+\sum_{i=4,i\neq m,m+1}^{2N}\prod_{k=4}^{i}\rho_k)^2}
\end{align*}

where

$$
A_2 =\rho_4\left( \sum_{j=3}^{N}a_{2,j} +\sum_{i=3}^{N}\sum_{j=i+1}^{N} \left(\prod_{k=5}^{2i-1}\rho_k \right)^2\rho_4\rho_{2i} a_{i,j}\right)
$$

and

\begin{align*}
U_I &=  \frac{2N-1}{\Delta_{m,m+1}^2}\frac{\sum_{i=1}^{N}\sum_{j=i+1}^{N} \left(\prod_{k=1}^{2i-1}\rho_k \right)^2\rho_{2i} a_{i,j}}{(\sum_{i=1,i\neq m,m+1}^{2N}\prod_{k=1}^{i}\rho_k)^2}\\
&= \frac{2N-1}{\Delta_{m,m+1}^2}\rho_2 \frac{\sum_{j=2}^{N}a_{1,2} +\sum_{i=2}^{N}\sum_{j=i+1}^{N} \left(\prod_{k=3}^{2i-1}\rho_k \right)^2\rho_2\rho_{2i} a_{i,j}}{(1+\sum_{i=2,,i\neq m,m+1}^{2N}\prod_{k=2}^{i}\rho_k)^2}\\
&= \frac{2N-1}{\Delta_{m,m+1}^2}\rho_2 \frac{\sum_{j=2}^{N}a_{1,j} +\rho_3^2\rho_2 A_2}{(1+\sum_{i=2,i\neq m,m+1}^{2N}\prod_{k=2}^{i}\rho_k)^2}.
\end{align*}

From the last two expressions, we have
$$
\lim_{\rho_2\to 0} \frac{U_D}{U_I} = +\infty.
$$

The following lower bound holds
\begin{align*}
a_{1,j} &=  \left(2 \left(1-\prod_{k=3}^{2j-1}\rho_k\right)\left(1-\prod_{k=2}^{2j-1}\rho_k\right)+\prod_{k=3}^{2j-1}\rho_k(1-\rho_{2})(1-\rho_{2j})\right) \geq 2(1-\rho_3)^2.
\end{align*}

Thus, $U_I$ is lower bounded as
\begin{align*}
    U_I \geq \rho_2 \frac{2N-1}{\Delta_{m,m+1}^2}\frac{2N(1-\rho_3)^2+\rho_3^2\rho_2 A_2}{(1+\sum_{i=2,i\neq m,m+1}^{2N}\prod_{k=2}^{i}\rho_k)^2}.
\end{align*}

Note that $A_2$ is at most of order $N^2$, thus if $\rho_2>1/2$, the ratio $U_D/U_I$ is upper bounded by $c_{\mathfrak{u}} N$. If, in addition, $\rho_3>1/2$, then the ratio $U_D/U_I$ is upper bounded by a constant $c_{\mathfrak{u}}$.

\section{\MatchingId\ algorithm}\label{MatchingIdAppendix}

\subsection{Algorithm description and pseudo-code}

The algorithm is defined by the pseudo-code in Algorithm~\ref{MIdAppendix}.

\begin{algorithm}[htb!]

\caption{\MatchingId}
\begin{algorithmic}
\label{MIdAppendix}
\STATE {\bfseries Input:} set of items $[2N]$, required precision $\delta$
\STATE $U=[2N],X=C=[0]^{[2N]\times[2N]}$
\WHILE{$U\neq \emptyset$}
    \STATE $\cB=\{\text{candidate }|\cS| \text{ best items}\}$
    \STATE $\cA=\cB\cup \cS$
    \IF{$|\cA|$ odd}
        \STATE $k= \argmax [2N] \setminus \cA$
        \STATE $A= A \cup k$
    \ENDIF
    \STATE $L(A) =$ \{round-robin tournament on $A$\}
    \STATE $d=$\{arbitrary matching on $[2N]\ \setminus A$\}
    \FOR{$m \in L(A)$}
        \STATE Sample $m \cup d$
        \FOR{$i \in U$}
        \IF{$(i,j) \in m \text{ and } j \in T$}
            \STATE  $X(i,j)+=x_{i,j,t}$
            \STATE $C(i,j)+=1$
        \ENDIF
        \ENDFOR
    \ENDFOR
     \STATE $Q_{+}^U,Q_{-}^U \leftarrow$ {\tt confidence\_bound}($X,C,U$)
    \STATE $U\leftarrow$ {\tt unranked}($Q_{+}^U,Q_{-}^U,U$)
\ENDWHILE
\end{algorithmic}
\end{algorithm}

The function {\tt confidence\_bound} in Algorithm~\ref{MIdAppendix} computes confidence bounds for the expected reward per item when matched with an item within $T$. If $\sum_{j=1}^{2N}C(i,j)=k_l$ with $k_l=\lceil4^{l+1}\log(\beta_l) \rceil$, $\beta_l=\pi \sqrt{(2N)/(3\delta)}\cdot l$, for some $l$, then
$$
Q_{-}(i)= \frac{\sum_{j=1}^{2N}X(i,j)}{k_l}-\sqrt{\frac{\log(\beta_l)}{k_l}}
$$
and
$$ Q_{+}(i)= \frac{\sum_{j=1}^{2N}X(i,j)}{k_l}+\sqrt{\frac{\log(\beta_l)}{k_l}}.
$$

The function {\tt unranked} in Algorithm~\ref{MIdAppendix} removes an item $i$ from $\cS$ if its associated confidence interval intersects with no other intervals, or if it intersects only with that of another item that would be its optimal match independently of their relative order. 

\subsection{Proof of the upper bound}

We prove the following theorem.

\begin{theorem}[upper bound]
For  any $\delta > 0$, the sample complexity of the \MatchingId\ algorithm satisfies
$$
\tau_{\delta}\leq c_\mathfrak{u}\frac{1}{\gamma_{\mathrm{min}}^2}\log(1/\delta) +c_\mathfrak{p}
$$
with probability at least $1-\delta$. Moreover, if $s\geq2$, then by denoting, 
$$
\alpha:=\min\Big\{\left(\frac{1}{4}\frac{(u_1+u_2)\Delta_{2s,2s+1}}{\mu_{[2N]\setminus\{2h,2h+1\}}\Delta_{2h,2h+1}}\right)^2 1\Big\},
$$
the following holds with probability at least $1-\delta$,
$$
\tau_{\delta} \leq c_\mathfrak{u}\frac{1}{(1-\alpha)^2(u_1^2+u_2^2)\Delta_{2s,2s+1}^2}\log(1/\delta) + c_\mathfrak{p} .
$$
else, if $s=1$, then by denoting, 
$$
\alpha:=\min\Big\{\left(\frac{u_1/4\Delta_{2s,2s+1}}{\mu_{[2N]\setminus\{2h,2h+1\}}\Delta_{2h,2h+1}}\right)^2, 1\Big\}
$$
the following holds with probability at least $1-\delta$,
$$
\tau_{\delta} \leq c_\mathfrak{u}\frac{1}{(1-\alpha)^2(u_1^2)\Delta_{2s,2s+1}^2}\log(1/\delta) + c_\mathfrak{p} .
$$
\end{theorem}

The set constructed by the algorithm, $\cB$ and $\cS$ have the following properties.

\begin{proposition}\label{meanT}
The mean of the elements in $|\cB|$ increases at the set of un-ranked items $\cS$ gets smaller. 
\end{proposition}

The proposition is implied by the definition of $\cB$.

\begin{proposition}\label{UsefulSamples}
It always holds that $|\cB|\leq 2|\cS|$. 
\end{proposition}

\textit{Proof}: If item $i$  with $i>|\cS|$ is ranked, then for any other item $j$ that is not $i$'s neighbor, it is known whether $i<j$ or $j<i$. If $j$ is $i$'s neighbor, either it is  known whether $i<j$ or $j<i$, or for all other item $k$, it is known whether $k< \min\{i,j\}$ or $k>\max\{i,j\}$. Thus $\cB \subset[i]$. \hfill \(\Box\)

We introduce the following notation:
$$
\mu_t(\cB\setminus\{k,l\})=\frac{1}{t}\sum_{s=1}^t\frac{1}{|\cB|}\sum_{i\in \cB\setminus\{k,l\}}u_i.
$$

Two relatively un-ranked $i,j$ items are items in $U$ s.t. $Q_+(i)>Q_-(j)$ and  $Q_+(j)>Q_-(i)$. Any two relatively un-ranked items $i,j$ either both belong to $\cB$, in which case,
$$
\frac{\sum_{l\in \cB\setminus{i}}  u_iu_l}{|\cB\setminus{j}|} -\frac{\sum_{l\in \cB\setminus{j}}  u_ju_l}{|\cB\setminus{j}|}= (u_i-u_j)\frac{\sum_{l\in T\setminus\{i,j\}}u_l}{|\cB|-1},
$$
or, none of them belongs to $\cB$, in which case
$$
\frac{\sum_{l\in \cB\setminus{i}}  u_iu_l}{|\cB\setminus{j}|} -\frac{\sum_{l\in \cB\setminus{j}}  u_ju_l}{|\cB\setminus{j}|}= (u_i-u_j)\frac{\sum_{l\in \cB}u_l}{|\cB|}.
$$

Thus, for any two relatively un-ranked items $i,j$, at every iteration
$$
\E[\frac{\sum_{l\in [2N]}X(i,l)}{\sum_{l\in [2N]}C(i,l)}-\frac{\sum_{l\in [2N]}X(j,l)}{\sum_{l\in [2N]}C(j,l)}] \geq \Delta_{i,j}\mu_t(\cB\setminus\{i,j\}).
$$

According to Proposition~\ref{meanT}, we have
$$
\mu_t(T\setminus\{i,j\}) \geq \frac{1}{2N}\sum_{l\in[2N]\setminus \{i,j\}}u_l.
$$

Thus, items $i$ and $j$ are guaranteed to be relatively ranked by the first iteration where $\sum_{j=1}^{2N}C(i,j)=k_l$  with
$$
\frac{1}{2^{l+1}}<\sqrt{\frac{\log(\beta_l)}{k_l}}<\frac{1}{4}\Delta_{i,j}(\frac{1}{2N}\sum_{l\in[2N]\setminus \{i,j\}}u_l):=\frac{1}{4}\tilde{\Delta}_{i,j}.
$$
This implies
\begin{align*}
    \frac{1}{4}\tilde{\Delta}_{i,j}&\leq \frac{1}{2^l}.
\end{align*}
Thus, we have

\begin{equation}\label{eqn:kl}
   k_l \leq \frac{32}{\tilde{\Delta}_{i,j}^2}\log(\frac{1}{\delta})+\frac{32}{\tilde{\Delta}_{i,j}^2}\log\left(2-\log_2(\tilde{\Delta}_{i,j})\right)^2+\frac{32}{\tilde{\Delta}_{i,j}^2}\log(\frac{2N\pi^2}{3}):=\frac{32}{\tilde{\Delta}_{i,j}^2}\left[\log(\frac{1}{\delta})+f(\tilde{\Delta}_{i,j})\right] 
\end{equation}

Proposition \ref{UsefulSamples} implies that at every iteration, for any $i\in U$ :
\begin{equation}\label{eqn:util}
\frac{\sum_{j \in [2N]}C(i,j)}{t}\geq \frac{4}{9}.
\end{equation}

Recall
$$
\mu_{[2N]\setminus \{2k,2k+1\}}= \frac{1}{2N}\sum_{i\in[2N]\setminus\{2k,2k+1\}}u_i
$$
and 
$$
\gamma_{\mathrm{min}} = \min_{k\in [N-1]} \{\mu_{[2N]\setminus\{2k,2k+1\}}\Delta_{2k,2k+1}\}.
$$
Equations \ref{eqn:kl} and \ref{eqn:util} imply that sample complexity of the \MatchingId\ algorithm is upper bounded as:
$$
\tau_{\delta}\leq \frac{72}{\gamma_{\mathrm{min}}^2}
\left[\log(\frac{1}{\delta})+f\left(\gamma_{\mathrm{min}}\right)\right]
$$
with probability at least $1-\delta$.

Let $s =\arg\min_{k \in [1,N-1] }\Delta_{2k,2k+1}$, and let $h=\arg\min_{k \in [1,N-1]\setminus s }\Delta_{2k,2k+1}\mu_{[2N]\setminus \{2k,2k+1\}}$. Let $\tau_2$ be the stopping time at which all items except those in pairs $s$ and $s+1$ are ranked.  Under the "good event", the number of counted samples for unranked items in pairs $s$ and $s+1$ before $\tau_2$, $\tilde{\tau}_2$ is upper bounded as

$$
\tilde{\tau}_2\leq  (\frac{32}{\Delta_{2h,2h+1}\mu_{[2N]\setminus \{2h,2h+1\}}})^2[\log(\frac{1}{\delta})+f(\Delta_{2h,2h+1}\mu_{[2N]\setminus \{2h,2h+1\}})].
$$

In the case where $s\geq2$, for any $t> \tau_2$, $\cB\subset [4]$, thus the following holds:
$$
\mu_t(T\setminus\{2s,2s+1\})\geq \frac{u_1+u_2+u_3+u_4}{4}+\frac{\tilde{\tau}_2}{\tilde{t}}(\frac{1}{2N}\sum_{l\in[2N]\setminus\{2s,2s+1\}}u_l-\frac{u_1+u_2+u_3+u_4}{4})
$$

At each iteration of the algorithm, the following bound holds on the number of counted samples for un-ranked items in pairs $s,s+1$ $\tilde{t}$
$$
\tilde{t}\leq \frac{32}{(\Delta_{2s,2s+1}\mu_t(T\setminus\{2s,2s+1\})^2}[\log(\frac{1}{\delta})+f(\Delta_{2s,2s+1}\mu_t(\cB\setminus\{2s,2s+1\})]
$$
Now, either
$$
\tilde{\tau}_{\delta}\leq  \frac{32}{(\Delta_{2s,2s+1}\frac{u_1+u_2+u_3+u_4}{4})^2}[\log(\frac{1}{\delta})+f(\Delta_{2s,2s+1}\frac{u_1+u_2+u_3+u_4}{4})^2)]
$$
which implies a bound on $\tau_{\delta}$ by $\tau_{\delta}\leq 9/4\tilde{\tau}_{\delta}$,
or 
$$
\sqrt{\frac{\tilde{\tau}_2}{\tilde{\tau}_{\delta}}}\leq \frac{\Delta_{2s,2s+1}\frac{u_1+u_2+u_3+u_4}{4}}{\Delta_{2h,2h+1}\mu_{[2N]\setminus \{2h,2h+1\}}}
\sqrt{\frac
{\log(\frac{1}{\delta})+f\left(\Delta_{2s,2s+1}\frac{u_1+u_2+u_3+u_4}{4}\right)}
{\log(\frac{1}{\delta})+f\left(\Delta_{2h,2h+1}\mu_{[2N]\setminus \{2h,2h+1\}}\right)}}.
$$

Since $f$ is a decreasing function, if for some $0<\alpha<1$,
$$
\left(\frac{\Delta_{2s,2s+1}\frac{u_1+u_2+u_3+u_4}{4}}{\Delta_{2h,2h+1}\mu_{[2N]\setminus \{2h,2h+1\}}}\right)^2\leq \alpha
$$
then
$$
 \tau_{\delta} \leq \frac{1}{(1-\alpha)^2}\frac{72}{(\Delta_{2s,2s+1}\frac{u_1+u_2+u_3+u_4}{4})^2}\left[\log(\frac{1}{\delta})+f\left((1-\alpha)\Delta_{2s,2s+1}\frac{u_1+u_2+u_3+u_4}{4}\right)\right]
$$
which complicates the proof in the case where $s \geq 2$.

In the case when $s=1$, we have
$$
\Delta_{1,3}\mu_t(\cB\setminus\{1,3\})\geq \Delta_{2,3}\mu_t(\cB\setminus\{2,3\}).
$$
Thus, at every iteration of the algorithm, it holds
$$
\tilde{t}\leq \frac{32}{\left(\Delta_{2,3}\mu_t\left(\cB\setminus\{2,3\}\right)\right)^2}[\log(\frac{1}{\delta})+f(\Delta_{2,3}\mu_t(\cB\setminus\{2,3\})].
$$
Therefore, repeating the steps of the previous proof, if for some $0<\alpha<1$,
$$
\left(\frac{\Delta_{2,3}\frac{u_1}{4}}{\Delta_{2h,2h+1}\frac{1}{2N}\sum_{l\in[2N]\setminus\{2h,2h+1\}}u_l}\right)^2\leq \alpha
$$
then
$$
 \tau_{\delta} \leq \frac{1}{(1-\alpha)^2}\frac{72}{(\Delta_{2,3}\frac{u_1}{4})^2}\left[\log(\frac{1}{\delta})+f\left((1-\alpha)\Delta_{2,3}\frac{u_1}{4}\right)\right].
$$

\subsection{Proof of the lower bound}

We prove the following theorem.

\begin{theorem}
Assume that stochastic rewards of item pairs have Gaussian distribution with unit variance. Then, for any $\delta$-PAC algorithm, we have
\begin{equation*}
\E[\tau_{\delta}]\geq c_\mathfrak{u}\hspace{-0.5cm} \sum_{i\in [2N]: \Delta_i > 0}\frac{1}{\sum_{j=1}^{2N} u_j^2}\frac{1}{\Delta_i^2}\log(1/\delta).
\end{equation*}
Moreover, if $s\geq2$, then
\begin{equation*}
\E[\tau_{\delta}]\geq c_\mathfrak{u} \frac{1}{u_1^2+u_2^2}\frac{1}{\Delta_{2s,2s+1}^2}\log(1/\delta)
\end{equation*}
else, if $s=1$, then
\begin{equation*}
\E[\tau_{\delta}]\geq c_\mathfrak{u} \frac{1}{u_1^2}\frac{1}{\Delta_{2,3}^2}\log(1/\delta).
\end{equation*}
\end{theorem}

Let us first rewrite the linear program that gives a lower bound on the sampling complexity. Let us define
$$\Lambda(U) = \{\lambda \in [0,1]^{2N} \mid m^*(\lambda)\neq m^*(U)\}/$$ to be the class of alternative models that give a different optimal arms. The expected sampling complexity is lower bounded by the solution of the following linear program:
\begin{align*}
    \hbox{ minimize }& \sum_{m \in \mathcal{M}}\eta_m\\
    \hbox{ subject to} & \sum_{m \in \mathcal{M}}\eta_m d(u,\lambda)\geq d(\delta,1-\delta),  \hbox{ for all } \lambda \in \Lambda(U).
\end{align*}

Which is equivalent to
\begin{align*}
    \hbox{ minimize }& \frac{1}{4N}\sum_{1=1}^{2N}\sum_{j=1}^{2N}\sum_{m \in \mathcal{M}}\eta_m\mathds{1}_{\{(i,j)\in m\}}\\
    \hbox{ subject to} & 
    \sum_{1=1}^{2N}\sum_{j=1}^{2N}\sum_{m \in \mathcal{M}}\eta_m \mathds{1}_{\{(i,j)\in m\}}d(u_iu_j,\lambda_i\lambda_j)\geq d(\delta,1-\delta),  \hbox{ for all } \lambda \in \Lambda(U).
\end{align*}

Note that for any matching sampling vector, the following always holds for some constant $c$ and every $(i,j) \in [2N]^2$:

\begin{equation*}
 \left\{
\begin{array}{l}
\sum_{j=1}^{2N}\sum_{m \in \mathcal{M}}\eta_m\mathds{1}_{\{(i,j)\in m\}}=c\\
\sum_{m \in \mathcal{M}}\eta_m\mathds{1}_{\{(i,j)\in m\}}=\sum_{m \in \mathcal{M}}\eta_m\mathds{1}_{\{(j,i)\in m\}}\\
\sum_{m \in \mathcal{M}}\eta_m\mathds{1}_{\{(i,i)\in m\}}=0.
\end{array}
\right .
\end{equation*}

Thus the solution of the previous linear program is lower bounded by that of the following one, $\mathbf{LP}_{m}$,

\begin{align*}
    \hbox{ minimize }& \frac{c}{2}\\
    \hbox{ subject to }& \frac{1}{2}\sum_{(i,j)\in [2N]^2}x_{i,j} d(u_iu_j,\lambda_i\lambda_j)\geq d(\delta,1-\delta),  \ \forall \lambda \in \Lambda(U)  \\
    & \sum_{j=1}^{2N}x_{i,j} = c, \hbox{ for all } i \in [2N]\\
    & x_{i,j}=x_{j,i}, \ x_{i,i}=0, \hbox{ for all } (i,j) \in [2N]^2. 
\end{align*}

In the case where the stochastic rewards are Gaussian random variables with unit variance, $d(u_iu_j,\lambda_i\lambda_j) = (u_iu_j-\lambda_i\lambda_j)^2$. 

We define $s =\arg\min_{k \in [2,N-1]}\Delta_{2k,2k+1}$. By considering alternative models $ \lambda_{2s}=u_{2s}+\Delta_{2s,2s+1}$ and $ \lambda_{2s+1}=u_{2s+1}-\Delta_{2s,2s+1}$, we obtain our first lower bound
\begin{equation}\label{LBsmallgap1}
\E[\tau_{\delta}]\geq \frac{1}{2}\frac{1}{u_1^2+u_2^2}\frac{1}{\Delta_{2s,2s+1}^2}\left(\log\left(\frac{1}{\delta}\right)-1\right).
\end{equation}

Let us change variables in the linear program $\mathbf{LP}_{m}$ for $z_{i,j}=u_j x_{i,j}$. By considering the class of alternative models $\lambda_i=u_i\pm \Delta_i$, and dropping the constraint $x_{i,j}=x_{j,i}$ \hbox{ for $i,j $ ...} we get that the solution of the following linear program is a lower bound on the sampling complexity 

\begin{align*}
    \hbox{ minimize } & \frac{c}{2}\\
    \hbox{ subject to } & \sum_{i=1}^{2N}z_{i,j} \geq \frac{1}{\Delta_j^2}d(\delta,1-\delta),  \ \hbox{ for all } j \in [2N]  \\
    & \sum_{j=1}^{2N}z_{i,j} = cu_i^2, \hbox{ for all } i \in [2N]\\
    & \ z_{i,i}=0, \hbox{ for all } (i,j) \in [2N]^2. 
\end{align*}

Note that $\sum_{i,j} z_{i,j}=c\sum_{j}u_j^2\geq \sum_{i: \Delta_i>0}(1/\Delta_i^2)d(\delta,1-\delta)$. This gives the second bound on the sampling complexity

\begin{equation}
\E[\tau_{\delta}]\geq \left(\sum_{i:\in \Delta_i > 0}\frac{1}{\Delta_i^2}\right)\frac{1}{\sum_j u_j^2}\left(\log\left(\frac{1}{\delta}\right)-1\right).
\end{equation}
\hfill \(\Box\)

Finally, consider alternative model $ \lambda_{2}=u_{2}+\Delta_{2,3}$ and $ \lambda_{3}=u_{3}-\Delta_{2,3}$.
 In this case, we first obtain the same lower bound as in Equation~(\ref{LBsmallgap1}) by removing the constraint $x_{2,2}=0$, of the linear program $\mathbf{LP}_{m}$, then note that the obtained lower bound implies the following one
\begin{equation}\label{LBsmallgap2}
\E[\tau_{\delta}]\geq \frac{1}{4}\frac{1}{u_1^2}\frac{1}{\Delta_{2,3}^2}\left(\log\left(\frac{1}{\delta}\right)-1\right).
\end{equation}
\hfill \(\Box\)

\end{document}